%% file: main.tex
\documentclass[11pt]{article}

\usepackage[final]{acl}

\usepackage{times}
\usepackage{latexsym}

\usepackage[T1]{fontenc} 

\usepackage[utf8]{inputenc}

\usepackage{microtype}

\usepackage{inconsolata}

\usepackage{graphicx}
\input{sections/new_add}

\title{
\logo\hspace{0.6em}\ours: A Holistic Benchmark for Evaluating\\
Forensic Analysis of AI-Generated Academic Images
}

\author{
Bo Zhang\thanks{Equal contribution.} \quad
Tzu-Yen Ma\footnotemark[1] \quad
Zichen Tang \quad
Junpeng Ding \quad
Zirui Wang \\
\textbf{Yizhuo Zhao} \quad
\textbf{Peilin Gao} \quad
\textbf{Zijie Xi} \quad
\textbf{Zixin Ding} \quad
\textbf{Haiyang Sun} \quad
\textbf{Haocheng Gao} \\
\textbf{Yuan Liu} \quad
\textbf{Liangjia Wang} \quad
\textbf{Yiling Huang} \quad
\textbf{Yujie Wang} \quad
\textbf{Yuyue Zhang} \\
\textbf{Ronghui Xi} \quad
\textbf{Yuanze Li} \quad
\textbf{Jiacheng Liu} \quad
\textbf{Zhongjun Yang} \quad
\textbf{Haihong E}\thanks{Corresponding author.}\\
Beijing University of Posts and Telecommunications\\
\homepage~~\href{https://bupt-reasoning-lab.github.io/AEGIS}{\texttt{bupt-reasoning-lab.github.io/AEGIS}} \\ \github~~\href{https://github.com/BUPT-Reasoning-Lab/AEGIS}{\texttt{BUPT-Reasoning-Lab/AEGIS}} 
\quad
\huggingface~~\href{https://huggingface.co/datasets/BUPT-Reasoning-Lab/AEGIS}{\texttt{BUPT-Reasoning-Lab/AEGIS}}
}

\begin{document}
\maketitle


\input{sections/0_abstract}
\input{sections/1_introduction}

\input{sections/2_benchmark}
\input{sections/3_experiment}
\input{sections/4_related_work}
\input{sections/5_conclusion}
\input{sections/6_limitations}

\input{sections/7_ethical}
\input{sections/acknowledgments}

\bibliography{main}

\appendix
\clearpage
\input{appendix/0_Table_Content}
\input{appendix/A_dataset_annotation_and_construction}

\input{appendix/B_experiment_details}
\input{appendix/C_multi-dimensional_evaluation_analysis}
\input{appendix/D_case_study}
\input{appendix/E_use_of_AI}
\input{appendix/N_page}

\end{document}

%% file: sections/new_add.tex
\usepackage{xspace}

\newcommand{\logo}{%
  \raisebox{-0.4em}{%
    \includegraphics[width=1.9em]{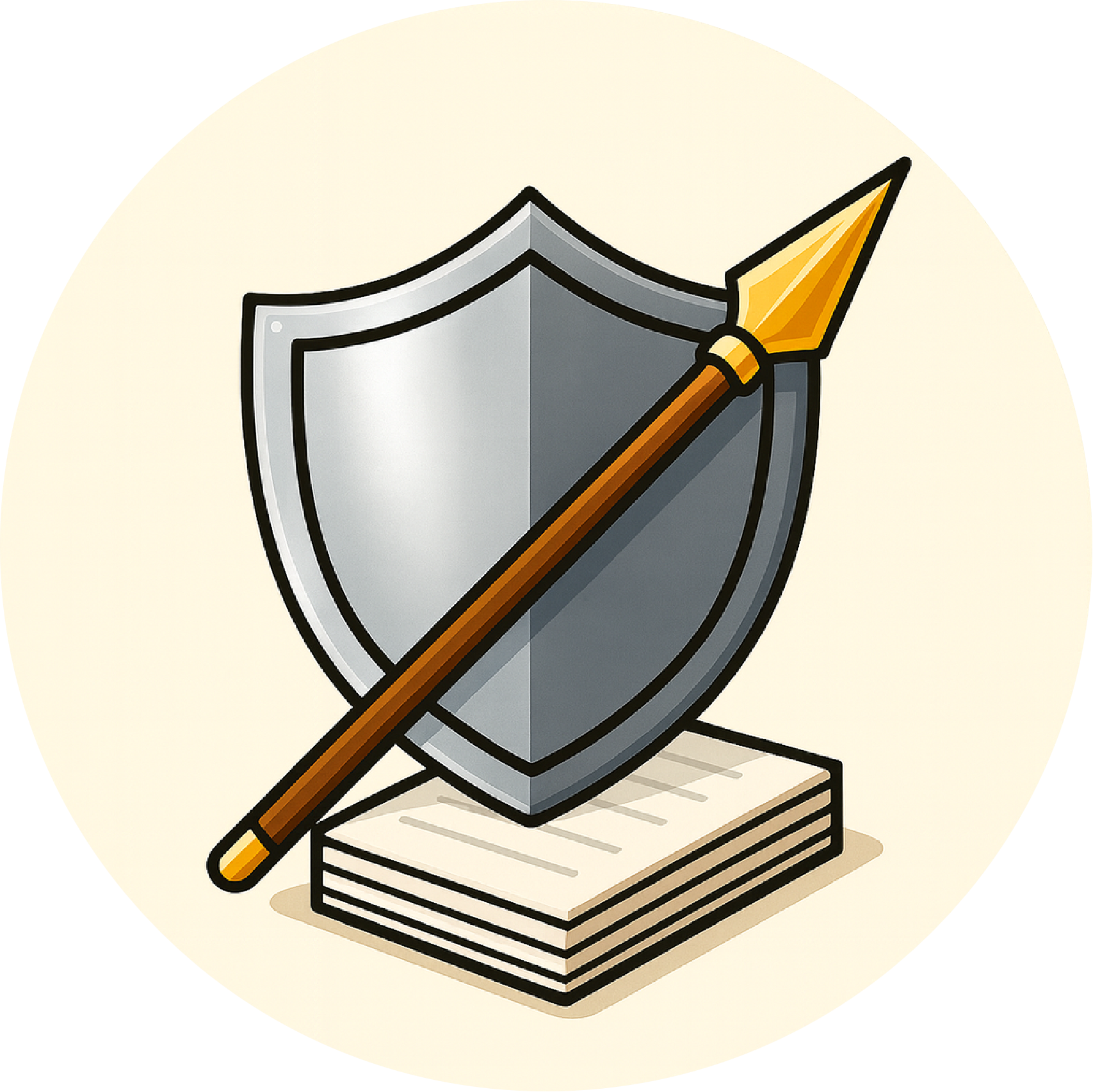}%
  }%
}

\newcommand{\ours}{\textbf{AEGIS}\xspace}

\usepackage{enumitem}

\usepackage{booktabs}   
\usepackage{multirow}   
\usepackage{array} 
\usepackage{adjustbox}

\usepackage{pifont}   
\usepackage{xcolor}   
\usepackage{amssymb}  
\definecolor{my_green}{RGB}{40,154,121}
\definecolor{my_yellow}{RGB}{255,165,0}
\definecolor{my_red}{RGB}{176,46,46}
\newcommand{\correctmark}{\textcolor{my_green}{\ding{52}}} 
\newcommand{\errormark}{\textcolor{my_red}{\ding{56}}}

\newcommand{\eg}{\hbox{e.g.,}\xspace}
\newcommand{\ie}{\hbox{i.e.,}\xspace}

\usepackage[table]{xcolor}   
\usepackage{colortbl} 
\definecolor{wkgreen}{RGB}{198,222,173}
\definecolor{wkyellow}{RGB}{250,228,159}
\newcommand{\best}[1]{%
  \begingroup
  \setlength{\fboxsep}{3pt}
  \colorbox{wkyellow}{\textbf{#1}}%
  \endgroup
}

\newcommand{\second}[1]{%
  \begingroup
  \setlength{\fboxsep}{3pt}%
  \colorbox{wkgreen}{\underline{#1}}%
  \endgroup
}

\usepackage{tabularx}
\usepackage{url}
\usepackage{amsmath}
\definecolor{markblue}{RGB}{74, 98,132}
\definecolor{markyellow}{RGB}{208,139, 54}
\usepackage{makecell}

\usepackage{hyperref}

\newcommand{\homepage}{\raisebox{-1.5pt}{\includegraphics[height=1.2em]{figures/icon/AEGIS.png}}}

\newcommand{\github}{\raisebox{-1.5pt}{\includegraphics[height=1em]{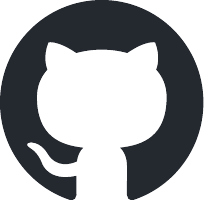}}}

\newcommand{\huggingface}{\raisebox{-1.5pt}{\includegraphics[height=1em]{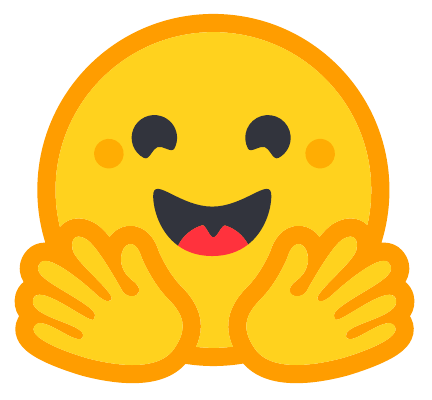}}}

%% file: sections/0_abstract.tex
\vspace{0.3cm}
\begin{abstract}
We introduce \ours, \textbf{A} holistic benchmark for \textbf{E}valuating forensic analysis of AI-\textbf{G}enerated academic \textbf{I}mage\textbf{S}. Compared to existing benchmarks, \ours features three key advances:  
(1) \textbf{Domain-Specific Complexity}: covering seven academic categories with 39 fine-grained subtypes, exposing intrinsic forensic difficulty, where even GPT-5.1 reaches 48.80\% overall performance and expert models achieve only limited localization accuracy (IoU 30.09\%); 
(2) \textbf{Diverse Forgery Simulations}: modeling four prevalent academic forgery strategies across 25 generative models, with 11 yielding average forensic accuracy below 50\%, showing that forensics lag behind generative advances;
and (3) \textbf{Multi-Dimensional Forensic Evaluation}: jointly assessing detection, reasoning, and localization, revealing complementary strengths between model families, with multimodal large language models (MLLMs) at 84.74\% accuracy in textual artifact recognition and expert detectors peaking at 79.54\% accuracy in binary authenticity detection.
By evaluating 25 leading MLLMs, nine expert models, and one unified multimodal understanding and generation model, \ours serves as a diagnostic testbed exposing fundamental limitations in academic image forensics. 
\end{abstract}

%% file: sections/1_introduction.tex
\input{figures/1_Overview}

\input{tables/1_contrast}

\section{Introduction}

With the rapid advancement of generative models, AI-generated images have reached high visual fidelity, raising concerns about the trustworthiness of visual evidence. In response, image forensics has evolved into multiple detection methods, including frequency-domain~\citep{corvi2023detection}, diffusion-process~\citep{pmlr-v235-chen24ay}, patch-level~\citep{chen2024singlesimplepatchneed}, and alignment-based approaches~\citep{rajan2025aligned}. Meanwhile, leveraging visual understanding and reasoning, multimodal large language models (MLLMs) have been increasingly applied to image forgery analysis, either directly~\citep{wen2025spot} or in conjunction with expert models~\citep{xu2025fakeshield}.

Despite recent advances, the applicability of current models to expert-level academic review in high-stakes, domain-specific settings remains underexplored.
From an expert artificial general intelligence (AGI) perspective (Figure~\ref{fig:1_overview}), academic image forensics requires models to progress from coarse authenticity judgment to precise region attribution and spatial grounding, as the academic domain presents three specific challenges:
(1) more complex visual distributions, (2) finer-grained manipulation patterns, and (3) knowledge-intensive reasoning requirements.
This raises the question of \textbf{\textit{whether current models can generalize their capabilities to support the multi-stage forensic analysis required for academic review}}.
As shown in Table~\ref{tab:1_contrast}, existing benchmarks fall short of supporting expert-level forensic analysis due to three key limitations:

\begin{itemize} [leftmargin=*]
    \item \textbf{\textit{Generic-Scene Bias.}}  \emph{Academic images are characterized by fine-grained, structured, and semantically dense visual content that poses challenges for forensic interpretation.} In contrast, benchmarks like Semi-Truths~\citep{NEURIPS2024_d5cdf7e5} and ForensicHub~\citep{du2025forensichubunifiedbenchmark} focus on generic imagery such as faces, natural scenes, or social media content with coarser structures, limiting their suitability for academic forensic reasoning.

    \item \textbf{\textit{Restricted Forgery Strategies.}} 
    \emph{In the context of academic imagery, forgeries exhibit diverse forms with high degrees of subtlety.} However, benchmarks such as AIGuard~\citep{zhang-etal-2025-aiguard} and GRE~\citep{10.1145/3664647.3681445} primarily focus on global forgery or single-type editing forgery, failing to simulate the complex and misleading manipulation behaviors.

    \item \textbf{\textit{Forensic-Agnostic Evaluation Protocols.}} 
    \emph{Expert academic review requires evidence-grounded forensic assessment beyond single authenticity prediction.} However, benchmarks such as AIGIBench~\citep{li2025is} and DFBench~\citep{10.1145/3746027.3758204} focus on detection accuracy, overlooking key forensic dimensions such as manipulation scope and localization.
\end{itemize}

\input{figures/2_Category}
To fill this gap, we introduce \ours, a holistic benchmark comprising over 20k forensic questions, incorporating advanced generative models and simulating diverse forgery scenarios. 

\begin{itemize} [leftmargin=*]
    \item \textbf{\textit{Domain-Specific Complexity.}} \ours captures the complexity of academic imagery through systematic coverage of \textbf{seven major categories and 39 fine-grained subtypes}. Our taxonomy spans diverse scientific visual forms, including microscopic particles, cellular structures, and heterogeneous imaging modalities (\eg{ microscopy, medical, and physical imaging}), enabling rigorous forensic evaluation in scholarly settings.

    \item \textbf{\textit{Diverse Forgery Simulations.}} \ours covers four representative forgery types commonly found in academic publications, including (1) Text Constraint Fabrication, (2) Image Inference Forgery, (3) Targeted Region Restoration, and (4) Targeted Region Editing. These forgeries are instantiated using \textbf{25 state-of-the-art generative models}, enabling realistic and adversarial simulation of academic image manipulation.

    \item \textbf{\textit{Multi-Dimensional Forensic Evaluation.}} \ours introduces a multi-level and multi-perspective evaluation suite, including (1) Forgery Scope Discrimination, (2) Textual Artifact Recognition, (3) Manipulation Classification, and (4) Tampering Pinpointing. These tasks comprehensively assess models' capabilities in judgment accuracy, localization precision, and explainability depth, closely aligning with real-world academic review needs.
\end{itemize}

We evaluate 14 proprietary MLLMs, 11 open-source MLLMs,
one representative unified multimodal understanding and generation model, 
as well as six vision-only expert models and three hybrid MLLM-assisted expert models. The experimental results reveal three key findings:

\begin{itemize} [leftmargin=*]
    \item \textbf{\textit{Domain-specific complexity challenges detection and localization.}} Both MLLMs and expert models struggle on academic images, with GPT-5.1 achieving 48.80\% overall performance and expert baselines exhibiting an approximately 30\% Real--Forgery F1 gap. This pattern indicates a structural bias toward geometric regularities, failing to generalize to the high-variance, texture-rich nature of academic imagery and leading to severely limited localization accuracy (IoU 30.09\%, with only two MLLMs exceeding 50\% region-level accuracy).

    \item \textbf{\textit{Generative diversity highlights forensic vulnerabilities.}} Generative diversity exposes an adversarial asymmetry between generation and forensic analysis. Averaged across models, 11 generative models reduce accuracy below 50\%, with four falling below 30\%, indicating that existing static forensic cues struggle to generalize against the rapid evolution of generative models.

    \item \textbf{\textit{Comprehensive forensic reasoning requires complementarity.}} MLLMs excel at forensic reasoning, achieving 84.74\% accuracy in textual artifact recognition and 60.07\% in manipulation classification, capabilities beyond expert models. In contrast, expert detectors achieve over 70\% accuracy in binary authenticity detection but are less robust to post-processing perturbations, indicating that reliable academic image scrutiny requires quantitatively complementary forensic evidence rather than any single paradigm.
\end{itemize}

%% file: figures/1_Overview.tex
\begin{figure}[t]
    \vspace{10pt}
    \centering
    \includegraphics[width=0.98\columnwidth]{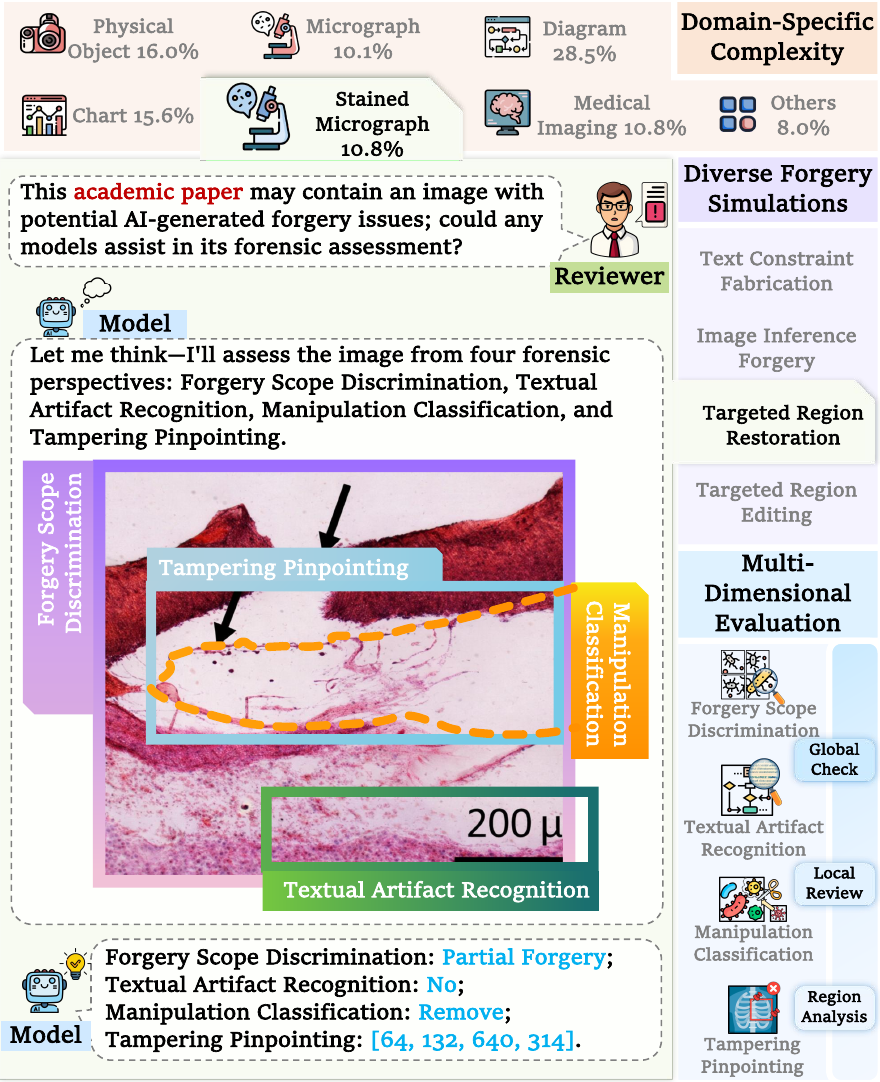}
    \caption{
		\ours investigates whether current models can effectively audit \emph{AI-generated images} in academic papers by performing holistic forensic analysis across four complementary tasks. 
    }
    \label{fig:1_overview}
    \vspace{-10pt}
\end{figure}

%% file: tables/1_contrast.tex
\begin{table*}[htbp]
  \centering
  \small
  \setlength{\tabcolsep}{1mm}
  
  \begin{tabular}{l m{1.6cm} c c c cc ccc cccc}
    \toprule
    \multirow{2}{*}{\textbf{Benchmark}} &
    \multirow{2}{*}{\textbf{Domain}}&
    \multirow{2}{*}{\textbf{Tasks}} &
    \multirow{2}{*}{\textbf{\# GMs}} &
    \multicolumn{5}{c}{\textbf{Forgery Strategies}} &
    \multicolumn{4}{c}{\textbf{Evaluation Tasks}} & 
    \multirow{2}{*}{\textbf{\# Samples}}
    \\
    \cmidrule(lr){5-9} \cmidrule(lr){10-13} 
    & & & & TCF & IIF & TRR & TRE & Total & FSD & TAR & MC & TP \\
    \midrule
    
    GenImage (NeurIPS'23) & 
     \includegraphics[height=3ex]{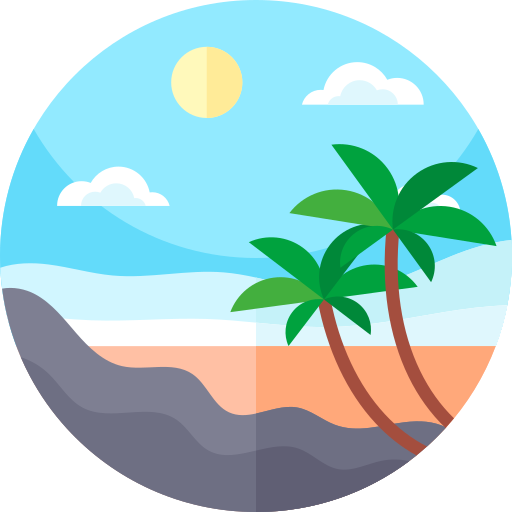} \includegraphics[height=3ex]{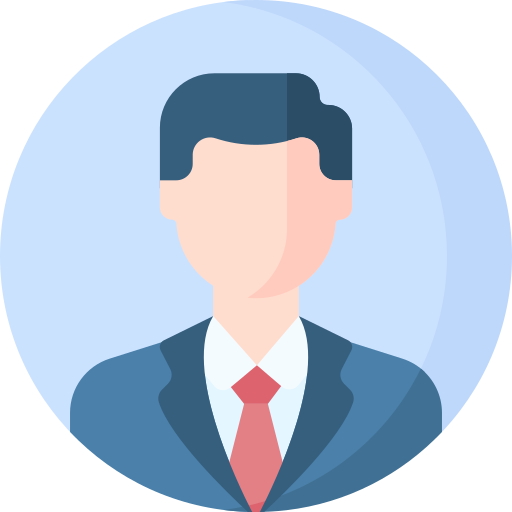} \includegraphics[height=3ex]{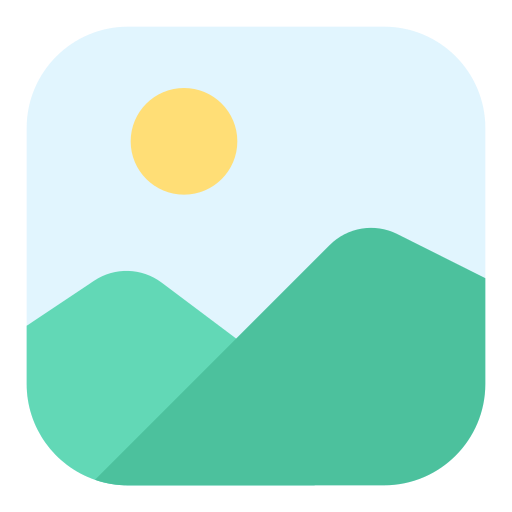} 
    & D & 6 & \correctmark & \correctmark & \errormark & \errormark & 2 & \correctmark & \errormark & \errormark & \errormark & 2.7M \\
    
    Semi-Truths (NeurIPS'24) & 
    \includegraphics[height=3ex]{figures/icon/Nature_Image.png} \includegraphics[height=3ex]{figures/icon/Generic_Image.png} 
    & D & 8 & \errormark & \correctmark & \correctmark & \errormark & 2 & \correctmark & \errormark & \errormark & \errormark & 1.5M \\

    AIGIBench (NeurIPS'25) & 
    \includegraphics[height=3ex]{figures/icon/Nature_Image.png} \includegraphics[height=3ex]{figures/icon/Human_Image.png} \includegraphics[height=3ex]{figures/icon/Generic_Image.png} 
    & D & 16 & \correctmark & \errormark & \errormark & \errormark & 1 & \correctmark & \errormark & \errormark & \errormark & 429k \\

    ForensicHub (NeurIPS'25)    & 
    \includegraphics[height=3ex]{figures/icon/Human_Image.png} \includegraphics[height=3ex]{figures/icon/Generic_Image.png}  
    & D+L & 10 & \correctmark & \errormark & \errormark & \correctmark & 2 & \correctmark & \errormark & \errormark & \correctmark & 4M \\
    
    GIM (AAAI'25) & \includegraphics[height=3ex]{figures/icon/Nature_Image.png} \includegraphics[height=3ex]{figures/icon/Generic_Image.png} 
    & D+L &  3  & \errormark & \errormark & \errormark & \correctmark & 1 & \correctmark & \errormark & \errormark & \correctmark & 2M \\
    
    GRE (ACM MM'24) & 
    \includegraphics[height=3.2ex]{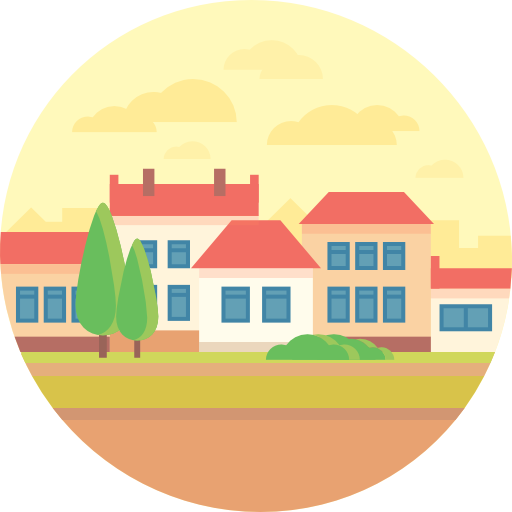} \includegraphics[height=3ex]{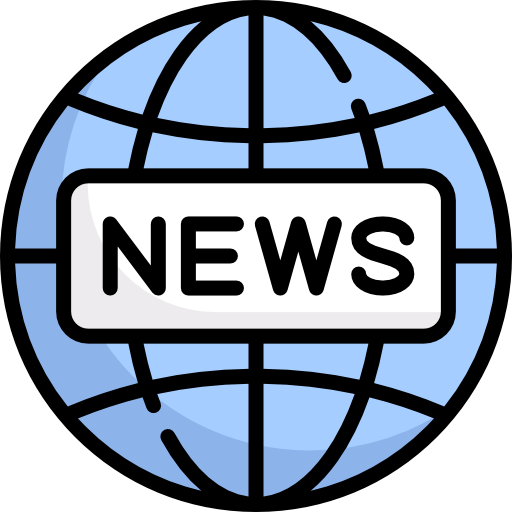}  
    & D+L & 3  & \errormark & \errormark & \errormark & \correctmark & 1 & \correctmark & \errormark & \correctmark & \correctmark & 2.3k \\
    
    DFBench (ACM MM'25) & 
    \includegraphics[height=3ex]{figures/icon/Nature_Image.png} \includegraphics[height=3ex]{figures/icon/Human_Image.png} \includegraphics[height=3ex]{figures/icon/Generic_Image.png} 
    & D & 12 & \correctmark & \correctmark & \correctmark & \correctmark & 4 & \correctmark & \errormark & \errormark & \errormark & 540k \\

    AIGuard (ACL Findings'25)   & 
    \includegraphics[height=3ex]{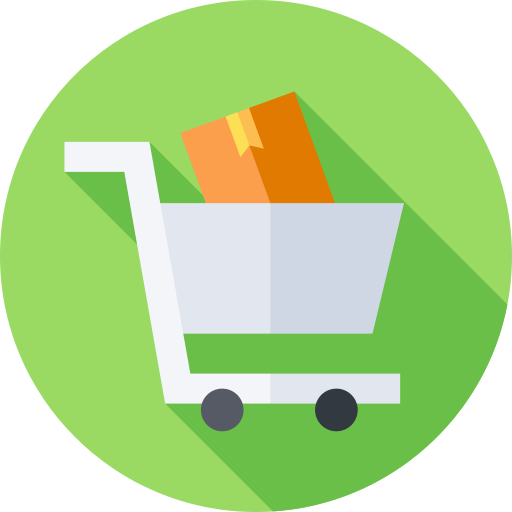}   
    & D+E & 1 & \errormark &  \correctmark & \errormark & \errormark & 1 & \correctmark & \errormark & \errormark & \errormark & 253k \\
    
    
    
    \midrule
    \textbf{\ours (ours)}   &\textbf{Academic} 
    & \textbf{D+L} & \textbf{25} 
    & \textbf{\correctmark} 
    & \textbf{\correctmark} 
    & \textbf{\correctmark} 
    & \textbf{\correctmark} 
    & \textbf{4} 
    & \textbf{\correctmark} 
    & \textbf{\correctmark} 
    & \textbf{\correctmark} 
    & \textbf{\correctmark} & \textbf{20k} \\
    
    \bottomrule
  \end{tabular} 
  
  \caption{
        \textbf{Comparison of \ours and other benchmarks.}      
		\includegraphics[height=2.2ex]{figures/icon/Nature_Image.png}: Nature;    
        \includegraphics[height=2.2ex]{figures/icon/Human_Image.png}: Human;  
        \includegraphics[height=2.2ex]{figures/icon/Generic_Image.png}: General;    
        \includegraphics[height=2.2ex]{figures/icon/Life_Image.png}: Daily Life;  
        \includegraphics[height=2.2ex]{figures/icon/News_Image.png}: News; 
        \includegraphics[height=2.2ex]{figures/icon/E-commerce.png}: E-commerce. 
        \textbf{D}: Detection; 
        \textbf{L}: Localization; 
        \textbf{E}: Explanation. 
        \textbf{GMs}: Generative Models (excluding pre-2022 models, which are primarily GAN-based).
        \textbf{TCF}: Text Constraint Fabrication; \textbf{IIF}: Image Inference Forgery; \textbf{TRR}: Targeted Region Restoration; \textbf{TRE}: Targeted Region Editing. 
        \textbf{FSD}: Forgery Scope Discrimination; \textbf{TAR}: Textual Artifact Recognition; \textbf{MC}: Manipulation Classification; \textbf{TP}: Tampering Pinpointing.}
        
  \label{tab:1_contrast}
\end{table*}

%% file: figures/2_Category.tex
\begin{figure*}[t]
	\centering
	\includegraphics[width=0.98\textwidth]{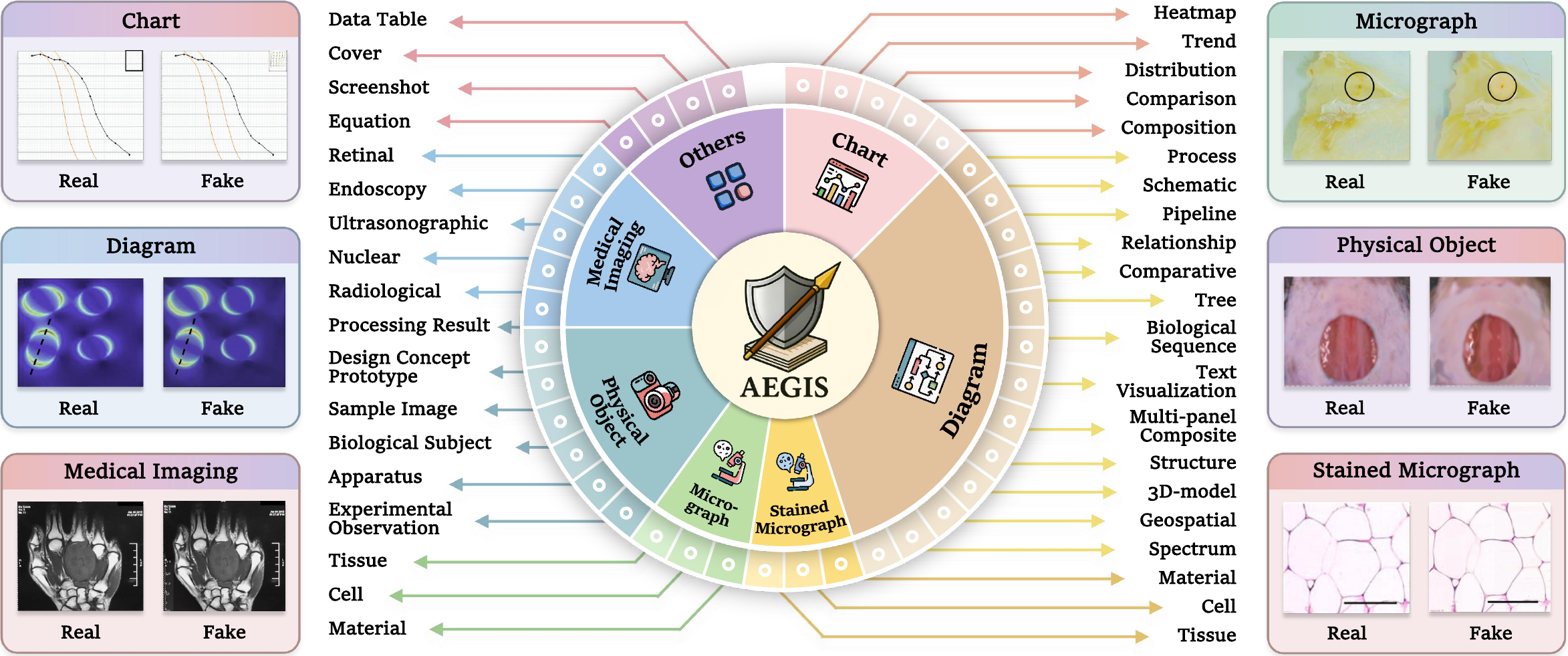}
	\caption{
		\textbf{Hierarchical taxonomy of \ours.} We organize academic images into seven categories and 39 fine-grained subtypes based on their structural and semantic characteristics. Real \textbf{(left)} and fake \textbf{(right)} examples are shown for comparison, where ``fake'' refers to AI-generated forgeries.}
	\label{fig:2_category}
\end{figure*}

%% file: sections/2_benchmark.tex
\section{\ours Benchmark}

\input{figures/3_Pipeline}

\subsection{Overview of \ours}
We introduce \ours, a comprehensive benchmark for systematic evaluation of model performance in academic image forensics. As illustrated in Figure~\ref{fig:2_category}, \ours establishes a hierarchical taxonomy with seven main categories (\ie{ Chart, Medical Imaging, Physical Object, Micrograph, Stained Micrograph, Diagram, and Others}), comprising 39 annotated subtypes. \ours is constructed based on forgery patterns inspired by genuine retraction cases and academic review observations. By synthesizing evidence from public platforms (\eg Retraction Watch\footnote{\url{https://retractionwatch.com}} and PubPeer\footnote{\url{https://pubpeer.com}}) and expert analysis, we identify four recurring forgery strategies. Qualitative expert evaluation, detailed in Appendix~\ref{subsec:A0_Qualitative expert evaluation}, further confirms the consistency between our synthetic data and real-world misconduct cases.

\subsection{Dataset Construction}

\paragraph{Data Curation.} Papers were parsed into figures and panels, then manually verified through expert curation to ensure data quality.


\begin{itemize}[leftmargin=*]
    \item \textbf{\textit{Paper Parsing.}} \ours collected more than 4,000 high-quality academic papers from the open-access PubMed Central (PMC) repository\footnote{\url{https://pmc.ncbi.nlm.nih.gov}} and performed document-level parsing to extract figures, captions, and panels, where a panel denotes the smallest indivisible visual unit in an academic image. Details for paper selection criteria are in Appendix~\ref{subsec:A2_Data Curation}.

    \item \textbf{\textit{Expert Curation and Validation.}} We conducted a two-stage quality control process involving manual screening and expert review. Panels with insufficient resolution, parsing errors, or lacking academic semantics were removed, while panels consistent with the predefined hierarchical taxonomy were retained and annotated. Non-academic or non-forgeable visual elements (\eg{ portraits, icons, and decorative graphics}) were excluded, resulting in over 8,000 high-quality annotated panels for dataset construction.
\end{itemize}

\paragraph{Generative Models.} To ensure representative quality and diversity for forensic evaluation, \ours incorporates 25 advanced generative models~\citep{Imagen4,Lugmayr_2022_CVPR} from three major architectures: (1) diffusion-based (\eg Flux, Midjourney V6/V7, DALL·E 2/3, and kling); (2) hybrid (\eg Wan2.1-T2I-Turbo and GPT Image 1); and (3) unified multimodal understanding and generation (\eg Janus-Pro-7B~\citep{chen2025janusprounifiedmultimodalunderstanding}). More details are in Appendix~\ref{subsubsec:A3_1_Generative Models}.

\input{figures/4_QA}
\paragraph{Forgery Strategy Simulations.} As illustrated in Figure~\ref{fig:3_pipeline}, to simulate realistic forgery behaviors in academic scenarios, \ours defines four representative forgery strategies, inspired by common manipulation practices in academic imagery and aligned with the capabilities of contemporary generative models.
More details are in Appendix~\ref{subsubsec:A3_2_Forgery Simulations}.

\begin{itemize} [leftmargin=*]
    \item \textbf{\textit{Text Constraint Fabrication (TCF). }}\emph{TCF models the fabrication of entire academic images from scratch under textual constraints.} We reconstructed captions of authentic academic images into semantically equivalent prompts, which were then used to guide text-to-image generative models, resulting in 3,121 Entire Forgeries.
    
    \item \textbf{\textit{Image Inference Forgery (IIF). }}\emph{IIF generates realistic and detailed globally forged images based on the original input, simulating reference-guided generation.} By using authentic images as references, we instructed generative models to synthesize globally consistent forgeries, resulting in 2,274 Entire Forgeries.
    
    \item \textbf{\textit{Targeted Region Restoration (TRR). }}\emph{TRR simulates the forgery risks associated with reconstructing missing local content.} 
    By applying masks to specific regions of authentic images, we instructed generative models to restore these areas, resulting in 1,650 Partial Forgeries. 

    \item \textbf{\textit{Targeted Region Editing (TRE). }}\emph{TRE emulates common ``lightweight tampering'' operations.} By providing masks or textual instructions, we performed localized edits on authentic images, including alterations, insertions, or removals, to produce 1,165 Partial Forgeries. 
\end{itemize}

\paragraph{Data Quality Assurance.} To mitigate hallucinations from generative models, all AI-generated images underwent a dual-review protocol, evaluating both local manipulation plausibility and global visual fidelity. 
This process led to the discard of 29\% of AI-generated images, resulting in a final set of 8,210 high-quality samples. 
Verification, performed by five trained annotators, took approximately 200 hours. 
Detailed assessment is shown in Appendix~\ref{subsubsec:A3_3_Data Quality Assurance}.

\subsection{Design of Evaluation Questions} 
\ours designs a set of progressively structured tasks to evaluate forensic capability from global authenticity judgment to fine-grained spatial localization, as shown in Figure~\ref{fig:4_QA}.  

\begin{itemize} [leftmargin=*]
    \item \textbf{\textit{Forgery Scope Discrimination (FSD).}} This task evaluates coarse-grained authenticity judgment, requiring models to determine whether an input image exhibits no forensic artifacts, contains regionally constrained AI-generated traces, or represents a comprehensively synthesized fabrication. 
    To enable abstention when confidence levels are insufficient, the prompt explicitly incorporates ``\textit{Not Sure}'' as an option. 
    
    \item \textbf{\textit{Textual Artifact Recognition (TAR).}} This task assesses fine-grained visual cue perception by requiring models to perform binary classification on whether textual regions in an input image exhibit traces of AI synthesis, relying on text region recognition and semantic consistency analysis.
    
    \item \textbf{\textit{Manipulation Classification (MC).}} This task evaluates structure-aware reasoning, requiring models to infer the type of manipulation applied to an image. Given a red-highlighted manipulated region and the original image caption, the model analyzes the structural and contextual role of the edited area to classify the manipulation as insertion, removal, or alteration.

    \item \textbf{\textit{Tampering Pinpointing (TP).}} This task focuses on fine-grained spatial grounding, evaluating the localization accuracy of partially edited regions in academic images. We adopt an adaptive granularity protocol: region-level bounding boxes for MLLMs and pixel-level masks for expert models, emphasizing spatial precision beyond global visual cues.
\end{itemize}

%% file: figures/3_Pipeline.tex
\begin{figure*}[t]
	\centering
	\includegraphics[width=0.98\textwidth]{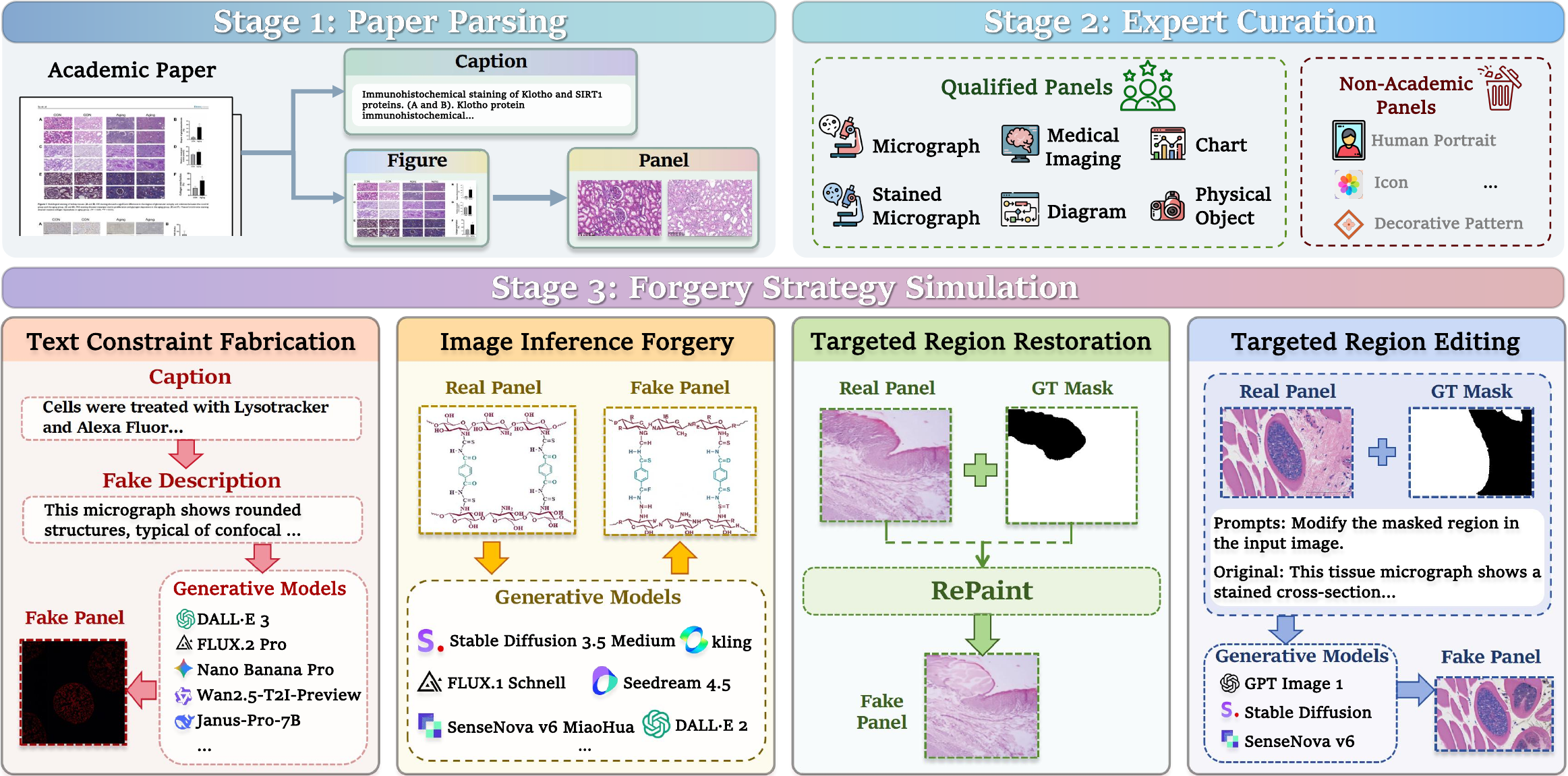}
	\caption{
        \textbf{Construction pipeline of \ours.}
        \textbf{Stage 1: Paper Parsing} extracts figures, captions, and panels from papers.
        \textbf{Stage 2: Expert Curation} retains qualified academic panels while excluding non-academic ones.
        \textbf{Stage 3: Forgery Strategy Simulation} synthesizes AI-generated academic image forgeries via four representative strategies.
        }
	\label{fig:3_pipeline}
\end{figure*}

%% file: figures/4_QA.tex
\begin{figure*}[t]
	\centering
	\includegraphics[width=0.98\textwidth]{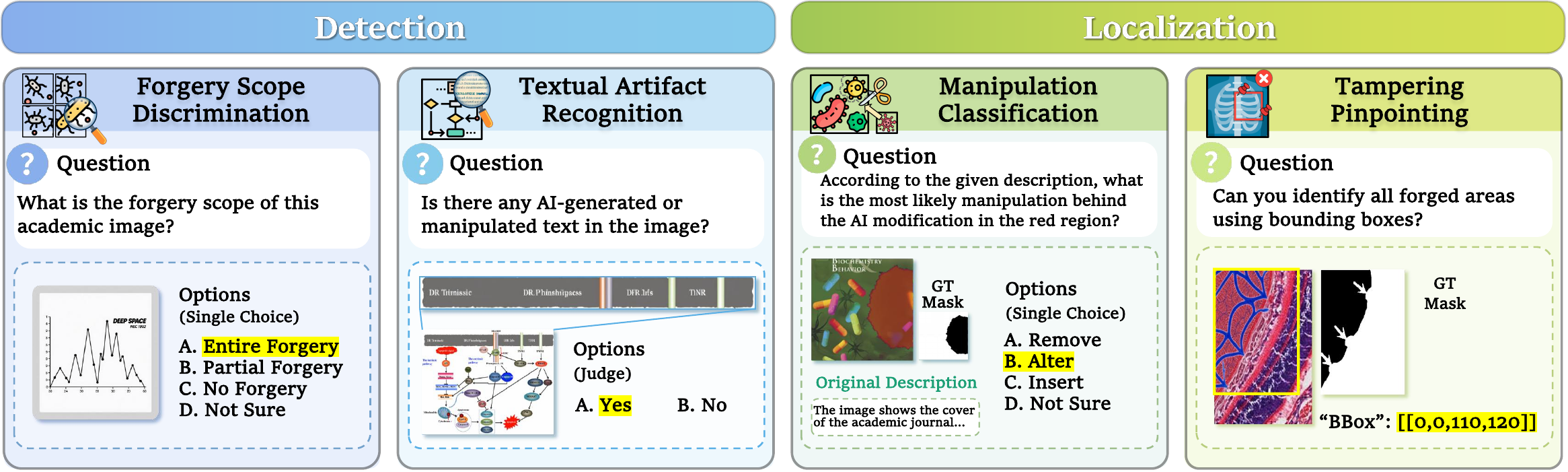}
	\caption{
		\textbf{Evaluation design of \ours.} 
        Four evaluation question types are designed to support staged forensic analysis, 
        from global authenticity assessment to fine-grained region-level localization,
        including \textbf{Forgery Scope Discrimination}, \textbf{Textual Artifact Recognition}, \textbf{Manipulation Classification}, and \textbf{Tampering Pinpointing}.
        }
	\label{fig:4_QA}
\end{figure*}

%% file: sections/3_experiment.tex
\input{tables/2_main_result}

\section{Experiments}
\subsection{Experimental Setup}

\input{figures/5_Experiment}

\paragraph{Benchmarked Models. }
We evaluate 14 proprietary MLLMs~\citep{GPT-4.1,o4mini,gemini2.5flashprpreview}, 11 open-source MLLMs~\citep{li2024llavanextinterleavetacklingmultiimagevideo,gemmateam2025gemma3technicalreport,llama4maverick}, one unified multimodal understanding and generation model~\citep{chen2025janusprounifiedmultimodalunderstanding}, and nine expert models, including six vision-only and three hybrid models, of which only SIDA~\citep{Huang_2025_CVPR} supports multi-granular forgery discrimination.

\paragraph{Evaluation Protocol.} We uniformly adopt high-resolution PNG images for all evaluations to avoid compression artifacts introduced by lossy formats such as JPEG, which may interfere with forensic cues and bias model judgments. 

\paragraph{Evaluation Metrics.} We adopt task-specific metrics to systematically evaluate model performance.

\begin{itemize}[leftmargin=*]
    \item \textbf{\textit{Tasks FSD, TAR, and MC.}} We employ Accuracy (ACC) and Macro-F1 Score (M-F1) to evaluate performance across tasks.

    \item \textbf{\textit{Task TP.}} We adopt an adaptive evaluation protocol: pixel-level metrics, including Intersection over Union (IoU) and F1 Score (F1), evaluate expert-generated masks, while region-level metrics assess MLLM-predicted bounding boxes, where Correct Localization Accuracy (CLA) measures coverage and Over Localization Rate (OLR) penalizes over-localization.
\end{itemize}

We further propose a new composite metric, the \textbf{Normalized Forensic Index (NFI)}, to reflect balanced forensic capability across tasks rather than peak performance on any single dimension. Further details are provided in Appendix~\ref{subsec:B3_Evaluation Metrics}.

\subsection{Experimental Analysis}
\subsubsection{Domain-Specific Complexity Challenges Current Models} We first investigate how the intrinsic complexity of academic imagery impacts forensic capability. Our analysis suggests that the high variance between structured and dense scientific visual forms poses a substantial challenge to the generalization capability of current architectures.

\begin{itemize} [leftmargin=*]
    \item \textbf{\textit{A holistic capability gap prevents any model from meeting the multifaceted forensic demands.}} As detailed in Table~\ref{tab:main_results}, models excel only at isolated tasks, a limitation quantified by the Normalized Forensic Index (NFI) where no evaluated MLLM exceeds 60\%, indicating limited balanced forensic capability.

    \item \textbf{\textit{Tampering Pinpointing remains a major bottleneck.}} Across all expert models, the highest pixel-level IoU reaches only 30.09\%, while among MLLMs, only two models exceed 50\% CLA.

    \item  \textbf{\textit{Dense semantics from visual density bias impede generalization.}} Performance correlates heavily with visual density. Figure~\ref{fig:5_experiment}(a) shows that robust generalization on structured images (\eg Chart and Diagram) contrasts sharply with marked degradation on dense categories (\eg Stained Micrograph and Medical Imaging). This suggests a structural bias: current models tend to over-rely on explicit geometric regularities, while struggling to capture the complex texture details.
    
    \item  \textbf{\textit{An authenticity modeling deficit causes expert models to exhibit a pronounced forgery bias.}} The challenge of modeling diverse scientific distributions predisposes expert models toward biased prediction strategies. Figure~\ref{fig:5_experiment}(b) shows most expert baselines achieve lower Real-F1 than Forgery-F1. This imbalance suggests that expert models, unable to encompass the manifold of \textit{authentic} academic imagery, default to flagging unseen patterns as forged.

\end{itemize}

\subsubsection{Generative Diversity Highlights Forensic Vulnerabilities}

Adversarial analysis reveals a capability gap: generative evolution outpaces forensic capability, particularly in fine-grained manipulations.

\begin{itemize} [leftmargin=*]

    \item \textbf{\textit{The granularity gap makes localized forgeries harder than global ones.}} Models perform reliably on globally synthesized images but show notable degradation on localized manipulations. As shown in Figure~\ref{fig:5_experiment}(a), performance on targeted restoration and editing consistently underperforms that on global forgeries, indicating the increased difficulty of localized forgery analysis.

    \item \textbf{\textit{Adversarial asymmetry leads generation to outpace forensic capabilities.}} Figure~\ref{fig:5_experiment}(c) reveals a clear adversarial asymmetry. Averaged across models, four generative models reduce accuracy below 30\% (\eg{ Nano Banana Pro}), while robust defense remains limited to a few models (\eg{ Gemini 3 Pro Preview and GPT-4.1}). This result shows that generative advances degrade performance more broadly than forensic capabilities can adapt, exposing a persistent gap in academic image forensics. 
\end{itemize}

\subsubsection{Towards Expert AGI: Synergizing Detection and Reasoning}
Our findings suggest that no single paradigm is sufficient; instead, the future of academic forensics likely lies in synergizing the sensitivity of experts with the reasoning of MLLMs.

\begin{itemize} [leftmargin=*]

    \item \textbf{\textit{Foreground--background entanglement complicates recognition.}} Figure~\ref{fig:5_experiment}(a) shows that \textit{insertion} and \textit{removal} are consistently harder to identify than \textit{alteration}. Even the strongest MLLM, GPT-5.1, reaches 60.07\% accuracy on Manipulation Classification, indicating reasonable reasoning capability but remaining challenges in separating foreground anomalies from coherent backgrounds.

    \item \textbf{\textit{Complementary synergies require orchestrating sensitive sensors with cognitive agents.}} The results reveal a functional orthogonality between expert models and MLLMs: experts act as precise but fragile visual ``sensors'' for authenticity discrimination, while MLLMs provide stronger semantic reasoning (\eg{ Textual Artifact Recognition}). This contrast is reinforced by their robustness profiles, where vision-centric experts degrade sharply under post-processing perturbations (\ie{ Gaussian blurring, JPEG compression and image scaling}; Figure~\ref{fig:6_post_process}), whereas MLLMs remain comparatively stable, indicating a weaker reliance on low-level visual fidelity. These findings motivate adaptive expert--MLLM orchestration for future \textbf{Expert AGI}.

\end{itemize}

\input{figures/6_Postprocessing}

\subsection{Prompting Strategies}
We evaluate two prompting strategies on \ours, namely \textbf{Chain-of-Thought} (CoT)~\citep{NEURIPS2022_9d560961} and \textbf{Few-Shot} prompting~\citep{NEURIPS2020_1457c0d6}. More details are provided in Appendix~\ref{subsec:C2_Impact of Prompting Strategies}.

\input{figures/7_Prompt}

\begin{itemize} [leftmargin=*]
    \item Few-Shot prompting introduces a clear trade-off (Figure~\ref{fig:7_prompt}): while it enhances pattern-level recognition in detection tasks (\eg{ Forgery Scope Discrimination and Textual Artifact Recognition}), it consistently degrades fine-grained reasoning required for localized spatial comparison (\eg{ Manipulation Classification}). This degradation suggests that few-shot examples bias the model toward shortcut pattern matching, which interferes with the multi-step reasoning required to distinguish subtle manipulation types. 

    \item CoT prompting yields consistent gains across most tasks, while substantially degrading manipulation classification. CoT improves detection-centric tasks, indicating that explicit reasoning helps surface more reliable authenticity cues and localization signals. In contrast, Manipulation Classification suffers a significant drop, suggesting limited capability to perform hypothetical judgments.
\end{itemize}

\subsection{Error Analysis}
We analyze representative error cases from GPT-5.1 by randomly sampling 100 instances for each task.
Full examples and further interpretation are provided in Appendix~\ref{sec:D_Case Study}.

\begin{itemize} [leftmargin=*]
    \item \textbf{\textit{Forgery Scope Discrimination (48\%). }}
    Caused by the model's tendency to overgeneralize localized edits as entire forgeries, or to overlook subtle tampering and misclassify partial forgeries.

    \item \textbf{\textit{Textual Artifact Recognition (22\%). }}
    Failure to detect fine-grained textual anomalies introduced by text-based modifications in partial forgery.
    
    \item \textbf{\textit{Manipulation Classification (36\%). }}
    Stemming from difficulty in distinguishing foreground manipulations from complex background regions.

    \item \textbf{\textit{Tampering Pinpointing (56\%). }}
    Stemming from a tendency to predict fewer but overly large bboxes, resulting in imprecise localization.
\end{itemize}

%% file: tables/2_main_result.tex
\begin{table*}[t!]
\centering
\adjustbox{width=\textwidth}{
\begin{tabular}{l|cc|cc|cc|cc|cc|c}
\toprule
\multirow{4}{*}{\textbf{Model}} 
& \multicolumn{4}{c|}{\textbf{Detection}} 
& \multicolumn{6}{c|}{\textbf{Localization}} 
& \multirow{4}{*}{\textbf{NFI}} \\

\cline{2-11}
& \multicolumn{2}{c|}{\textbf{FSD}} 
& \multicolumn{2}{c|}{\textbf{TAR}} 
& \multicolumn{2}{c|}{\textbf{MC}} 
& \multicolumn{4}{c|}{\textbf{TP}} 
& \\

\cline{2-11}
& \multirow{2}{*}{ACC} & \multirow{2}{*}{M-F1} 
& \multirow{2}{*}{ACC} & \multirow{2}{*}{M-F1} 
& \multirow{2}{*}{ACC} & \multirow{2}{*}{M-F1} 
& \multicolumn{2}{c|}{\textbf{Region-Level}} 
& \multicolumn{2}{c|}{\textbf{Pixel-Level}} 
& \\

\cline{8-11}
&  &  
&  &  
&  &  
& CLA & OLR $\downarrow$
& IoU & F1 
& \\
\midrule
\rowcolor{yellow!10}
\multicolumn{12}{c}{\textbf{Human}} \\
\midrule
Human 
& 44.20 & 37.83 
& 76.14 & 75.02 
& 68.01 & 51.11 
& -- & -- 
& -- & -- 
& -- \\

\midrule
\rowcolor{gray!10}
\multicolumn{12}{c}{\textbf{Proprietary MLLMs}} \\
\midrule

GPT-5.1 
& 50.99 & 46.16 
& 76.43 & 76.38 
& \best{60.07} & \best{48.34} 
& 46.87 & 12.25 
& -- & -- 
& \best{48.80} \\

GPT-4.1 
& 66.34 & 59.98 
& \second{83.57} & \second{82.74} 
& 44.55 & 40.05 
& 25.93 & \best{0.14} 
& -- & -- 
& 43.31 \\

OpenAI o4-mini-high 
& 54.04 & 49.35 
& 80.71 & 80.02 
& 36.80 & 35.54 
& 33.29 & 5.44 
& -- & -- 
& 42.77 \\

Claude Sonnet 4.5 
& 25.36 & 19.62 
& 58.76 & 58.70 
& 44.80 & 39.70 
& 27.41 & 24.32 
& -- & -- 
& 26.83 \\

Gemini 3 Pro Preview 
& 64.37 & 53.44 
& \best{84.74} & \best{84.67} 
& 48.54 & \second{45.42} 
& 39.14 & 20.09 
& -- & -- 
& 45.79 \\

Gemini 2.5 Flash 
& 60.29 & 53.78 
& 81.96 & 80.97 
& 44.80 & 37.28 
& 47.42 & 14.07 
& -- & -- 
& \second{47.02} \\

Doubao-Seed-1.6 
& 41.31 & 35.38 
& 77.23 & 76.90 
& 50.05 & 37.08 
& 36.38 & 0.28
& -- & -- 
& 41.73 \\

Doubao-Seed-1.6-thinking 
& 44.57 & 37.13 
& 77.92 & 77.57 
& 58.47 & 41.85 
& 46.00 & 0.64 
& -- & -- 
& 46.66 \\

Doubao-Seed-1.6-flash 
& 29.26 & 23.77 
& 60.67 & 60.39 
& 35.89 & 30.96 
& \best{52.36} & 1.78 
& -- & -- 
& 36.03 \\

Doubao-1.5-thinking-vision-pro 
& 39.78 & 35.31 
& 75.55 & 75.44 
& 36.65 & 32.33 
& 47.53 & 1.71 
& -- & -- 
& 42.39 \\

Doubao-1.5-vision-pro 
& 39.87 & 32.52 
& 72.35 & 72.32 
& 29.17 & 28.96 
& 26.96 & 2.02 
& -- & -- 
& 34.07 \\

Qwen3-VL-Plus 
& 38.77 & 35.72 
& 79.25 & 78.98 
& \second{59.28} & 36.89 
& 5.76 & 0.53 
& -- & -- 
& 16.53 \\

Qwen-VL-Max 
& 43.24 & 37.30 
& 73.76 & 73.74 
& 52.06 & 36.76 
& 40.30 & 1.32 
& -- & -- 
& 43.01 \\

Qwen-VL-Plus 
& 38.20 & 31.81 
& 51.09 & 48.23 
& 42.99 & 32.23 
& 29.37 & 1.32 
& -- & -- 
& 33.89 \\

\midrule
\rowcolor{gray!10}
\multicolumn{12}{c}{\textbf{Open-Source MLLMs}} \\
\midrule

Gemma 3 27B 
& 44.86 & 42.36 
& 66.88 & 58.39 
& 42.49 & 34.76 
& 48.70 & 28.56 
& -- & -- 
& 37.55 \\

Qwen2.5-VL-72B  
& 37.63 & 33.16 
& 70.81 & 70.76 
& 51.76 & 35.96 
& 32.29 & 0.53 
& -- & -- 
& 38.71 \\

Llama 4 Maverick 
& 39.66 & 37.00 
& 68.37 & 68.32 
& 54.62 & 40.18 
& 14.56 & 2.88 
& -- & -- 
& 29.15 \\

Ministral 3 14B 
& 27.19 & 25.44 
& 62.41 & 62.25 
& 53.62 & 33.89 
& 9.73 & 12.86 
& -- & -- 
& 19.90 \\

LLaVA-NeXT-7B 
& 10.05 & 12.23 
& 53.82 & 53.36 
& 59.04 & 27.01 
& 10.37 & 6.47 
& -- & -- 
& 16.53 \\

LLaVA-NeXT-34B 
& 18.57 & 17.01 
& 41.71 & 35.16 
& 32.65 & 27.19 
& \second{51.01} & 0.78 
& -- & -- 
& 27.74 \\

LLaVA-Interleave-7B 
& 33.53 & 20.61 
& 64.68 & 54.01 
& 26.00 & 26.67 
& 5.65 & 8.28 
& -- & -- 
& 13.61 \\

LLaVA-Interleave-7B-DPO 
& 35.56 & 26.09 
& 59.01 & 53.98 
& 25.15 & 24.99 
& 25.22 & 7.50 
& -- & -- 
& 28.18 \\

LLaVA-OneVision-7B 
& 39.20 & 38.76 
& 50.46 & 50.45 
& 37.80 & 31.34 
& 3.94 & 9.98 
& -- & -- 
& 11.45 \\

LLaVA-OneVision-72B 
& 30.37 & 28.57 
& 56.28 & 55.92 
& 55.31 & 39.42 
& 33.89 & 17.05 
& -- & -- 
& 33.81 \\

LLaVA-OneVision-7B-Chat 
& 40.54 & 39.94 
& 50.44 & 50.26 
& 39.47 & 32.45 
& 4.14 & 6.45 
& -- & -- 
& 12.19 \\

\midrule
\rowcolor{gray!10}
\multicolumn{12}{c}{\textbf{Unified Multimodal Understanding and Generation Model}} \\
\midrule

Janus-Pro-7B 
& 24.97 & 27.18 
& 60.25 & 42.59  
& 24.18 & 24.10 
& 5.31 & \second{0.25} 
& -- & -- 
& 13.77 \\

\midrule
\rowcolor{gray!10}
\multicolumn{12}{c}{\textbf{Vision-Only Expert Models}} \\
\midrule

DRCT~\citep{pmlr-v235-chen24ay} $^*$ 
& 55.05 & 39.55 
& -- & -- 
& -- & -- 
& -- & -- 
& -- & -- 
& -- \\

DIRE~\citep{Wang_2023_ICCV} $^*$ 
& 58.16 & 50.41 
& -- & -- 
& -- & -- 
& -- & -- 
& -- & -- 
& -- \\

AIDE~\citep{yan2025a} $^*$ 
& \best{79.54} & 44.74 
& -- & -- 
& -- & -- 
& -- & -- 
& -- & -- 
& -- \\

AlignedForensics~\citep{rajan2025aligned} $^*$ 
& 47.57 & 42.48 
& -- & -- 
& -- & -- 
& -- & -- 
& -- & -- 
& -- \\

Denoising Trajectory~\citep{liang2025denoising} $^*$ 
& 73.68 & \second{65.37} 
& -- & -- 
& -- & -- 
& -- & -- 
& -- & -- 
& -- \\

DDA~\citep{chen2025dual} $^*$ 
& \second{74.19} & \best{67.07} 
& -- & -- 
& -- & -- 
& -- & -- 
& -- & -- 
& -- \\

\midrule
\rowcolor{gray!10}
\multicolumn{12}{c}{\textbf{Hybrid MLLM-Assisted Expert Models}} \\
\midrule

FakeShield~\citep{xu2025fakeshield} $^*$
& 59.72 & 62.08 
& -- & -- 
& -- & -- 
& -- & -- 
& \best{30.09} & \best{69.11} 
& -- \\

SIDA~\citep{Huang_2025_CVPR} 
& 52.91 & 64.38 
& -- & -- 
& -- & -- 
& -- & -- 
& \second{3.66} & \second{54.28} 
& -- \\

FakeVLM~\citep{wen2025spot} $^*$ 
& 58.73 & 55.39 
& -- & -- 
& -- & -- 
& -- & -- 
& -- & -- 
& -- \\

\bottomrule
\end{tabular}
}
\caption{
\textbf{Performance on \ours across four dimensions.} $^*$ denotes binary authenticity classifiers. 
\textbf{FSD}: Forgery Scope Discrimination; \textbf{TAR}: Textual Artifact Recognition; \textbf{MC}: Manipulation Classification; \textbf{TP}: Tampering Pinpointing. 
\textbf{CLA}: Correct Localization Accuracy; \textbf{OLR}: Over Localization Rate; \textbf{NFI}: Normalized Forensic Index.
}
\label{tab:main_results}
\end{table*}

%% file: figures/5_Experiment.tex
\begin{figure*}[t]
	\centering
	\includegraphics[width=0.98\textwidth]{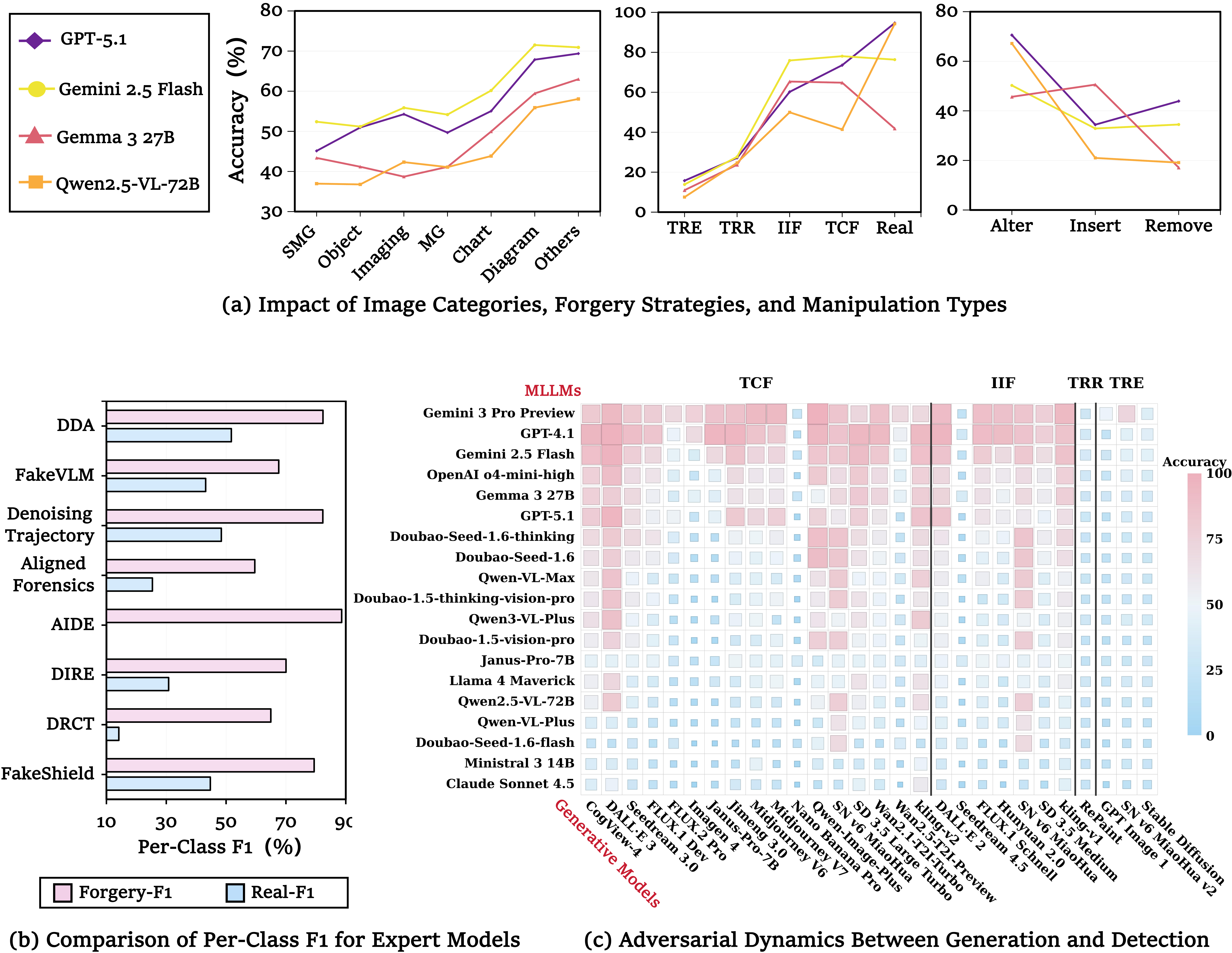}
	\caption{
		\textbf{Fine-grained experimental analysis of \ours.} 
        \textbf{TCF}: Text Constraint Fabrication; \textbf{IIF}: Image Inference Forgery; \textbf{TRR}: Targeted Region Restoration; \textbf{TRE}: Targeted Region Editing. 
        \textbf{SMG}: Stained Micrograph; 
        \textbf{MG}: Micrograph. 
        \textbf{Seedream}: Doubao-Seedream; 
        \textbf{SD}: Stable Diffusion; 
        \textbf{SN}: SenseNova. 
        }
	\label{fig:5_experiment}
\end{figure*}

%% file: figures/6_Postprocessing.tex
\begin{figure}[t]
	\centering
	\includegraphics[width=0.98\columnwidth]{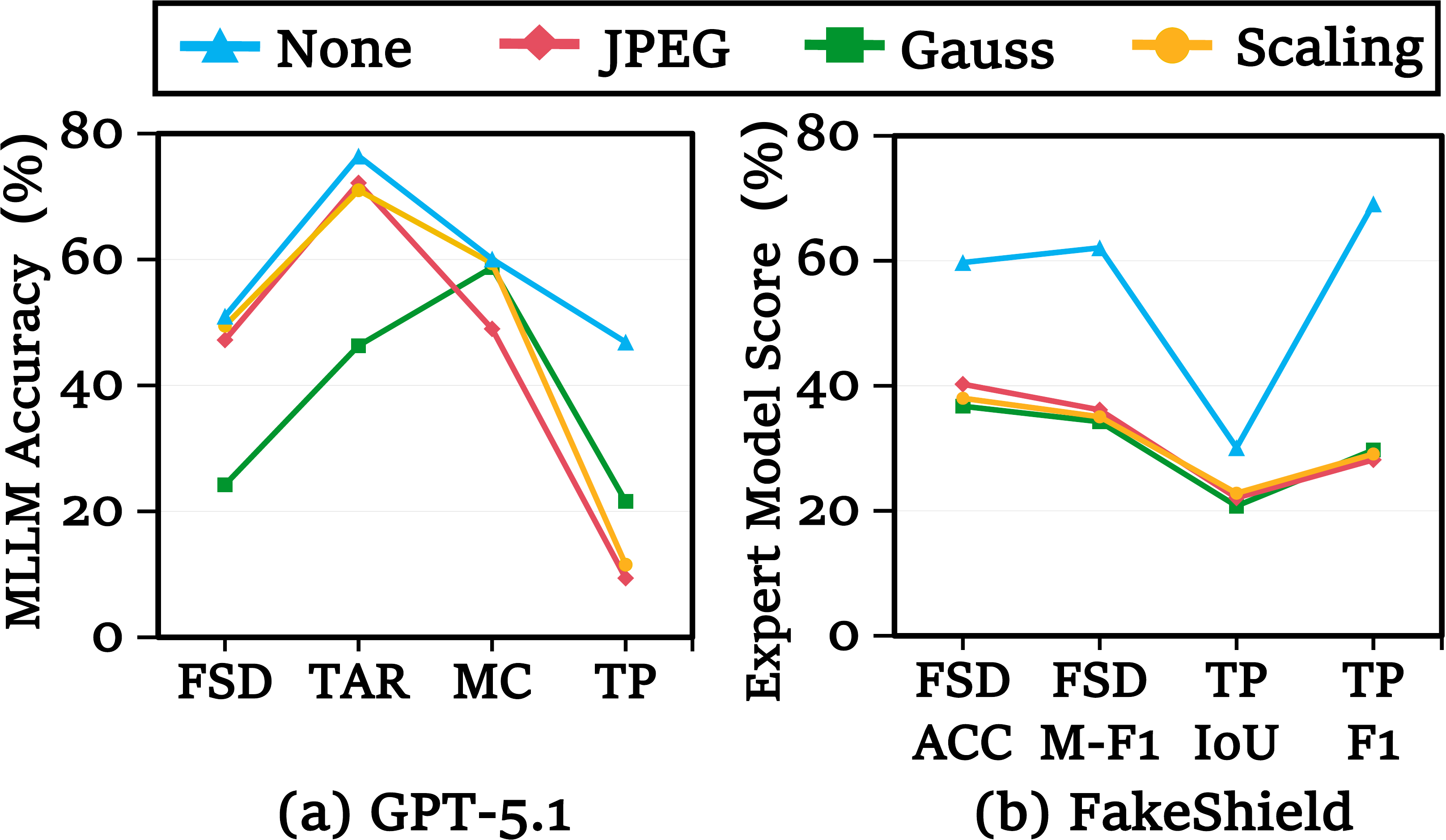}
	\caption{
		\textbf{Impact of post-processing perturbations on \ours.} 
		\textbf{FSD}: Forgery Scope Discrimination; 
		\textbf{TAR}: Textual Artifact Recognition; 
		\textbf{MC}: Manipulation Classification; 
		\textbf{TP}: Tampering Pinpointing.
	}
	\label{fig:6_post_process}
 \end{figure}

%% file: figures/7_Prompt.tex
\begin{figure}[t]
	\centering
	\includegraphics[width=0.98\columnwidth]{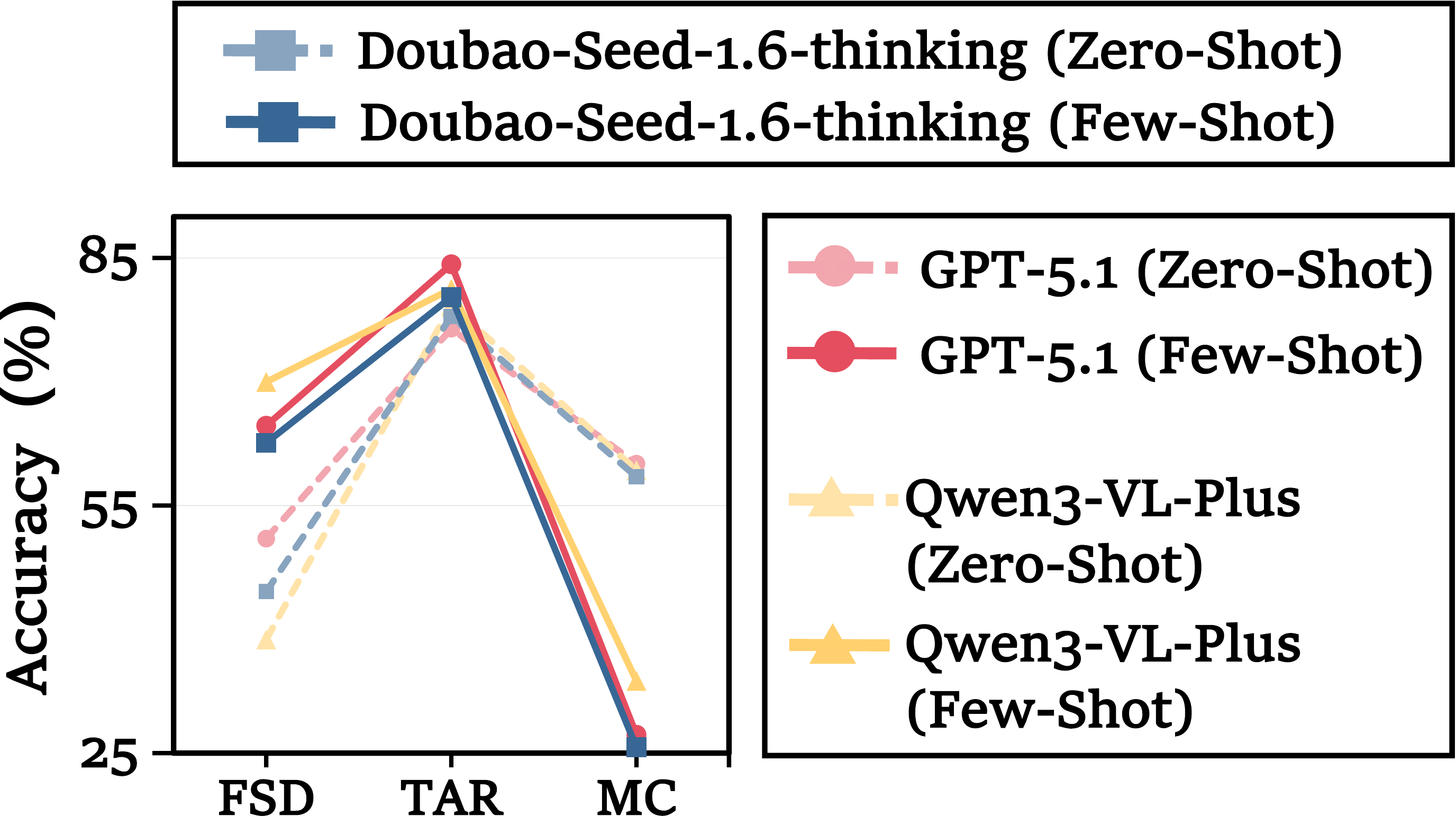}
	\caption{
		\textbf{Impact of Few-Shot prompting on \ours.}
	}
	\label{fig:7_prompt}
 \end{figure}

%% file: sections/4_related_work.tex
\section{Related Work}

\paragraph{Existing Benchmarks and Datasets.} Prior efforts fall short of supporting expert-level forensic analysis for academic images, due to three structural limitations: (1) \textbf{\textit{Generic-scene bias}}, where datasets such as Semi-Truths~\citep{NEURIPS2024_d5cdf7e5} and GenImage~\citep{DBLP:conf/nips/ZhuCYHLLT0H023} are dominated by faces or natural scenes rather than structured academic images; (2) \textbf{\textit{Restricted forgery strategies}}, as benchmarks including GRE~\citep{10.1145/3664647.3681445} and AIGuard~\citep{zhang-etal-2025-aiguard} largely focus on global or single-type edits and fail to model subtle academic manipulations; and (3) \textbf{\textit{Forensic-agnostic evaluation protocols}}, where datasets such as AIGIBench~\citep{li2025is} and DFBench~\citep{10.1145/3746027.3758204} emphasize detection accuracy while overlooking manipulation scope and localization.

\paragraph{Forensic Models: Expert Models, MLLMs, and Hybrid Approaches.} In parallel, the literature has explored a broad spectrum of forensic models for generative forgery analysis, spanning vision-only expert models, MLLMs, and hybrid systems.
Vision-only expert models cover multiple detection paradigms, including diffusion-process-based methods such as DIRE~\citep{Wang_2023_ICCV}, DRCT~\citep{pmlr-v235-chen24ay}, and denoising-trajectory-based methods~\citep{liang2025denoising}, as well as alignment-based approaches including AlignedForensics~\citep{rajan2025aligned} and DDA~\citep{chen2025dual}.
Meanwhile, MLLMs have exhibited strong cross-modal understanding and reasoning capabilities, exemplified by OpenAI o4-mini-high~\citep{o4mini}, Gemini 2.5 Flash~\citep{gemini2.5flashprpreview}, and the Qwen-VL series~\citep{qwen25vl}, motivating their evaluation for generative image forensics in complex scenarios. 
Hybrid approaches integrate MLLMs with expert models either directly, as in FakeVLM~\citep{wen2025spot}, or in an assisted manner, as exemplified by SIDA~\citep{Huang_2025_CVPR} and FakeShield~\citep{xu2025fakeshield}.

%% file: sections/5_conclusion.tex
\section{Conclusion}
We introduce \ours, a holistic benchmark comprising over 20k forensic questions, designed to evaluate vision-only models, MLLMs, and hybrid systems for academic image forensics. 
The primary contribution of \ours lies in revealing systematic forensic capability gaps that were not exposed by prior benchmarks. 
These gaps stem from the nature of academic images, which involve fine-grained details, knowledge-intensive content, and structurally complex layouts. 

Similar challenges are \textbf{not unique to academic imagery}. Domains such as legal evidence analysis (\eg{ forged visual exhibits}), financial document forensics, and other structured visual materials also require \textbf{fine-grained inspection}, \textbf{domain knowledge}, and \textbf{precise region-level reasoning}. These shared characteristics suggest that the capability gaps exposed by \ours are rooted in common challenges rather than being confined to academic images alone.
We hope \ours provides a robust foundation for advancing trustworthy, interpretable, and generalizable multimodal forensic systems.

%% file: sections/6_limitations.tex
\section*{Limitations}
\label{sec:limitations}



While \ours provides a comprehensive benchmark for evaluating AI-generated academic images, several limitations remain:

(1) The availability of \emph{real-world retracted papers involving AI-generated images} is limited. Only a small number of confirmed real cases are publicly accessible, which constrains the scale at which authentic misconduct examples can be incorporated into our benchmark. We expect to include more real-world cases as attention to generative image misuse in academic publishing grows.

(2) Baseline coverage is constrained by the availability of open-source implementations. Although we evaluate a broad range of state-of-the-art models, some recent approaches (\eg{ AIGI-Holmes}) lack released training weights. Future updates will incorporate additional baselines as more expert models become publicly accessible.

%% file: sections/7_ethical.tex
\section*{Ethical Considerations}
\label{sec: ethical}

\ours complies with ACL ethics guidelines. This study involved no human subjects or animal experimentation. All data were collected from open-source repositories in accordance with relevant usage licenses, ensuring that privacy is preserved and no personally identifiable information is included. \ours is released under the Creative Commons Attribution 4.0 International License (CC BY 4.0), and the associated codebase is distributed under the Apache License 2.0, supporting both commercial and open-source applications. We have made consistent efforts to minimize bias and ensure transparency throughout the dataset construction and evaluation process.

%% file: sections/acknowledgments.tex
\section*{Acknowledgments}
This work is supported by the National Natural Science Foundation of China (Grant Nos. 62473271, 62176026), the Fundamental Research Funds for the Beijing University of Posts and Telecommunications (Grant No. 2025AI4S03), and the BUPT Innovation and Entrepreneurship Support Program (Grant No. 2025-YC-A042). This work is also supported by the Engineering Research Center of Information Networks, Ministry of Education, China. We would also like to thank the anonymous reviewers and area chairs for constructive discussions and feedback.

%% file: appendix/0_Table_Content.tex
\newpage
\section*{Appendix Roadmap}

\paragraph{Appendix~\ref{sec:A_Dataset Annotation and Construction}.} This appendix provides a detailed description of the annotation and construction procedures underlying the \ours benchmark:
\begin{itemize} [leftmargin=2em]
    \item \textbf{Appendix~\ref{subsec:A0_Limitations of Including Real-World Retracted Cases}} details the collection process of real-world retracted cases and clarifies the practical constraints associated with using real-world misconduct cases.
    
    \item \textbf{Appendix~\ref{subsec:A0_Qualitative expert evaluation}} provides qualitative expert evaluations on the consistency between synthetic data and real-world misconduct cases.

    \item \textbf{Appendix~\ref{subsec:A1_Dataset Statistics}} summarizes key dataset statistics, including distribution across forgery simulations and annotated categories. 

    \item \textbf{Appendix~\ref{subsec:A2_Data Curation}} describes the data curation process. 

    \item \textbf{Appendix~\ref{subsec:A3_Synthetic Data Generation}} details the synthetic data generation pipeline, including the generative models employed and the design of representative forgery simulations. 
\end{itemize}

\paragraph{Appendix~\ref{sec:B_Experiment Details}.} This appendix provides the detailed experimental setup supporting \ours evaluation:  

\begin{itemize} [leftmargin=2em]
    \item \textbf{Appendix~\ref{subsec:B1_Experiment Environment}} describes experiment environment.

    \item \textbf{Appendix~\ref{subsec:B2_Benchmarked Models}} lists all benchmarked models, including expert models and MLLMs. 

    \item \textbf{Appendix~\ref{subsec:B3_Evaluation Metrics}} defines the evaluation metrics. 

    \item \textbf{Appendix~\ref{subsec:B4_Evaluation Protocol}} presents the evaluation procedures and task-specific protocols.
\end{itemize}

\paragraph{Appendix~\ref{sec:C_Multi-dimensional Evaluation Analysis}.}
This appendix provides supplementary results and analyses, including the effects of Few-Shot prompting, Chain-of-Thought prompting, and post-processing perturbations.

\begin{itemize} [leftmargin=2em]

    \item \textbf{Appendix~\ref{subsec:C2_Impact of Prompting Strategies}} analyzes the impact of prompting strategies.

    \item \textbf{Appendix~\ref{subsec:C3_Analysis of Post-Processing Results}} analyzes the impact of post-processing perturbations.

\end{itemize}

\paragraph{Appendix~\ref{sec:D_Case Study}.}
This appendix presents detailed case studies.

\paragraph{Appendix~\ref{sec:E_Use of AI Assistants}.}
This appendix provides details regarding the use of AI assistants.

%% file: appendix/A_dataset_annotation_and_construction.tex
\section{Dataset Annotation and Construction}
\label{sec:A_Dataset Annotation and Construction} 

\subsection{Limitations of Including Real-World Retracted Cases}
\label{subsec:A0_Limitations of Including Real-World Retracted Cases}

\paragraph{Collection of Real-World Misconduct Cases.} During dataset construction, we systematically examined publicly available academic misconduct resources, including Retraction Watch and PubPeer. Retraction Watch reports retracted scientific publications, while PubPeer is a post-publication discussion platform where researchers raise concerns about potential issues in published work. From these platforms, we identified three documented cases involving AI-generated academic images, including two formally retracted papers and one paper under public discussion.

\input{appendix/figures/F/real-world_case1}
\input{appendix/figures/F/real-world_case2}

Notably, as shown in Figures~\ref{fig:real-world case1} and~\ref{fig:real-world case2}, only a very limited number of cases can be confidently attributed to AI-generated academic images. This scarcity does not indicate that AI-generated image misuse is rare in practice, but rather reflects the inherent difficulty of detecting such forgeries in real-world scientific publishing.

\paragraph{Practical Constraints of Using Real-World Misconduct Cases.} Currently available and verifiable misconduct cases involving AI-generated academic images exhibit several limitations:

\begin{itemize} [leftmargin=*]
    \item \textbf{\textit{Lack of Structured Annotations.}} Most cases do not provide region- or pixel-level ground truth, which prevents reliable computation of localization metrics and undermines evaluation reproducibility.
    
    \item \textbf{\textit{Legal and Ethical Sensitivity.}} Some cases involve ongoing investigations or sensitive ethical issues, making them unsuitable for inclusion in a publicly released benchmark.
    
    \item \textbf{\textit{Lack of Editing Provenance.}} Real cases rarely disclose detailed editing procedures or generation sources (\eg generative models), making it impossible to control manipulation granularity or construct fully verified ground truth.

    \item \textbf{\textit{Insufficient Sample Size.}} The number of currently available cases is too limited to support statistically stable and systematic evaluation.
\end{itemize}

Under these conditions, incorporating real-world cases into a systematic benchmark would compromise controllability, reproducibility, and consistency in manipulation granularity. Therefore, \ours adopts a structured modeling framework grounded in forgery patterns identified by academic review experts, enabling controlled, fine-grained, and reproducible evaluation.

\subsection{Qualitative Expert Evaluation}
\label{subsec:A0_Qualitative expert evaluation}

To validate the consistency between synthetic samples and real-world misconduct cases, we conducted a blinded expert evaluation on 200 images randomly sampled from the combined pool of real-world misconduct cases and \ours synthetic samples. All available authentic cases were included, with the remaining images randomly drawn from the synthetic set. Five experts reviewed the images without knowing their origin. Each image was rated on a 5-point Likert scale (1 = very low, 5 = very high) based on the following criteria:

\begin{itemize} [leftmargin=*]
    \item \textbf{\textit{Visual Realism}} (how natural the image appears)
    \item \textbf{\textit{Structural Plausibility}} (whether the image structure is coherent and scientifically reasonable)
    \item \textbf{\textit{Consistency with Known Misconduct Patterns}} (whether the forgery resembles patterns observed in real-world retraction cases)
    \item \textbf{\textit{Overall Credibility}} (as an academic image)
\end{itemize}

\input{appendix/tables/A/0_consistency}

Table~\ref{tab:A0_consistency} shows that \ours synthetic samples received scores highly comparable to real-world retraction cases across all four criteria. These findings suggest that \textbf{the synthetic data reasonably aligns with real-world misconduct cases in terms of visual realism, structural plausibility, and manipulation characteristics}.

\subsection{Dataset Statistics}
\label{subsec:A1_Dataset Statistics}

As shown in Figure~\ref{fig:forgery_pie}, we summarize the data distribution of \ours, comprising real images (1,795 images) and four forgery strategies: Text Constraint Fabrication (3,121 images), Image Inference Forgery (2,274 images), Targeted Region Restoration (1,650 images), and Targeted Region Editing (1,165 images). 

\input{appendix/figures/A/forgery_pie}

As shown in Figure~\ref{fig:category_bar}, the seven academic image categories exhibit a deliberately diverse distribution, reflecting broad coverage of structured, dense, and heterogeneous scientific visual forms.

\input{appendix/figures/A/category_bar}

\subsection{Data Curation}
\label{subsec:A2_Data Curation}

\paragraph{Paper Parsing. }
\ours collected 4,362 high-quality academic papers from the open-access PMC repository and performed document-level parsing to construct the initial visual corpus. The paper selection criteria were as follows:
\begin{itemize} [leftmargin=*]
    \item \textbf{\textit{Structural Completeness.}} Each paper must contain at least four independent figures to ensure sufficient visual information density. We prioritized papers with multi-panel figures, where a single figure can be reasonably segmented into multiple panels to support subsequent panel-level forgery simulation.

    \item \textbf{\textit{Image Parsability and Resolution Requirements.}} Figures must maintain adequate clarity in the original publication. Papers were excluded if all images had a shortest side of 100 pixels or less, or exhibited severe scanning blur or other distortions that hindered reliable parsing.

    \item \textbf{\textit{Data Cleaning and Deduplication.}} We removed duplicated papers and excluded documents lacking structural value, such as those containing only decorative or non-informative images.
\end{itemize}

As illustrated in Figure~\ref{fig:paper_parsing}, a \emph{figure} refers to a complete figure unit paired one-to-one with its corresponding caption, whereas a \emph{panel} denotes the smallest indivisible visual unit within an academic image.
The entire extraction and parsing process required approximately 200 hours of human and computational effort.

The extraction and parsing pipeline consisted of the following steps:

\input{appendix/figures/A/paper_parsing}

\begin{itemize} [leftmargin=*]
    \item \textbf{\textit{Step 1: Figure and Caption Extraction.}} We employed the Fitz library to extract figures from PDF documents, as it consistently preserved higher visual fidelity than alternative document parsing tools, which was critical for subsequent panel-level analysis. 
    Captions were parsed using dots.ocr~\citep{li2025dotsocrmultilingualdocumentlayout} and were further used to cross-validate the correctness of the extracted figures.
    Finally, three human experts with academic peer-review experience manually verified the figure--caption correspondences to ensure extraction accuracy.

    \item \textbf{\textit{Step 2: Panel Segmentation from Figures.}} We applied YOLOv7~\citep{Wang_2023_CVPR} to segment figures into panels.
    The YOLOv7 model was fully trained on 200,000 manually annotated academic panels and had been validated to meet production-level reliability and commercial deployment standards. Panels with irregular layouts, poor image quality, or lacking experimental visual content were subsequently filtered out through joint inspection by three expert reviewers and five trained graduate annotators.
\end{itemize}

\subsection{Synthetic Data Generation}
\label{subsec:A3_Synthetic Data Generation}
As discussed in the limitations, real-world misconduct cases are inherently scarce and present multiple practical constraints for benchmark construction. Therefore, we adopted a synthetic data generation paradigm, which enabled scalable, controllable, and well-annotated construction of forensic samples. 
This subsection describes the synthetic data generation pipeline employed in \ours.

\subsubsection{Generative Models}
\label{subsubsec:A3_1_Generative Models}
As shown in Table~\ref{tab:file-model-counts}, we incorporate 25 state-of-the-art generative models spanning diffusion-based models, hybrid architectures, and one unified multimodal understanding and generation model to support diverse academic forgery simulations.

\subsubsection{Forgery Simulations}
\label{subsubsec:A3_2_Forgery Simulations}

To systematically simulate realistic manipulations in academic imagery, \ours adopted a strategy-driven forgery simulation framework. Inspired by common misconduct patterns in scholarly publications and aligned with the capabilities of contemporary generative models, we define four representative forgery strategies: \textit{Text Constraint Fabrication}, \textit{Image Inference Forgery}, \textit{Targeted Region Restoration}, and \textit{Targeted Region Editing}. These strategies cover global and localized manipulation scenarios, ranging from semantic-level content fabrication to fine-grained region-specific edits.

Specifically, the first two strategies focus on holistic image synthesis conditioned on textual or contextual constraints, yielding globally generated academic figures. For \textit{Text Constraint Fabrication}, we instructed GPT-4o mini to semantically reconstruct the captions of authentic academic images.

In contrast, the latter two strategies involve localized manipulations applied to selected regions within otherwise authentic images. For these region-level forgeries, we employed the Segment Anything Model (SAM)~\citep{Kirillov_2023_ICCV} to automatically generate precise manipulation masks, ensuring accurate spatial control over the forged regions and enabling consistent pixel-level ground-truth annotations for localization evaluation.

\subsubsection{Data Quality Assurance}
\label{subsubsec:A3_3_Data Quality Assurance}

At the local level, annotators examined whether manipulated regions exhibited coherent textures, well-defined boundaries, and semantic consistency with surrounding content, while avoiding trivial artifacts such as unnatural edges, repetitive patterns, or mismatched visual semantics. At the global level, images were assessed for overall structural coherence, adherence to scientific conventions, and consistency with the corresponding academic context, including figure layout, visual hierarchy, and modality-specific characteristics. The review process involved five human experts with prior academic peer-review experience, together with three Ph.D. students and six trained Master's students. 

In addition to expert review, we conducted an automated quality assessment to contextualize the realism and semantic fidelity of the synthesized images relative to datasets used by baseline forensic models. As shown in Table~\ref{tab:quality}, we compared the quality scores of \ours against baseline datasets using three metrics: Inception Score (IS), Fréchet Inception Distance (FID), and CLIP Score.

\input{appendix/tables/A/5_quality}

\setlist[itemize]{leftmargin=*, itemsep=0.05em, topsep=0.1em}
\begin{itemize}
    \item \textbf{\textit{Inception Score (IS).}}  IS measures visual quality and content diversity of generated images based on the entropy of class predictions produced by a pretrained Inception network. Higher scores generally indicate images that are both visually coherent and diverse. We note that \textbf{IS was originally proposed for evaluating GAN-based image generators, and its sensitivity may be reduced for modern non-GAN architectures} (\eg diffusion-based or hybrid generative models), whose outputs often depart from the ImageNet-centric class distributions assumed by the Inception model. As a result, lower IS values in our setting should not be interpreted as diminished perceptual quality, but rather as a known limitation of the metric when applied beyond its original design scope.

    \item \textbf{\textit{Fréchet Inception Distance (FID).}}  
    FID~\citep{Seitzer2020FID} evaluates the distributional similarity between generated images and real academic images by computing the Fréchet distance between their feature embeddings extracted from a pretrained Inception model. Lower FID scores indicate higher visual fidelity and closer alignment with real-image statistics.

    \item \textbf{\textit{CLIP Score.}}  
    CLIP Score~\citep{hessel-etal-2021-clipscore} measures cross-modal semantic consistency between generated images and their corresponding textual prompts. It is computed as the cosine similarity between visual and textual embeddings extracted by a pretrained CLIP model, where higher scores reflect stronger semantic alignment.
\end{itemize}

Together, these complementary metrics provide an objective reference for evaluating perceptual fidelity, distributional realism, and semantic alignment, supporting the validity of the curated dataset alongside expert-driven quality control.

%% file: appendix/figures/F/real-world_case1.tex
\begin{figure}[!b]
	\centering
	\includegraphics[width=0.90\columnwidth]{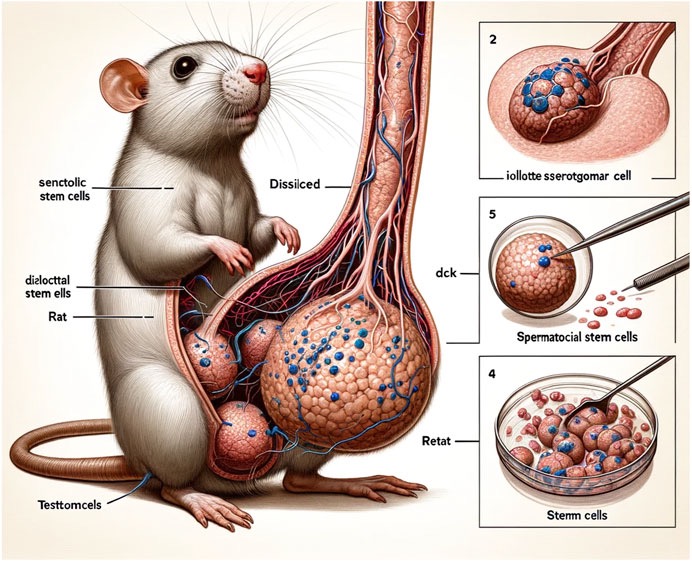}
	\caption{Example of a retracted paper published in \textit{Frontiers in Cell and Developmental Biology} that employed an AI-generated illustrative image.}
	\label{fig:real-world case1}
 \end{figure}
 

%% file: appendix/figures/F/real-world_case2.tex
\begin{figure}[!b]
	\centering
	\includegraphics[width=0.80\columnwidth]{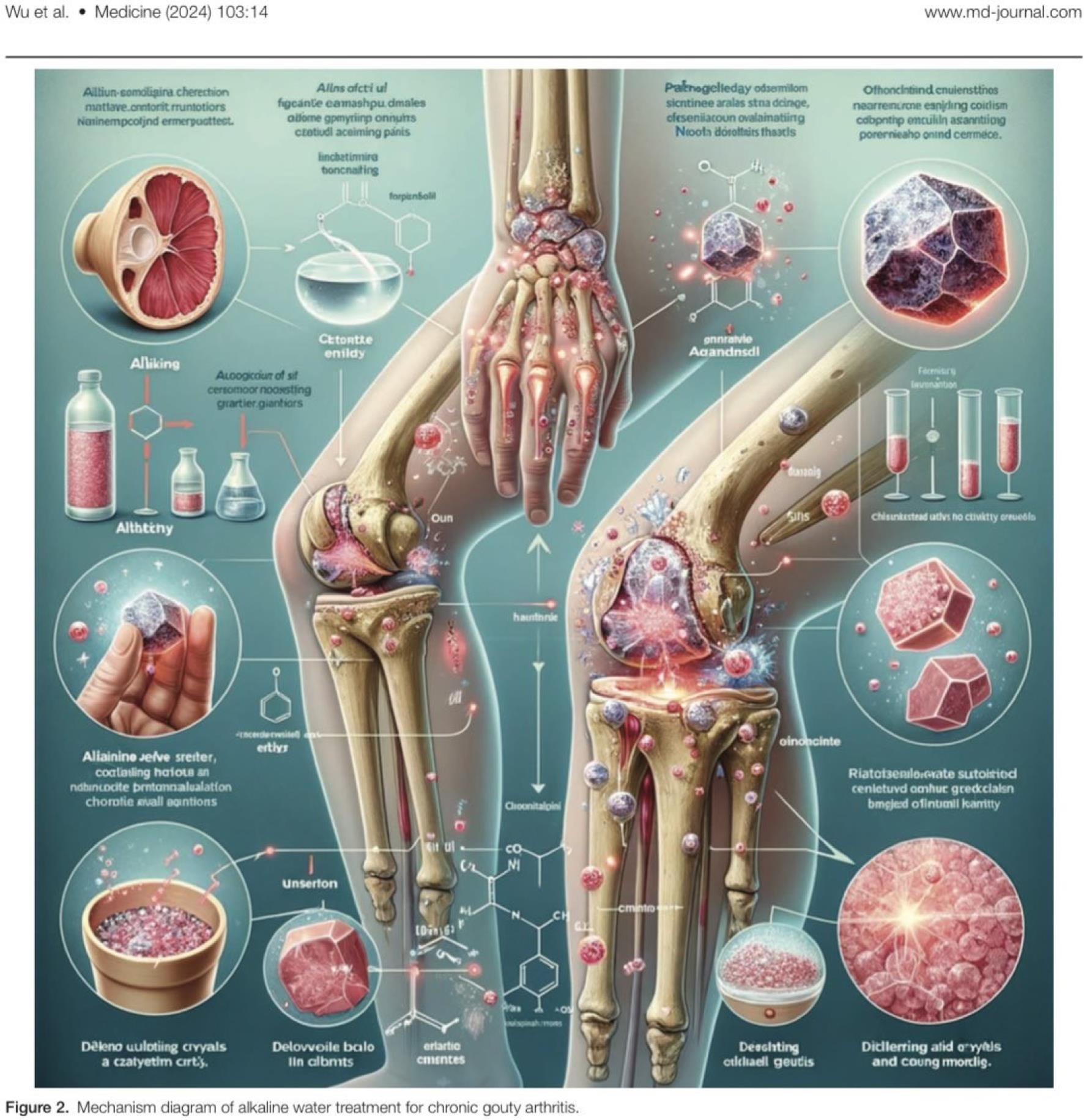}
	\caption{Example of a retracted paper from the Lippincott journal \textit{Medicine}.}
	\label{fig:real-world case2}
\end{figure}

%% file: appendix/tables/A/0_consistency.tex
\begin{table}[!t]
\centering
\small
\begin{tabular}{lcc}
\toprule
\textbf{Criterion} & \textbf{Real-World} & \textbf{Synthetic} \\
\midrule
\makecell[l]{Visual Realism $\uparrow$} 
& 3.86 $\pm$ 0.52 & 4.21 $\pm$ 0.56 \\

\makecell[l]{Structural Plausibility $\uparrow$} 
& 3.98 $\pm$ 0.47 & 4.09 $\pm$ 0.50 \\

\makecell[l]{Consistency with \\ Misconduct Patterns $\uparrow$} 
& 4.18 $\pm$ 0.60 & 4.15 $\pm$ 0.63 \\

\makecell[l]{Overall Credibility $\uparrow$} 
& 4.05 $\pm$ 0.54 & 4.19 $\pm$ 0.58 \\
\bottomrule
\end{tabular}
\caption{\textbf{Qualitative expert evaluation.} Comparison between \textit{real-world misconduct cases} and \textit{synthetic samples}. Higher scores indicate better performance.}
\label{tab:A0_consistency}
\end{table}

%% file: appendix/figures/A/forgery_pie.tex
\begin{figure}[b]
	\centering
	\includegraphics[width=0.98\columnwidth]{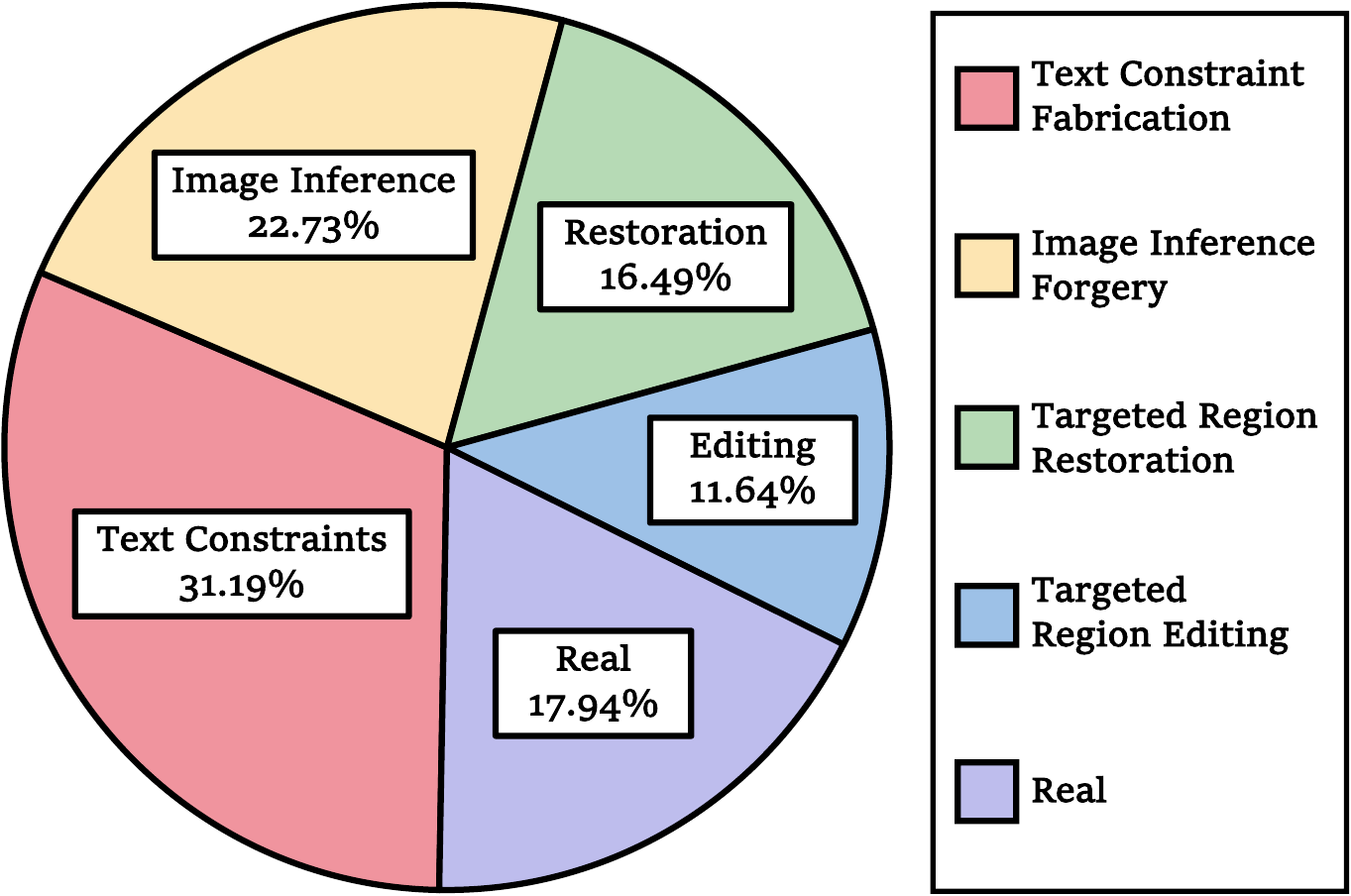}
	\caption{\textbf{Distribution of real images and images generated by four forgery strategies in \ours.}}
	\label{fig:forgery_pie}
 \end{figure}

%% file: appendix/figures/A/category_bar.tex
\begin{figure}[t]
	\centering
	\includegraphics[width=0.98\columnwidth]{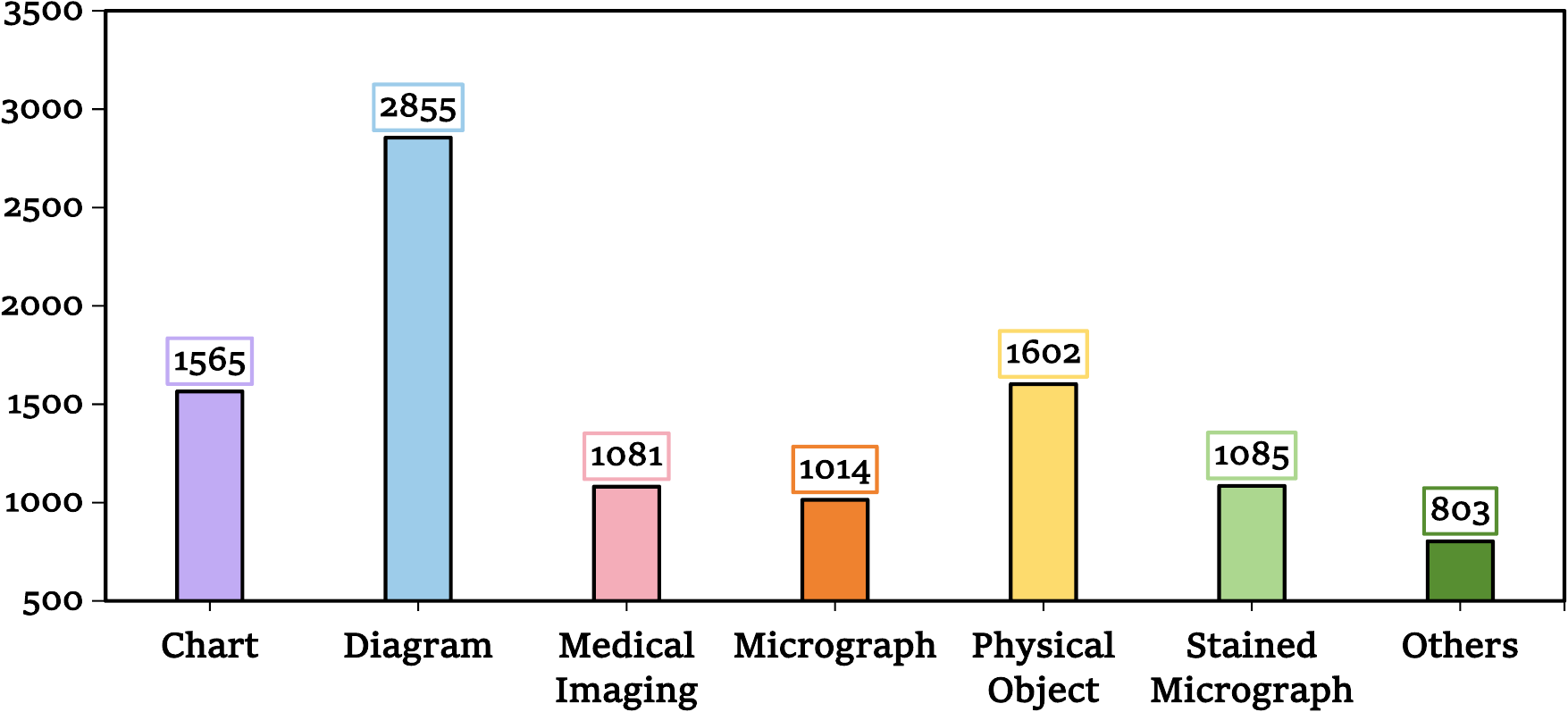}
	\caption{\textbf{Distribution of seven academic image categories in \ours.}}
	\label{fig:category_bar}
\end{figure}

%% file: appendix/figures/A/paper_parsing.tex
\begin{figure}[t]
	\centering
	\includegraphics[width=0.98\columnwidth]{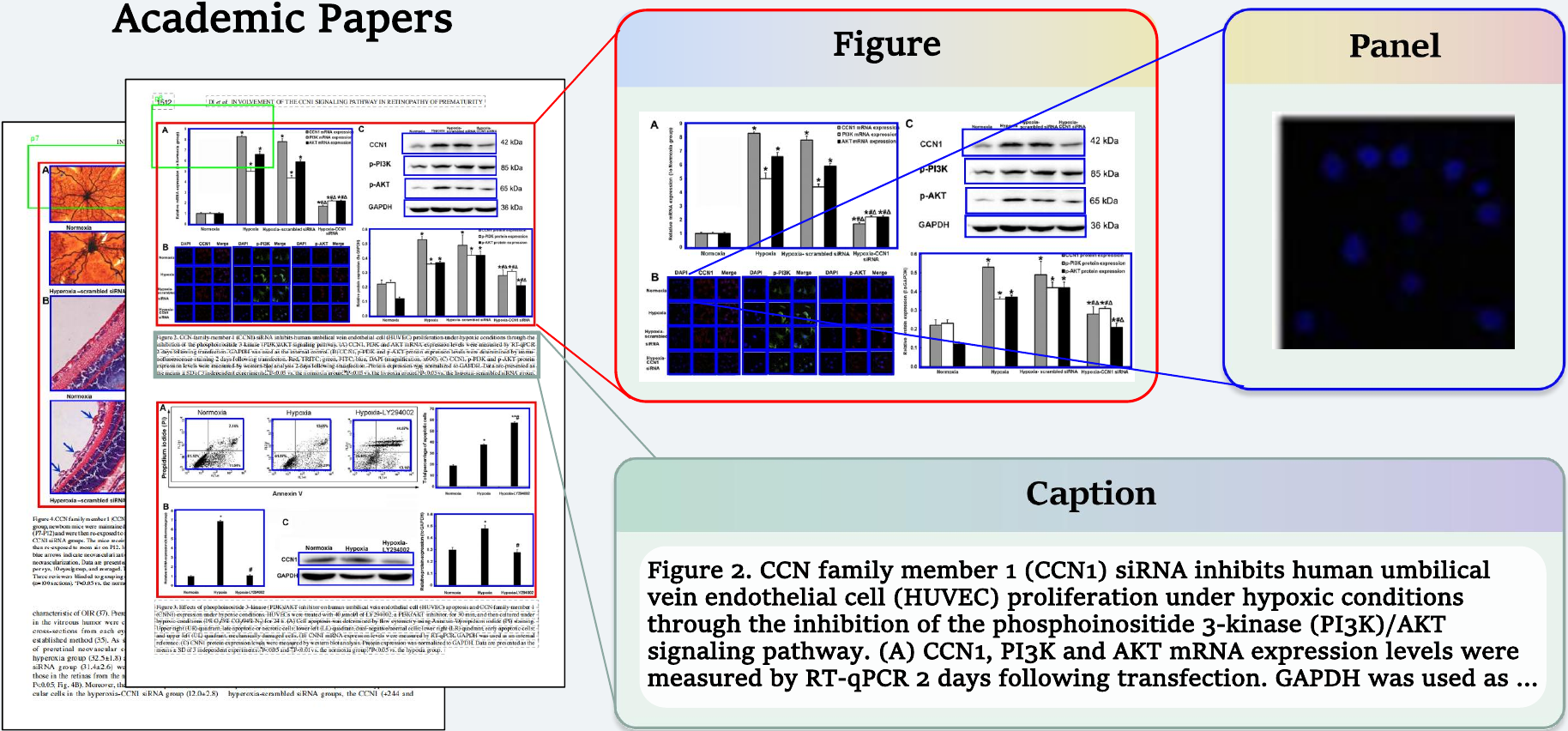}
	\caption{\textbf{Extraction and Parsing of academic papers.} The regions highlighted with \textcolor{red}{red boxes} correspond to figures, while those highlighted with \textcolor{blue}{blue boxes} correspond to panels. \textcolor{green}{Green boxes} are used to verify whether the visualization is effective.
	}
	\label{fig:paper_parsing}
 \end{figure}

%% file: appendix/tables/A/5_quality.tex
\begin{table}[t]
\centering
\small
\setlength{\tabcolsep}{4pt}
\begin{tabular}{lccc}
\toprule
\multirow{2}{*}{\textbf{Dataset}} & \multicolumn{3}{c}{\textbf{Score}} \\
\cmidrule(lr){2-4}
& \textbf{CLIP} $\uparrow$ & \textbf{FID} $\downarrow$ & \textbf{IS} $\uparrow$ \\
\midrule
DiffusionForensics & -- & 27.46 & 13.47 $\pm$ 0.97 \\
DRCT-2M & 75.74 & 22.97 & 34.57 $\pm$ 1.57 \\
SID-Set (Full\_Synthetic) & \multirow{2}{*}{--} & 60.02 & 38.73 $\pm$ 0.95 \\
SID-Set (Tampered) &  & 30.20 & 32.28 $\pm$ 2.39 \\
New Generator & -- & 75.58 & 24.02 $\pm$ 1.35 \\
Robust\_LDM & -- & 99.80 & 13.76 $\pm$ 1.02 \\
FakeClue & 70.95 & 86.59 & 13.30 $\pm$ 1.69 \\
EvalGEN & -- & 69.60 & 29.67 $\pm$ 0.67 \\
Chameleon & -- & 53.88 & 16.87 $\pm$ 0.46 \\
\midrule
\textbf{Ours} & \textbf{78.66} & \textbf{32.22} & \textbf{13.41 $\pm$ 1.23} \\
\bottomrule
\end{tabular}
\caption{\textbf{Comparison of different datasets.}}
\label{tab:quality}
\end{table}

%% file: appendix/B_experiment_details.tex
\section{Experiment Details}
\label{sec:B_Experiment Details}

\subsection{Experiment Environment}
\label{subsec:B1_Experiment Environment}
For evaluation experiments, most model inferences were conducted via the OpenRouter API\footnote{\url{https://openrouter.ai}}. Exceptions include the LLaVA series, which were downloaded from Hugging Face\footnote{\url{https://huggingface.co}} and executed locally, as well as the Doubao series, which were accessed via the Volcano Engine API\footnote{\url{https://www.volcengine.com}}.

The system configuration is summarized below:

\begin{itemize}[leftmargin=*]
\item \textbf{CPU}: Dual-socket Intel Xeon Gold 6148 (2.40 GHz), 20 cores per socket, 80 threads total
\item \textbf{GPU}: 8 $\times$ NVIDIA A40 (48 GB VRAM each)
\item \textbf{GPU Driver}: 575.57.08
\item \textbf{CUDA}: 11.8
\item \textbf{cuDNN}: 8.9.6 (compiled with CUDA 11.8)
\item \textbf{Operating System}: Ubuntu 22.10
\end{itemize}

\subsection{Benchmarked Models}
\label{subsec:B2_Benchmarked Models}

As summarized in Table~\ref{tab:expert_models}, we evaluate nine expert models in total, including six vision-only expert models and three hybrid expert models.

As shown in Table~\ref{tab:mllms}, we evaluate 25 MLLMs (consisting of 14 proprietary and 11 open-source models) as well as one unified multimodal understanding and generation model. The proprietary models are released by OpenAI~\citep{GPT-4.1,GPT-5.1,o4mini}, Google~\citep{gemini2.5flashprpreview,google2025gemini3pro}, Alibaba~\citep{bai2025qwen3vltechnicalreport}, and ByteDance~\citep{doubao1.5thinkingvisionpro,doubao1.5thinkingpro,doubaoseed1.6flash,doubaoseed1.6thinking}. The open-source models are released by Meta~\citep{llama4maverick}, Mistral, Alibaba~\citep{qwen25vl}, Google~\citep{gemmateam2025gemma3technicalreport}, and the LLaVA community~\citep{li2024llavaonevisioneasyvisualtask,li2024llavanextinterleavetacklingmultiimagevideo,liu2024llavanext}.

\subsection{Evaluation Metrics}
\label{subsec:B3_Evaluation Metrics}
To comprehensively assess model performance across different dimensions, we design \textbf{seven core metrics}. 

\paragraph{(1) Accuracy (ACC).}
ACC measures the proportion of correctly predicted answers among all evaluated instances. All tasks evaluated with ACC are formulated as \emph{single-choice classification problems}, where each image corresponds to exactly one ground-truth label and the model selects a single option.

Let $N$ denote the total number of valid evaluation samples, and let $\mathbb{I}(\cdot)$ be the indicator function. ACC is computed as:
\begin{equation}
\mathrm{ACC} = \frac{1}{N} \sum_{i=1}^{N} \mathbb{I}\bigl(\hat{y}_i = y_i\bigr),
\end{equation}
where $\hat{y}_i$ and $y_i$ denote the predicted and ground-truth labels of the $i$-th sample, respectively.

For the Forgery Scope Discrimination (FSD) task, ACC is reported under two settings depending on model capability. Models that support fine-grained scope prediction are evaluated under a multi-class setting (\textit{Real}, \textit{Entire Forgery}, \textit{Partial Forgery}). Vision-only expert models that only support authenticity detection are evaluated under a binary setting (\textit{Real}, \textit{Forgery}), where both global and local forgeries are mapped to the \textit{Forgery} class.

\paragraph{(2) Macro-F1 Score (M-F1).}
M-F1 is adopted to measure class-balanced performance by averaging F1 scores across classes with equal weight. Similar to ACC, M-F1 is computed over single-choice classification outputs and mitigates the influence of class imbalance.

For each class $c \in \mathcal{C}$, precision $P_c$, recall $R_c$, and F1 Score $F1_c$ are defined as:
\begin{equation}
P_c = \frac{\mathrm{TP}_c}{\mathrm{TP}_c + \mathrm{FP}_c}, \quad
R_c = \frac{\mathrm{TP}_c}{\mathrm{TP}_c + \mathrm{FN}_c},
\end{equation}
\begin{equation}
F1_c = \frac{2 \cdot P_c \cdot R_c}{P_c + R_c},
\end{equation}
where $\mathrm{TP}_c$, $\mathrm{FP}_c$, and $\mathrm{FN}_c$ denote the numbers of true positives, false positives, and false negatives for class $c$, respectively.

M-F1 is then computed as:
\begin{equation}
\mathrm{M\text{-}F1} = \frac{1}{|\mathcal{C}|} \sum_{c \in \mathcal{C}} F1_c.
\end{equation}

For the FSD task, M-F1 is computed separately for the binary and multi-class settings, consistent with the corresponding ACC evaluation protocol. Binary M-F1 is reported for expert models distinguishing \textit{Real} versus \textit{Forgery}, while multi-class M-F1 is reported for models predicting \textit{Real}, \textit{Entire Forgery}, and \textit{Partial Forgery}. Samples with abstention responses (\ie{ \textit{Not Sure}}) are excluded from M-F1 computation.

\paragraph{(3) Correct Localization Accuracy (CLA).} CLA is a region-level localization metric that evaluates whether all ground-truth tampered regions are sufficiently covered by the predicted regions. CLA is computed at the \emph{sample level} and then aggregated across samples.

For a given sample, let $N_{\mathrm{gt}}$ denote the number of ground-truth connected components, and let $\mathrm{GT}_i$ denote the $i$-th ground-truth component. Let $\mathrm{Pred}$ represent the union of all predicted regions. A sample is considered \emph{CLA-correct} if and only if every ground-truth component is sufficiently covered by the predicted regions:
\begin{equation}
\resizebox{0.95\hsize}{!}{$
\mathrm{CLA}_{\mathrm{correct}} =
\begin{cases}
\text{False}, & \text{if } N_{\mathrm{gt}} = 0, \\
\text{True},  & \text{if } \forall i \in [1, N_{\mathrm{gt}}], 
\displaystyle \frac{|\mathrm{Pred} \cap \mathrm{GT}_i|}{|\mathrm{GT}_i|} \ge \tau, \\
\text{False}, & \text{otherwise},
\end{cases}
$}
\end{equation}
where $|\cdot|$ denotes the number of pixels in a region, and $\tau$ is a coverage threshold (set to $0.5$ in all experiments).

The final CLA score is reported as the percentage of CLA-correct samples over all evaluated samples:
\begin{equation}
\mathrm{CLA}\,(\%) = 
\frac{\#\{\text{CLA-correct samples}\}}{\#\{\text{total samples}\}} \times 100\%.
\end{equation}

\paragraph{(4) Over Localization Rate (OLR).}
OLR measures whether a model produces excessively many predicted regions relative to the ground-truth, capturing the tendency of over-fragmented or overly sensitive localization.

For a given sample, let $N_{\mathrm{gt}}$ denote the number of ground-truth connected components and $N_{\mathrm{pred}}$ denote the number of predicted regions. A sample is considered \emph{over-localized} if:
\begin{equation}
\resizebox{0.95\hsize}{!}{$
\mathrm{OLR}_{\mathrm{over}} =
\begin{cases}
\text{True}, & \text{if } N_{\mathrm{gt}} = 0 \ \text{and} \ N_{\mathrm{pred}} > 0, \\
\text{True}, & \text{if } N_{\mathrm{gt}} > 0 \ \text{and} \ 
N_{\mathrm{pred}} > N_{\mathrm{gt}} \times (1 + \alpha), \\
\text{False}, & \text{otherwise},
\end{cases}
$}
\end{equation}
where $\alpha$ is an over-localization tolerance factor (set to $0.5$ by default).

The final OLR score is computed as the proportion of over-localized samples among all evaluated samples:
\begin{equation}
\resizebox{0.95\hsize}{!}{$
\mathrm{OLR}\,(\%) =
\frac{\#\{\text{over-localized samples}\}}{\#\{\text{total samples}\}} \times 100\%.
$}
\end{equation}

\paragraph{(5) Intersection over Union (IoU).}
For pixel-level localization tasks, we adopt IoU to measure the spatial overlap between predicted tampering masks and ground-truth masks. IoU evaluates localization accuracy at the pixel granularity and is widely used for segmentation-based forensic analysis.

Let $\mathrm{Pred}$ and $\mathrm{GT}$ denote the predicted binary mask and the ground-truth binary mask for a given sample, respectively. IoU is defined as:
\begin{equation}
\mathrm{IoU} = \frac{|\mathrm{Pred} \cap \mathrm{GT}|}{|\mathrm{Pred} \cup \mathrm{GT}|},
\end{equation}
where $|\cdot|$ denotes the number of pixels in the corresponding region.

The final IoU score is reported as the average IoU over all evaluated samples.

\paragraph{(6) Pixel-level F1 Score (F1).}
To complement IoU, we further report the pixel-level F1 Score, which balances precision and recall of predicted tampering masks. This metric emphasizes accurate boundary delineation and penalizes both over- and under-segmentation.

At the pixel level, true positives (TP), false positives (FP), and false negatives (FN) are defined based on pixel-wise correspondence between the predicted mask and the ground-truth mask. Precision $P$, recall $R$, and F1 Score are computed as:
\begin{equation}
P = \frac{\mathrm{TP}}{\mathrm{TP} + \mathrm{FP}}, \quad
R = \frac{\mathrm{TP}}{\mathrm{TP} + \mathrm{FN}},
\end{equation}
\begin{equation}
\mathrm{F1} = \frac{2 \cdot P \cdot R}{P + R}.
\end{equation}

The reported pixel-level F1 Score is averaged across all samples in the evaluation set.

\paragraph{(7) Normalized Forensic Index (NFI).}
Beyond task-specific metrics, we define a composite score to holistically measure a model's overall forensic capability with an emphasis on balanced performance.
NFI is designed to reward models that perform consistently across heterogeneous forensic tasks, while penalizing pathological localization behaviors such as excessive over-localization.

For each model $i$, we first collect its core forensic scores:
\begin{equation}
\mathcal{S}_i = \{ s_{i,1}, s_{i,2}, s_{i,3}, s_{i,4} \},
\end{equation}
where $\mathcal{S}_i$ includes the FSD-F1, TAR-F1, MC-F1, and CLA scores of model $i$.
All scores are normalized to $[0,1]$ for numerical consistency.

We compute the harmonic mean to emphasize balanced performance:
\begin{equation}
\mathrm{HM}_i = \frac{|\mathcal{S}_i|}{\sum_{s \in \mathcal{S}_i} \frac{1}{s + \epsilon}},
\end{equation}
where $\epsilon$ is a small constant to ensure numerical stability.

To penalize over-localized predictions, we introduce an over-localization factor based on the OLR:
\begin{equation}
P_i = \left(1 - \mathrm{OLR}_i \right)^{\gamma},
\end{equation}
where $\mathrm{OLR}_i \in [0,1]$ and $\gamma$ controls the penalty strength.

Finally, the NFI is defined as:
\begin{equation}
\mathrm{NFI}_i = 100 \cdot \mathrm{HM}_i \cdot P_i,
\quad \gamma = 0.5.
\end{equation}

A higher NFI indicates stronger and more reliable forensic capability, reflecting both balanced task performance and structurally reasonable localization behavior.

\paragraph{Sensitivity Analysis of NFI.}
To examine the robustness of the proposed Normalized Forensic Index (NFI), we conducted a sensitivity analysis with respect to the OLR penalty exponent $\gamma$, as reported in Table~\ref{tab:nfi_gamma_sens}.

Table~\ref{tab:nfi_gamma_sens} shows that varying $\gamma$ within a reasonable range ($\gamma \in \{0.3, 0.5, 0.7\}$) leads to only minor changes in absolute NFI values, while the relative ranking of top-performing models remains largely stable.
In particular, leading proprietary MLLMs consistently occupy the top ranks, indicating that NFI is not overly sensitive to the exact strength of the OLR-based penalty.

\subsection{Evaluation Protocol}
\label{subsec:B4_Evaluation Protocol}
We uniformly adopted \textbf{high-resolution PNG images} for all evaluations to avoid compression artifacts introduced by lossy formats (\eg{ JPEG}), which may interfere with forensic cues and bias model judgments. When an input image exceeded a model's maximum input resolution, a two-stage adaptive compression strategy was applied: (1) the long edge was first scaled to the model-specific maximum input size while preserving the aspect ratio; and (2) the total pixel count was further reduced, if necessary, through iterative 10\% downscaling until the model's input constraints were satisfied.

All prompts were intentionally kept minimal, containing only task definitions and direct instructions, in order to reduce extraneous noise and better expose the intrinsic forensic capabilities of the evaluated models.

%% file: appendix/C_multi-dimensional_evaluation_analysis.tex
\section{Multi-Dimensional Evaluation Analysis}
\label{sec:C_Multi-dimensional Evaluation Analysis}

\subsection{Impact of Prompting Strategies}
\label{subsec:C2_Impact of Prompting Strategies}

\subsubsection{Impact of Few-Shot Prompting}
\label{subsubsec:C21_Impact of Few-Shot Prompting}

For Few-Shot prompting, we built an indexed reference library of curated examples. Given an input image, retrieval was restricted to the same category based on dataset annotations. Visual similarity was then computed using DINO3~\citep{simeoni2025dinov3} embeddings, and the single most similar reference image was selected as the in-context exemplar. The user prompt included the image description, the input image, and an output-format example to guide the model's inference.

Tables~\ref{tab:appendix_fewshot_fsd}--\ref{tab:appendix_fewshot_mc} reveal a task-dependent impact of Few-Shot prompting.

\subsubsection{Impact of Chain-of-Thought Prompting}
\label{subsubsec:C22_Impact of Chain-of-Thought Prompting}

For Chain-of-Thought (CoT) prompting, we tailored task-specific prompts for each of the seven categories across all four tasks. The corresponding experimental results for GPT-5.1 are summarized in Table~\ref{tab:experiment_CoT}.

Overall, CoT prompting yields broad gains for the MLLM across categories, improving detection and localization performance on all except Manipulation Classification (MC). Averaged over seven categories, CoT increases Forgery Scope Discrimination (FSD) accuracy (+4.38\%) and improves Textual Artifact Recognition (TAR) (+3.33\%), while boosting Tampering Pinpointing (TP) localization quality (+4.25\%), albeit with a slight rise in over-localization (+1.07\%). These trends indicate that explicit reasoning steps mainly help the model surface more reliable authenticity cues and provide more correct localization hits, but may also encourage slightly more aggressive region proposals.

\subsection{Analysis of Post-Processing Results}
\label{subsec:C3_Analysis of Post-Processing Results}
As shown in Table~\ref{tab:post_process}, to examine the robustness of models under common post-processing perturbations, we applied three representative image post-processing strategies.
Specifically, we considered \textbf{Gaussian blurring}, \textbf{JPEG compression}, and \textbf{image scaling}.
Gaussian blurring was implemented with a fixed blur radius of $5$ to suppress high-frequency details.
JPEG compression was applied with a quality factor of $50$, introducing moderate compression artifacts.
For image scaling, inputs were uniformly downsampled by a factor of $0.5$ using the LANCZOS resampling method to preserve interpolation quality.

%% file: appendix/D_case_study.tex
\section{Case Study}
\label{sec:D_Case Study}

As shown in Figures~\ref{fig:appendix_case_FSD_correct}--\ref{fig:appendix_case_TP_error1}, we provide 12 representative cases selected from the outputs of GPT-5.1. These cases span all seven academic image categories
(Chart, Diagram, Micrograph, Stained Micrograph, Physical Object, Medical Imaging, and Others), and are systematically organized across all forensic dimensions.

%% file: appendix/E_use_of_AI.tex
\section{Use of AI Assistants}
\label{sec:E_Use of AI Assistants}

This research was driven entirely by the authors, who provided all core scientific insights, experimental designs, and analyses. We acknowledge the use of AI assistants during the preparation of this manuscript: Cursor was utilized to aid in code writing and data processing, large language models (LLMs) assisted with language editing to enhance readability, and generative models played a role in the construction of our synthetic data (Table~\ref{tab:file-model-counts}). Furthermore, \texttt{GPT-4o mini} was employed to semantically reconstruct the captions of authentic academic images in \textit{Text Constraint Fabrication}, with detailed methodologies provided in Appendix~\ref{subsubsec:A3_2_Forgery Simulations}. We emphasize that all AI-assisted content was thoroughly reviewed and validated by the authors, who bear full responsibility for the scientific integrity and fundamental contributions of this work.

%% file: appendix/N_page.tex
\clearpage
\input{appendix/tables/A/2_forgery_strategies}

\clearpage
\input{appendix/tables/B/1_expert}
\input{appendix/tables/B/2_mllms}

\clearpage
\input{appendix/tables/B/3_nfi_gamma_sens}

\clearpage
\input{appendix/tables/C/2_fewshot_FSD}
\input{appendix/tables/C/2_fewshot_TAR}
\input{appendix/tables/C/2_fewshot_MC}

\clearpage
\input{appendix/tables/C/3_cot}

\input{appendix/tables/C/4_post_progress}

\clearpage
\input{appendix/figures/case/FSD_correct}
\clearpage
\input{appendix/figures/case/FSD_error1}
\clearpage
\input{appendix/figures/case/FSD_error2}
\clearpage
\input{appendix/figures/case/FSD_error3}
\clearpage
\input{appendix/figures/case/FSD_error4}

\clearpage
\input{appendix/figures/case/TAR_correct}
\clearpage
\input{appendix/figures/case/TAR_error}

\clearpage
\input{appendix/figures/case/MC_correct}
\clearpage
\input{appendix/figures/case/MC_error1}
\clearpage
\input{appendix/figures/case/MC_error2}

\clearpage
\input{appendix/figures/case/TP_correct}
\clearpage
\input{appendix/figures/case/TP_error1}

%% file: appendix/tables/A/2_forgery_strategies.tex
\begin{table*}[t]
\scriptsize
\centering
\small
\setlength{\tabcolsep}{2pt} 

\begin{tabularx}{\textwidth}{l r X}
\toprule
\textbf{Model} & \textbf{Count} & \textbf{Access Links} \\
\midrule
\rowcolor{red!10}
\multicolumn{3}{l}{\textbf{\textit{Text Constraint Fabrication}}} \\
\midrule

Janus-Pro-7B & 364 & \url{https://huggingface.co/deepseek-ai/Janus-Pro-7B} \\
DALL·E 3 & 273 & \url{https://openai.com/zh-Hans-CN/index/dall-e-3/} \\
FLUX.1 Dev & 273 & \url{https://huggingface.co/black-forest-labs/FLUX.1-dev} \\
FLUX.2 Pro & 70 & \url{https://openrouter.ai/black-forest-labs/flux.2-pro} \\
Midjourney V6 & 249 & \url{https://www.midjourney.com} \\
Midjourney V7 & 247 & \url{https://www.midjourney.com} \\
Stable Diffusion 3.5 Large Turbo & 248 & \url{https://huggingface.co/stabilityai/stable-diffusion-3.5-large-turbo} \\
Imagen 4 & 128 & \url{https://deepmind.google/models/imagen} \\
Nano Banana Pro & 71 & \url{https://deepmind.google/models/gemini-image/pro/} \\
kling-v2 & 190 & \url{https://klingai.com/} \\
CogView-4 & 183 & \url{https://docs.z.ai/guides/image/cogview-4} \\
Jimeng 3.0 & 145 & \url{https://jimeng.jianying.com/} \\
Seedream 3.0 & 128 & \url{https://console.volcengine.com/ark/region:ark+cn-beijing/model/detail?Id=doubao-seedream-3-0-t2i} \\
Wan2.1-T2I-Turbo & 191 & \url{https://bailian.console.aliyun.com/cn-beijing?tab=model\#/model-market/detail/wanx2.1-t2i-turbo?serviceSite=asia-pacific-china} \\
Wan2.5-T2I-Preview & 53 & \url{https://bailian.console.aliyun.com/cn-beijing?tab=model\#/model-market/detail/wan2.5-t2i-preview?serviceSite=asia-pacific-china} \\
Qwen-Image-Plus & 35 & \url{https://bailian.console.aliyun.com/cn-beijing?tab=model\#/model-market/detail/qwen-image-plus?serviceSite=asia-pacific-china} \\
SenseNova v6 MiaoHua & 273 & \url{https://www.sensenova.cn/en} \\

\midrule
\rowcolor{yellow!10}
\multicolumn{3}{l}{\textbf{\textit{Image Inference Forgery}}} \\
\midrule

DALL·E 2 & 292 & \url{https://openai.com/zh-Hans-CN/index/dall-e-2/} \\
FLUX.1 Schnell & 482 & \url{https://huggingface.co/black-forest-labs/FLUX.1-schnell} \\
Stable Diffusion 3.5 Medium & 462 & \url{https://huggingface.co/stabilityai/stable-diffusion-3.5-medium} \\
kling-v1 & 253 & \url{https://klingai.com/} \\
Seedream 4.5 & 69 & \url{https://console.volcengine.com/ark/region:ark+cn-beijing/model/detail?Id=doubao-seedream-4-5} \\
SenseNova v6 MiaoHua & 506 & \url{https://www.sensenova.cn/en} \\
Hunyuan 2.0 & 210 & \url{https://aistudio.tencent.com/modelSquare/home/list?modelKey=Multimodal} \\

\midrule
\rowcolor{green!10}
\multicolumn{3}{l}{\textbf{\textit{Targeted Region Restoration}}} \\
\midrule

RePaint & 1,650 & \url{https://github.com/andreas128/RePaint} \\

\midrule
\rowcolor{blue!10}
\multicolumn{3}{l}{\textbf{\textit{Targeted Region Editing}}} \\
\midrule

SenseNova v6 MiaoHua v2 & 701 & \url{https://www.sensenova.cn/en} \\
Stable Diffusion & 325 & \url{https://platform.stability.ai/docs/api-reference\#tag/Edit/paths/~1v2beta~1stable-image~1edit~1inpaint/post} \\
GPT Image 1 & 139 & \url{https://developers.openai.com/api/docs/models/gpt-image-1} \\

\bottomrule
\end{tabularx}

\caption{\textbf{Generative models used for each forgery strategy.}}
\label{tab:file-model-counts}
\end{table*}

%% file: appendix/tables/B/1_expert.tex
\begin{table*}[t]
\small
\centering
\renewcommand{\arraystretch}{1.2} 
\begin{tabularx}{\linewidth}{lcX}
\toprule
\textbf{Model} & \textbf{Capabilities} & \textbf{Access Links} \\
\midrule

\rowcolor{gray!10}
\multicolumn{3}{l}{\textit{\textbf{Hybrid MLLM-Assisted Expert Models}}} \\
\midrule
FakeShield & Detection + Localization & \url{https://github.com/zhipeixu/FakeShield} \\
SIDA & Detection + Localization & \url{https://github.com/hzlsaber/SIDA} \\
FakeVLM & Detection & \url{https://github.com/opendatalab/FakeVLM} \\
\midrule

\rowcolor{gray!10}
\multicolumn{3}{l}{\textit{\textbf{Vision-Only Expert Models}}} \\
\midrule
DRCT & Detection & \url{https://github.com/beibuwandeluori/DRCT} \\
DIRE & Detection & \url{https://github.com/ZhendongWang6/DIRE} \\
AIDE & Detection & \url{https://github.com/shilinyan99/AIDE} \\
AlignedForensics & Detection & \url{https://github.com/AniSundar18/AlignedForensics} \\
Denoising Trajectory & Detection & \url{https://github.com/Eleven4AI/DTAD} \\
DDA & Detection & \url{https://github.com/roy-ch/Dual-Data-Alignment} \\
\bottomrule
\end{tabularx}
\caption{\textbf{Detailed information of benchmarked expert models.}}
\label{tab:expert_models}
\end{table*}

%% file: appendix/tables/B/2_mllms.tex
\begin{table*}[htbp]
\scriptsize 
\centering
\renewcommand{\arraystretch}{1.1} 
\begin{tabularx}{\linewidth}{llcX}
\toprule
\textbf{Provider} & \textbf{Model} & \textbf{Size} & \textbf{Access Links} \\
\midrule

\rowcolor{gray!10}
\multicolumn{4}{l}{\textit{\textbf{Proprietary MLLMs}}} \\
\midrule
OpenAI & GPT-5.1 & -- & \url{https://openrouter.ai/openai/gpt-5.1} \\
& GPT-4.1 & -- & \url{https://openrouter.ai/openai/gpt-4.1} \\
& OpenAI o4-mini-high & -- & \url{https://openrouter.ai/openai/o4-mini-high} \\
\midrule
Anthropic & Claude Sonnet 4.5 & -- & \url{https://openrouter.ai/anthropic/claude-sonnet-4.5} \\
\midrule
Google & Gemini 3 Pro Preview & -- & \url{https://openrouter.ai/google/gemini-3-pro-preview} \\
& Gemini 2.5 Flash & -- & \url{https://openrouter.ai/google/gemini-2.5-flash} \\
\midrule
ByteDance & Doubao-Seed-1.6 & -- & \url{https://console.volcengine.com/ark/model/detail?Id=doubao-seed-1-6} \\
& Doubao-Seed-1.6-thinking & -- & \url{https://console.volcengine.com/ark/model/detail?Id=doubao-seed-1-6-thinking} \\
& Doubao-Seed-1.6-flash & -- & \url{https://console.volcengine.com/ark/model/detail?Id=doubao-seed-1-6-flash} \\
& Doubao-1.5-thinking-vision-pro & -- & \url{https://console.volcengine.com/ark/model/detail?Id=doubao-1-5-thinking-vision-pro} \\
& Doubao-1.5-vision-pro & -- & \url{https://console.volcengine.com/ark/model/detail?Id=doubao-1-5-vision-pro} \\
\midrule
Alibaba & Qwen3-VL-Plus & -- & \url{https://bailian.console.aliyun.com/cn-beijing?tab=model\#/model-market/detail/qwen3-vl-plus?serviceSite=asia-pacific-china} \\
& Qwen-VL-Max & -- & \url{https://bailian.console.aliyun.com/cn-beijing?tab=model\#/model-market/detail/qwen-vl-max?serviceSite=asia-pacific-china} \\
& Qwen-VL-Plus & -- & \url{https://bailian.console.aliyun.com/cn-beijing?tab=model\#/model-market/detail/qwen-vl-plus?serviceSite=asia-pacific-china} \\
\midrule

\rowcolor{gray!10}
\multicolumn{4}{l}{\textit{\textbf{Open-Source MLLMs}}} \\
\midrule
Google & Gemma 3 27B & 27B & \url{https://openrouter.ai/google/gemma-3-27b-it} \\
\midrule
Alibaba & Qwen2.5-VL-72B & 72B & \url{https://bailian.console.aliyun.com/cn-beijing?tab=model\#/model-market/detail/qwen2.5-vl-72b-instruct?serviceSite=asia-pacific-china} \\
\midrule
Meta & Llama 4 Maverick & 400B-A17B & \url{https://openrouter.ai/meta-llama/llama-4-maverick} \\
\midrule
Mistral AI & Ministral 3 14B & 14B & \url{https://openrouter.ai/mistralai/ministral-14b-2512} \\
\midrule
LLaVA Community & LLaVA-NeXT-7B & 7B & \url{https://huggingface.co/llava-hf/llava-v1.6-vicuna-7b-hf} \\
& LLaVA-NeXT-34B & 34B & \url{https://huggingface.co/llava-hf/llava-v1.6-34b-hf} \\
& LLaVA-Interleave-7B & 7B & \url{https://huggingface.co/llava-hf/llava-interleave-qwen-7b-hf} \\
& LLaVA-Interleave-7B-DPO & 7B & \url{https://huggingface.co/llava-hf/llava-interleave-qwen-7b-dpo-hf} \\
& LLaVA-OneVision-7B & 7B & \url{https://huggingface.co/lmms-lab/llava-onevision-qwen2-7b-ov} \\
& LLaVA-OneVision-72B & 72B & \url{https://huggingface.co/lmms-lab/llava-onevision-qwen2-72b-ov} \\
& LLaVA-OneVision-7B-Chat & 7B & \url{https://huggingface.co/lmms-lab/llava-onevision-qwen2-7b-ov-chat} \\
\midrule

\rowcolor{gray!10}
\multicolumn{4}{l}{\textit{\textbf{Unified Multimodal Understanding and Generation Model}}} \\
\midrule
DeepSeek & Janus-Pro-7B & 7B & \url{https://huggingface.co/deepseek-ai/Janus-Pro-7B} \\
\bottomrule
\end{tabularx}
\caption{\textbf{Detailed information of benchmarked MLLMs.}}
\label{tab:mllms}
\end{table*}

%% file: appendix/tables/B/3_nfi_gamma_sens.tex
\begin{table*}[t]
\centering
\scriptsize
\renewcommand{\arraystretch}{1.25}
\adjustbox{width=0.8\textwidth}{%
\begin{tabular}{l cc cc cc}
\toprule
\multirow{2}{*}{\textbf{Model}} 
& \multicolumn{2}{c}{\textbf{$\gamma=0.3$}} 
& \multicolumn{2}{c}{\textbf{$\gamma=0.5$}} 
& \multicolumn{2}{c}{\textbf{$\gamma=0.7$}} \\
\cmidrule(lr){2-3}
\cmidrule(lr){4-5}
\cmidrule(lr){6-7}
& \textbf{NFI} & \textbf{Rank} 
& \textbf{NFI} & \textbf{Rank} 
& \textbf{NFI} & \textbf{Rank} \\

\midrule
\rowcolor{gray!10}
\multicolumn{7}{c}{\textbf{Proprietary MLLMs}} \\
\midrule

GPT-5.1 & 50.09 & 1 & 48.80 & 1 & 47.54 & 1 \\
GPT-4.1 & 43.32 & 5 & 43.31 & 5 & 43.30 & 5 \\
OpenAI o4-mini-high & 43.25 & 6 & 42.77 & 7 & 42.30 & 7 \\
Claude Sonnet 4.5 & 28.37 & 18 & 26.83 & 19 & 25.38 & 19 \\
Gemini 3 Pro Preview & 47.89 & 3 & 45.79 & 4 & 43.78 & 4 \\
Gemini 2.5 Flash & 48.47 & 2 & 47.02 & 2 & 45.62 & 3 \\
Doubao-Seed-1.6 & 41.75 & 9 & 41.73 & 9 & 41.70 & 9 \\
Doubao-Seed-1.6-thinking & 46.72 & 4 & 46.66 & 3 & 46.60 & 2 \\
Doubao-Seed-1.6-flash & 36.16 & 12 & 36.03 & 11 & 35.90 & 11 \\
Doubao-1.5-thinking-vision-pro & 42.54 & 8 & 42.39 & 8 & 42.25 & 8 \\
Doubao-1.5-vision-pro & 34.21 & 14 & 34.07 & 14 & 33.94 & 14 \\
Qwen3-VL-Plus & 16.55 & 22 & 16.53 & 21 & 16.51 & 21 \\
Qwen-VL-Max & 43.12 & 7 & 43.01 & 6 & 42.89 & 6 \\
Qwen-VL-Plus & 33.98 & 15 & 33.89 & 16 & 33.80 & 16 \\

\midrule
\rowcolor{gray!10}
\multicolumn{7}{c}{\textbf{Open-Source MLLMs}} \\
\midrule

Gemma 3 27B & 40.16 & 10 & 37.55 & 10 & 35.11 & 13 \\
Qwen2.5-VL-72B & 38.75 & 11 & 38.71 & 10 & 38.67 & 10 \\
Llama 4 Maverick & 31.01 & 16 & 29.15 & 17 & 27.40 & 17 \\
Ministral 3 14B & 21.51 & 20 & 19.90 & 20 & 18.40 & 20 \\
LLaVA-NeXT-7B & 19.54 & 21 & 16.53 & 21 & 13.99 & 24 \\
LLaVA-NeXT-34B & 28.37 & 18 & 27.74 & 18 & 27.12 & 18 \\
LLaVA-Interleave-7B & 13.81 & 23 & 13.61 & 24 & 13.41 & 24 \\
LLaVA-Interleave-7B-DPO & 28.61 & 17 & 28.18 & 16 & 27.75 & 16 \\
LLaVA-OneVision-7B & 11.63 & 26 & 11.45 & 26 & 11.28 & 26 \\
LLaVA-OneVision-72B & 35.10 & 13 & 33.81 & 15 & 32.57 & 15 \\
LLaVA-OneVision-7B-Chat & 12.36 & 25 & 12.19 & 25 & 12.03 & 25 \\

\midrule
\rowcolor{gray!10}
\multicolumn{7}{c}{\textbf{Unified Multimodal Understanding and Generation Model}} \\
\midrule

Janus-Pro-7B & 13.78 & 24 & 13.77 & 23 & 13.76 & 23 \\
\bottomrule
\end{tabular}}
\caption{\textbf{Sensitivity of NFI to the OLR penalty exponent $\gamma$.} Lower ranks indicate better performance.}
\label{tab:nfi_gamma_sens}
\end{table*}

%% file: appendix/tables/C/2_fewshot_FSD.tex
\begin{table*}[t]
\footnotesize
\centering
\setlength{\tabcolsep}{3pt}
\renewcommand{\arraystretch}{1.25}
\begin{tabular}{l | c | c | c | c | c | c}
\toprule
\textbf{Model} & \textbf{Prompting} & \textbf{ACC} & \textbf{M-F1} & \textbf{No-Forgery} & \textbf{Entire-F1} & \textbf{Partial-F1} \\
\midrule

\multirow{2}{*}{GPT-5.1} & None
& 50.99 & 46.16 & 52.54 & 74.70 & 11.25 \\
& Few-Shot
& 64.67 \textcolor{markyellow}{(+13.68)}
& 57.67 \textcolor{markyellow}{(+11.51)}
& 59.34 \textcolor{markyellow}{(+6.80)}
& 79.78 \textcolor{markyellow}{(+5.08)}
& 33.90 \textcolor{markyellow}{(+22.65)} \\
\midrule

\multirow{2}{*}{Doubao-Seed-1.6-thinking} & None
& 44.57 & 37.13 & 41.97 & 66.44 & 2.97 \\
& Few-Shot
& 62.60 \textcolor{markyellow}{(+18.03)}
& 55.51 \textcolor{markyellow}{(+18.38)}
& 57.06 \textcolor{markyellow}{(+15.09)}
& 76.72 \textcolor{markyellow}{(+10.28)}
& 32.75 \textcolor{markyellow}{(+29.78)} \\
\midrule

\multirow{2}{*}{Qwen3-VL-Plus} & None
& 38.77 & 35.72 & 41.23 & 50.12 & 15.82 \\
& Few-Shot
& 69.97 \textcolor{markyellow}{(+31.20)}
& 65.20 \textcolor{markyellow}{(+29.48)}
& 50.42 \textcolor{markyellow}{(+9.19)}
& 80.07 \textcolor{markyellow}{(+29.95)}
& 65.09 \textcolor{markyellow}{(+49.27)} \\
\bottomrule
\end{tabular}

\caption{\textbf{Forgery Scope Discrimination} performance under \textbf{Few-Shot prompting} strategies.}
\label{tab:appendix_fewshot_fsd}
\end{table*}

%% file: appendix/tables/C/2_fewshot_TAR.tex
\begin{table*}[t]
\footnotesize
\centering
\setlength{\tabcolsep}{3pt}
\renewcommand{\arraystretch}{1.25}
\begin{tabular}{l | c | c | c | c | c }
\toprule
\textbf{Model} & \textbf{Prompting} & \textbf{ACC} & \textbf{M-F1} & \textbf{Yes-F1} & \textbf{No-F1} \\
\midrule

\multirow{2}{*}{GPT-5.1} & None
& 76.43 & 76.38 & 77.50 & 75.25 \\
& Few-Shot
& 84.23 \textcolor{markyellow}{(+7.80)}
& 83.37 \textcolor{markyellow}{(+6.99)}
& 87.16 \textcolor{markyellow}{(+9.66)}
& 79.57 \textcolor{markyellow}{(+4.32)} \\
\midrule

\multirow{2}{*}{Doubao-Seed-1.6-thinking} & None
& 77.92 & 77.57 & 80.38 & 74.76 \\
& Few-Shot
& 80.25 \textcolor{markyellow}{(+2.33)}
& 78.98 \textcolor{markyellow}{(+1.41)}
& 84.14 \textcolor{markyellow}{(+3.76)}
& 73.83 \textcolor{markblue}{(-0.93)} \\
\midrule

\multirow{2}{*}{Qwen3-VL-Plus} & None
& 79.25 & 78.98 & 81.38 & 76.59 \\
& Few-Shot
& 81.24 \textcolor{markyellow}{(+1.99)}
& 80.63 \textcolor{markyellow}{(+1.65)}
& 84.06 \textcolor{markyellow}{(+2.68)}
& 77.19 \textcolor{markyellow}{(+0.60)} \\
\bottomrule
\end{tabular}

\caption{\textbf{Textual Artifact Recognition} performance under \textbf{Few-Shot prompting} strategies.}
\label{tab:appendix_fewshot_tar}
\end{table*}

%% file: appendix/tables/C/2_fewshot_MC.tex
\begin{table*}[t]
\footnotesize
\centering
\setlength{\tabcolsep}{3pt}
\renewcommand{\arraystretch}{1.25}
\begin{tabular}{l | c | c | c | c | c | c}
\toprule
\textbf{Model} & \textbf{Prompting} & \textbf{ACC} & \textbf{M-F1} & \textbf{Insert-F1} & \textbf{Remove-F1} & \textbf{Alter-F1} \\
\midrule

\multirow{2}{*}{GPT-5.1} & None
& 60.07 & 48.34 & 34.93 & 36.65 & 73.43 \\
& Few-Shot
& 27.22 \textcolor{markblue}{(-32.85)}
& 30.53 \textcolor{markblue}{(-17.81)}
& 18.18 \textcolor{markblue}{(-16.75)}
& 42.25 \textcolor{markyellow}{(+5.60)}
& 31.15 \textcolor{markblue}{(-42.28)} \\
\midrule

\multirow{2}{*}{Doubao-Seed-1.6-thinking} & None
& 58.47 & 41.85 & 26.10 & 26.00 & 73.43 \\
& Few-Shot
& 25.80 \textcolor{markblue}{(-32.67)}
& 23.75 \textcolor{markblue}{(-18.10)}
& 23.88 \textcolor{markblue}{(-2.22)}
& 23.46 \textcolor{markblue}{(-2.54)}
& 23.90 \textcolor{markblue}{(-49.53)} \\
\midrule

\multirow{2}{*}{Qwen3-VL-Plus} & None
& 59.28 & 36.89 & 15.54 & 20.37 & 74.75 \\
& Few-Shot
& 33.75 \textcolor{markblue}{(-25.53)}
& 31.52 \textcolor{markblue}{(-5.37)}
& 20.80 \textcolor{markyellow}{(+5.26)}
& 25.60 \textcolor{markyellow}{(+5.23)}
& 48.15 \textcolor{markblue}{(-26.60)} \\
\bottomrule
\end{tabular}

\caption{\textbf{Manipulation Classification} performance under \textbf{Few-Shot prompting} strategies.}
\label{tab:appendix_fewshot_mc}
\end{table*}

%% file: appendix/tables/C/3_cot.tex
\begin{table*}[t]
\small
\centering
\setlength{\tabcolsep}{4pt}
\renewcommand{\arraystretch}{1.25}
\begin{tabular}{l|c|cc|cc|cc|cc}
\toprule
\multirow{3}{*}{\textbf{Category}} & 
\multirow{3}{*}{\textbf{Prompting}} & 
\multicolumn{4}{c|}{\textbf{Detection}} & 
\multicolumn{4}{c}{\textbf{Localization}} \\
\cline{3-10} 
& &
\multicolumn{2}{c|}{\textbf{FSD}} &
\multicolumn{2}{c|}{\textbf{TAR}} &
\multicolumn{2}{c|}{\textbf{MC}} &
\multicolumn{2}{c}{\textbf{TP}} \\
\cline{3-10} 
& &
\textbf{ACC} &
\textbf{M-F1} & 
\textbf{ACC} &
\textbf{M-F1} & 
\textbf{ACC} &
\textbf{M-F1} & 
\textbf{CLA} &
\textbf{OLR} \\
\midrule

\multirow{3}{*}{Chart} & None 
& 41.10 & 41.99 & 77.34 & 77.30 
& 70.73 & 48.50 & 50.47 & 10.17 \\

& \multirow{2}{*}{CoT} & 
50.50 & 44.98 & 80.81 & 79.77 & 
62.42 & 50.64 & 53.67 & 11.86 \\

& & 
\textcolor{markyellow}{(+9.40)}  &  
\textcolor{markyellow}{(+2.99)}  & 
\textcolor{markyellow}{(+3.47)}  & 
\textcolor{markyellow}{(+2.47)}  &
\textcolor{markblue}{(-8.31)} & 
\textcolor{markyellow}{(+2.14)} & 
\textcolor{markyellow}{(+3.20)} & 
\textcolor{markblue}{(+1.69)} \\

\midrule
\multirow{3}{*}{Diagram} & None 
& 65.35 & 55.49 & 80.19 & 79.35 & 
58.57 & 46.57 & 40.50 & 11.83 \\

& \multirow{2}{*}{CoT} & 
73.35 & 62.15 & 83.75 & 82.60 &
47.31 & 40.22 & 50.81 & 10.27 \\

& & \textcolor{markyellow}{(+8.00)}  &  
\textcolor{markyellow}{(+6.66)}  & 
\textcolor{markyellow}{(+3.56)}  & 
\textcolor{markyellow}{(+3.25)}  &
\textcolor{markblue}{(-11.26)} & 
\textcolor{markblue}{(-6.35)} & 
\textcolor{markyellow}{(+10.31)} & 
\textcolor{markyellow}{(-1.56)} \\

\midrule
\multirow{3}{*}{Micrograph} & None & 
39.09 & 43.33 & 71.98 & 64.39 & 
55.30 & 45.05 & 46.13 & 13.18 \\

& \multirow{2}{*}{CoT} & 
56.78 & 52.02 & 73.95 & 68.85 & 
36.73 & 35.03 & 51.36 & 15.18\\

& & \textcolor{markyellow}{(+17.69)}  &  
\textcolor{markyellow}{(+8.69)}  & 
\textcolor{markyellow}{(+1.97)}  & 
\textcolor{markyellow}{(+4.46)}  &
\textcolor{markblue}{(-18.57)} & 
\textcolor{markblue}{(-10.02)} & 
\textcolor{markyellow}{(+5.23)} & 
\textcolor{markblue}{(+2.00)} \\

\midrule
\multirow{3}{*}{Stained Micrograph} & None &
34.79 & 42.16 & 57.72 & 52.90 &
50.47 & 47.21 & 53.43 & 16.31 \\

& \multirow{2}{*}{CoT} & 
52.26 & 48.68 & 58.76 & 55.83 & 
35.20 & 34.46 & 54.61 & 10.64\\

& & \textcolor{markyellow}{(+17.47)} &
\textcolor{markyellow}{(+6.52)} & 
\textcolor{markyellow}{(+1.04)} & 
\textcolor{markyellow}{(+2.93)} &
\textcolor{markblue}{(-15.27)} & 
\textcolor{markblue}{(-12.75)} & 
\textcolor{markyellow}{(+1.18)} & 
\textcolor{markyellow}{(-5.67)} \\

\midrule
\multirow{3}{*}{Physical Object} & None & 
48.49 & 47.81 & 60.77 & 57.93 & 
59.23 & 42.01 & 42.58 & 9.46 \\

& \multirow{2}{*}{CoT} & 
57.66 & 51.05 & 65.74 & 64.39 & 
40.16 & 36.64 & 40.65 & 14.84 \\

& & \textcolor{markyellow}{(+9.17)}  &  
\textcolor{markyellow}{(+3.24)}  & 
\textcolor{markyellow}{(+4.97)}  & 
\textcolor{markyellow}{(+6.46)}  &
\textcolor{markblue}{(-19.07)} & 
\textcolor{markblue}{(-5.37)} & 
\textcolor{markblue}{(-1.93)} & 
\textcolor{markblue}{(+5.38)} \\

\midrule
\multirow{3}{*}{Medical Imaging} & None & 
48.92 & 50.31 & 75.12 & 74.78 & 
59.47 & 45.47 & 49.81 & 13.69 \\

& \multirow{2}{*}{CoT} & 
64.84 & 60.42 & 79.66 & 79.64 & 
32.47 & 32.60 & 54.02 & 21.84\\

& & \textcolor{markyellow}{(+15.92)}  &  
\textcolor{markyellow}{(+10.11)}  & 
\textcolor{markyellow}{(+4.54)}  & 
\textcolor{markyellow}{(+4.86)}  &
\textcolor{markblue}{(-27.00)} & 
\textcolor{markblue}{(-12.87)} & 
\textcolor{markyellow}{(+4.21)} & 
\textcolor{markblue}{(+8.15)} \\

\midrule
\multirow{3}{*}{Others} & None & 
61.89 & 51.55 & 83.00 & 82.78 & 
66.67 & 57.37 & 48.40 & 13.24 \\

& \multirow{2}{*}{CoT} & 
67.80 & 59.61 & 86.79 & 86.57 & 
58.33 & 55.46 & 55.95 & 10.71\\

& & \textcolor{markyellow}{(+5.91)}  & 
\textcolor{markyellow}{(+8.06)}  & 
\textcolor{markyellow}{(+3.79)}  & 
\textcolor{markyellow}{(+3.79)}  &
\textcolor{markblue}{(-8.34)} & 
\textcolor{markblue}{(-1.91)} & 
\textcolor{markyellow}{(+7.55)} & 
\textcolor{markyellow}{(-2.53)} \\

\midrule
\multirow{3}{*}{Avg.} & None & 
48.52 & 47.52 & 72.30 & 69.92 & 
60.06 & 47.45 & 47.33 & 12.55 \\

& \multirow{2}{*}{CoT} & 
52.90 & 47.36 & 75.64 & 73.95 & 
44.66 & 40.72 & 51.58 & 13.62\\

& & \textcolor{markyellow}{(+4.38)}  &  
\textcolor{markblue}{(-0.16)}  & 
\textcolor{markyellow}{(+3.33)}  & 
\textcolor{markyellow}{(+4.03)}  &
\textcolor{markblue}{(-15.40)} & 
\textcolor{markblue}{(-6.73)} & 
\textcolor{markyellow}{(+4.25)} & 
\textcolor{markblue}{(+1.07)} \\

\bottomrule
\end{tabular}
\caption{Performance comparison of \textbf{GPT-5.1} across categories under \textbf{CoT prompting} strategies.}
\label{tab:experiment_CoT}
\end{table*}

%% file: appendix/tables/C/4_post_progress.tex
\begin{table*}[t]
\centering
\small
\setlength{\tabcolsep}{4pt}
\renewcommand{\arraystretch}{1.15}

\begin{tabular}{l | c | cc | cc | cc | cc | cc}
\toprule
\multirow{4}{*}{\textbf{Model}} 
& \multirow{4}{*}{\textbf{\makecell{Post-\\Processing}}} 
& \multicolumn{4}{c|}{\textbf{Detection}} 
& \multicolumn{6}{c}{\textbf{Localization}} \\

\cline{3-12}
& 
& \multicolumn{2}{c|}{\textbf{FSD}} 
& \multicolumn{2}{c|}{\textbf{TAR}} 
& \multicolumn{2}{c|}{\textbf{MC}} 
& \multicolumn{4}{c}{\textbf{TP}} \\

\cline{3-12}
& 
& \multirow{2}{*}{ACC} & \multirow{2}{*}{M-F1} 
& \multirow{2}{*}{ACC} & \multirow{2}{*}{M-F1} 
& \multirow{2}{*}{ACC} & \multirow{2}{*}{M-F1} 
& \multicolumn{2}{c|}{\textbf{Region-Level}} 
& \multicolumn{2}{c}{\textbf{Pixel-Level}} \\

\cline{9-12}
&  &  
&  &  
&  &  
 & 
& CLA & OLR 
& IoU & F1 \\
\midrule

\multirow{7}{*}{GPT-5.1}
& None 
& 50.99 & 46.16 
& 76.43 & 76.38 
& 60.07 & 48.34 
& 46.87 & 12.25 
& -- & -- \\

& Gauss
& \makecell{24.22 \\ \textcolor{markblue}{(-26.77)}} 
& \makecell{25.74 \\ \textcolor{markblue}{(-20.42)}} 
& \makecell{46.32 \\ \textcolor{markblue}{(-30.11)}} 
& \makecell{42.62 \\ \textcolor{markblue}{(-33.76)}} 
& \makecell{58.74 \\ \textcolor{markblue}{(-1.33)}} 
& \makecell{40.22 \\ \textcolor{markblue}{(-8.12)}} 
& \makecell{21.58 \\ \textcolor{markblue}{(-25.29)}} 
& \makecell{13.03 \\ \textcolor{markyellow}{(+0.78)}} 
& -- & -- \\

& JPEG
& \makecell{47.22 \\ \textcolor{markblue}{(-3.77)}} 
& \makecell{40.38 \\ \textcolor{markblue}{(-5.78)}} 
& \makecell{72.16 \\ \textcolor{markblue}{(-4.27)}} 
& \makecell{72.13 \\ \textcolor{markblue}{(-4.25)}} 
& \makecell{48.99 \\ \textcolor{markblue}{(-11.08)}} 
& \makecell{39.91 \\ \textcolor{markblue}{(-8.43)}} 
& \makecell{9.39 \\ \textcolor{markblue}{(-37.48)}} 
& \makecell{0.11 \\ \textcolor{markblue}{(-12.14)}} 
& -- & -- \\

& Scaling
& \makecell{49.49 \\ \textcolor{markblue}{(-1.50)}} 
& \makecell{41.25 \\ \textcolor{markblue}{(-4.91)}} 
& \makecell{71.00 \\ \textcolor{markblue}{(-5.43)}} 
& \makecell{70.99 \\ \textcolor{markblue}{(-5.39)}} 
& \makecell{59.40 \\ \textcolor{markblue}{(-0.67)}} 
& \makecell{46.91 \\ \textcolor{markblue}{(-1.43)}} 
& \makecell{11.53 \\ \textcolor{markblue}{(-35.34)}} 
& \makecell{0.12 \\ \textcolor{markblue}{(-12.13)}} 
& -- & -- \\
\midrule

\multirow{7}{*}{FakeShield}
& None 
& 59.72 & 62.08 
& -- & -- 
& -- & -- 
& -- & -- 
& 30.09 & 69.11 \\

& Gauss
& \makecell{36.73 \\ \textcolor{markblue}{(-22.99)}} 
& \makecell{34.26 \\ \textcolor{markblue}{(-27.82)}} 
& -- & -- 
& -- & -- 
& -- & -- 
& \makecell{20.77 \\ \textcolor{markblue}{(-9.32)}} 
& \makecell{29.73 \\ \textcolor{markblue}{(-39.38)}} \\

& JPEG
& \makecell{40.26 \\ \textcolor{markblue}{(-19.46)}} 
& \makecell{36.15 \\ \textcolor{markblue}{(-25.93)}} 
& -- & -- 
& -- & -- 
& -- & -- 
& \makecell{22.07 \\ \textcolor{markblue}{(-8.02)}} 
& \makecell{28.16 \\ \textcolor{markblue}{(-40.95)}} \\

& Scaling
& \makecell{38.01 \\ \textcolor{markblue}{(-21.71)}} 
& \makecell{35.04 \\ \textcolor{markblue}{(-27.04)}} 
& -- & -- 
& -- & -- 
& -- & -- 
& \makecell{22.82 \\ \textcolor{markblue}{(-7.27)}} 
& \makecell{29.09 \\ \textcolor{markblue}{(-40.02)}} \\
\bottomrule
\end{tabular}

\caption{\textbf{Impact of post-processing perturbations.}}
\label{tab:post_process}
\end{table*}

%% file: appendix/figures/case/FSD_correct.tex
\begin{figure*}[t]
    \centering
    \includegraphics[width=0.90\linewidth]{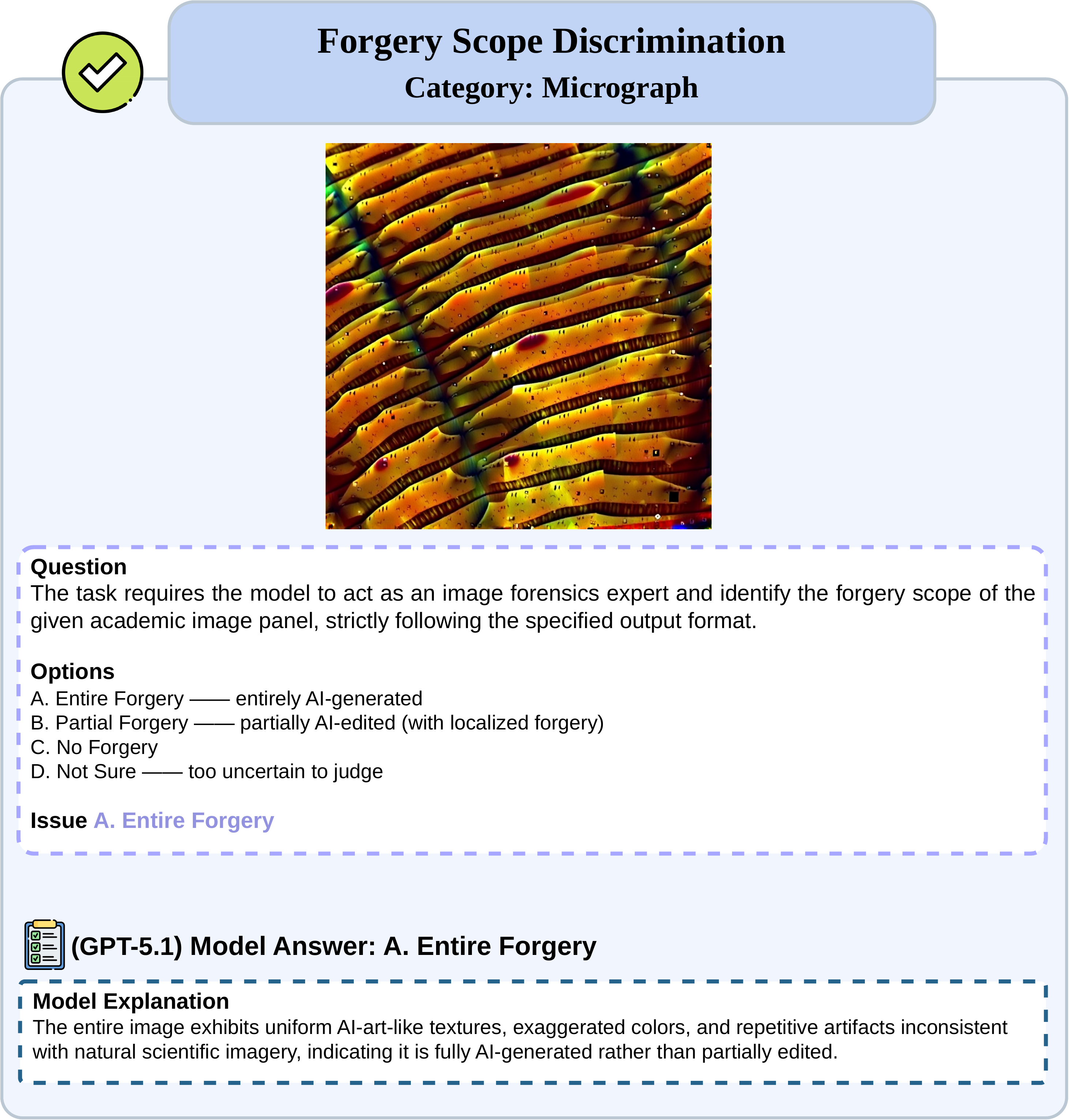}
    \caption{\textbf{Success case by GPT-5.1 on Forgery Scope Discrimination.} The panel belongs to the Micrograph category.}
    \label{fig:appendix_case_FSD_correct}
\end{figure*}

%% file: appendix/figures/case/FSD_error1.tex
\begin{figure*}[t]
    \centering
    \includegraphics[width=0.90\linewidth]{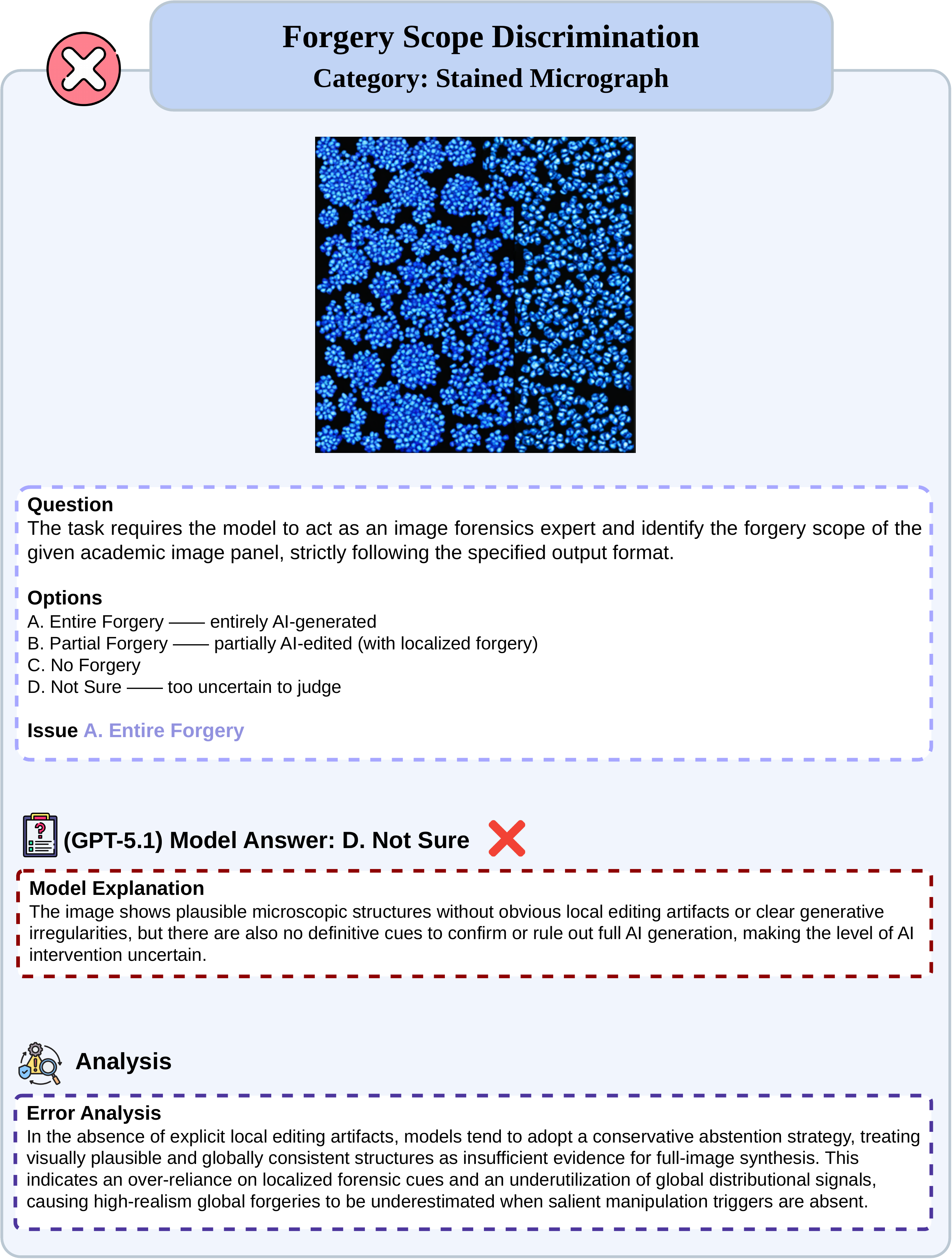}
    \caption{\textbf{Failure case by GPT-5.1 on Forgery Scope Discrimination.} The panel belongs to the Stained Micrograph category.}
    \label{fig:appendix_case_FSD_error1}
\end{figure*}

%% file: appendix/figures/case/FSD_error2.tex
\begin{figure*}[t]
    \centering
    \includegraphics[width=0.90\linewidth]{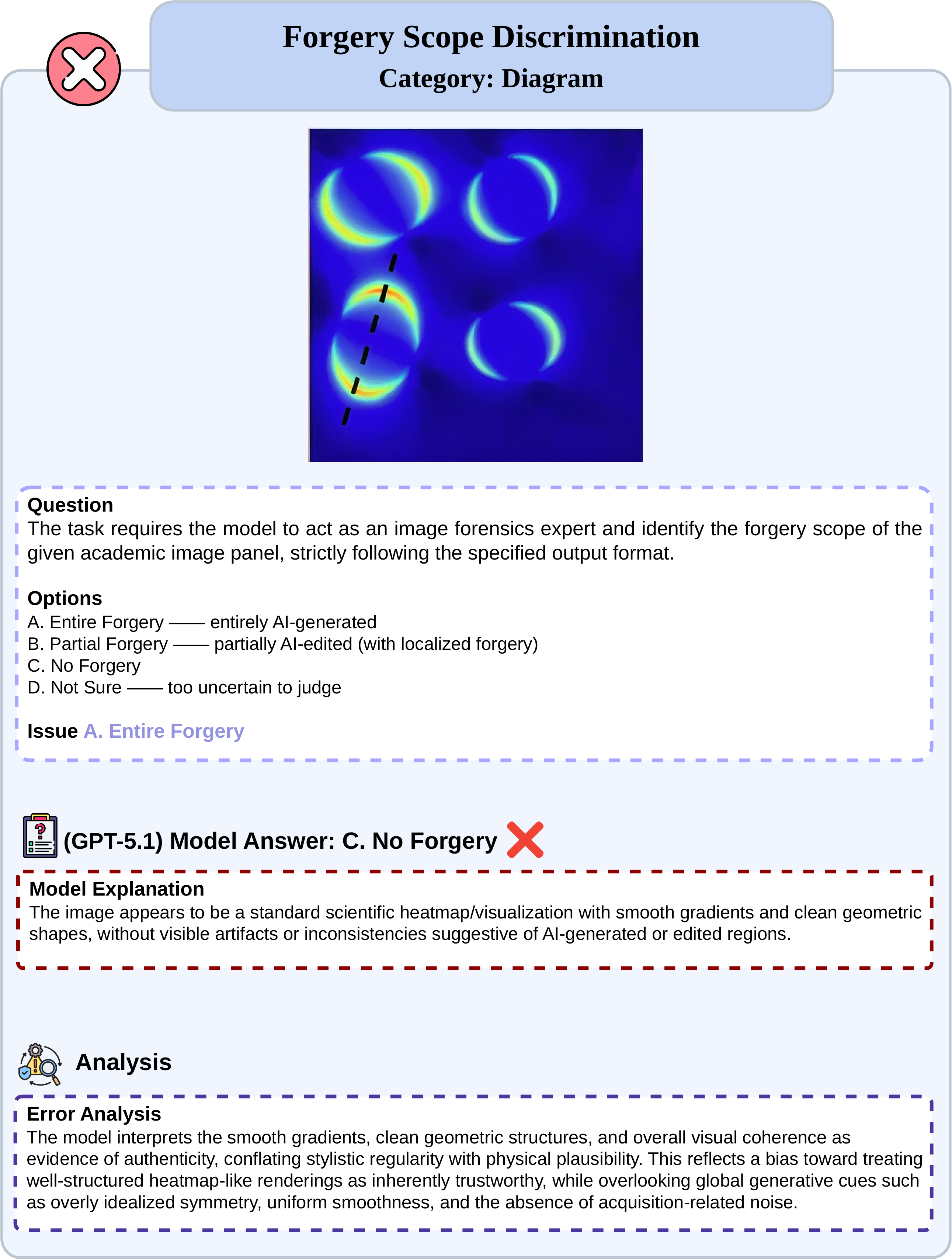}
    \caption{\textbf{Failure case by GPT-5.1 on Forgery Scope Discrimination.} The panel belongs to the Diagram category.}
    \label{fig:appendix_case_FSD_error2}
\end{figure*}

%% file: appendix/figures/case/FSD_error3.tex
\begin{figure*}[t]
    \centering
    \includegraphics[width=0.90\linewidth]{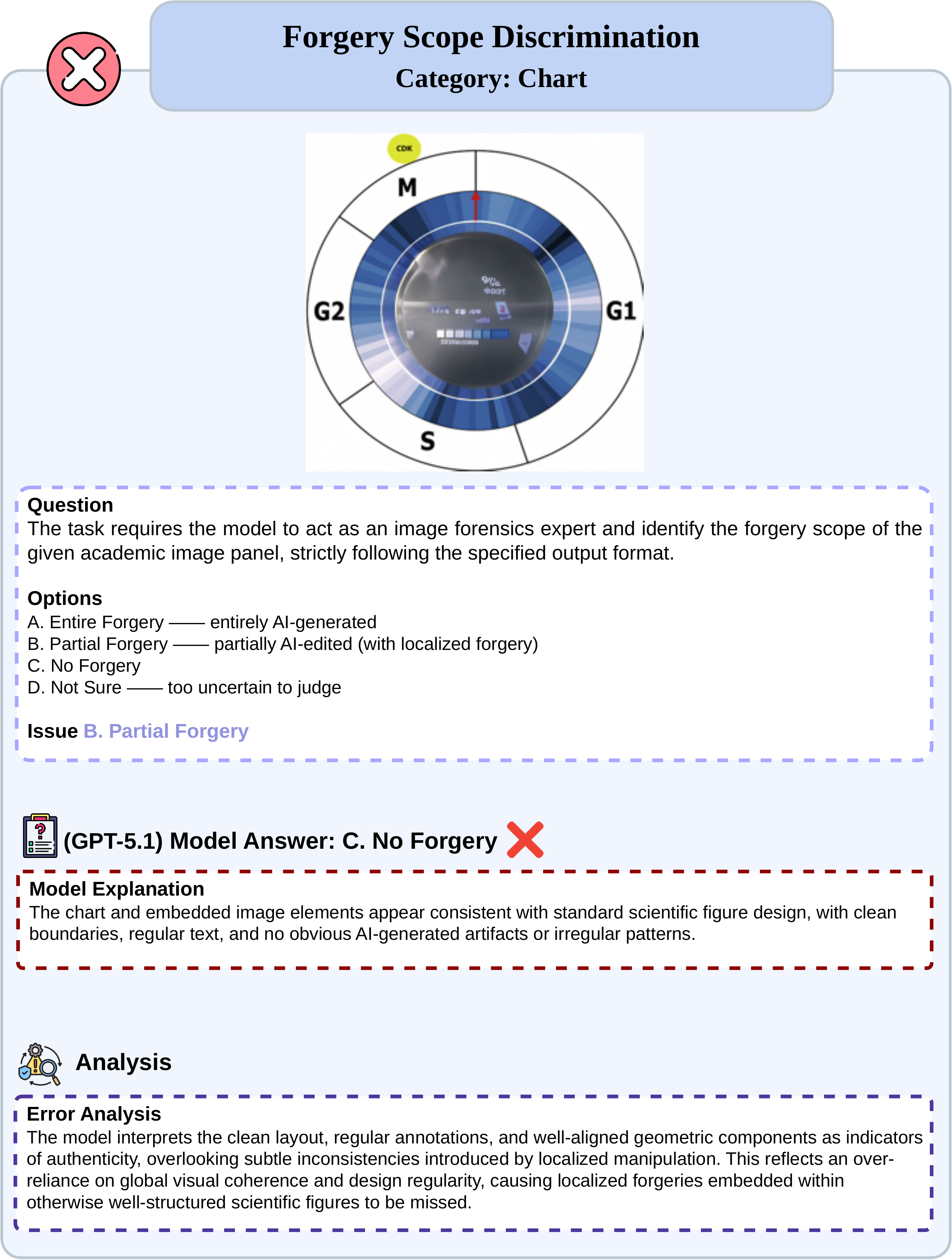}
    \caption{\textbf{Failure case by GPT-5.1 on Forgery Scope Discrimination.} The panel belongs to the Chart category.}
    \label{fig:appendix_case_FSD_error3}
\end{figure*}

%% file: appendix/figures/case/FSD_error4.tex
\begin{figure*}[t]
    \centering
    \includegraphics[width=0.90\linewidth]{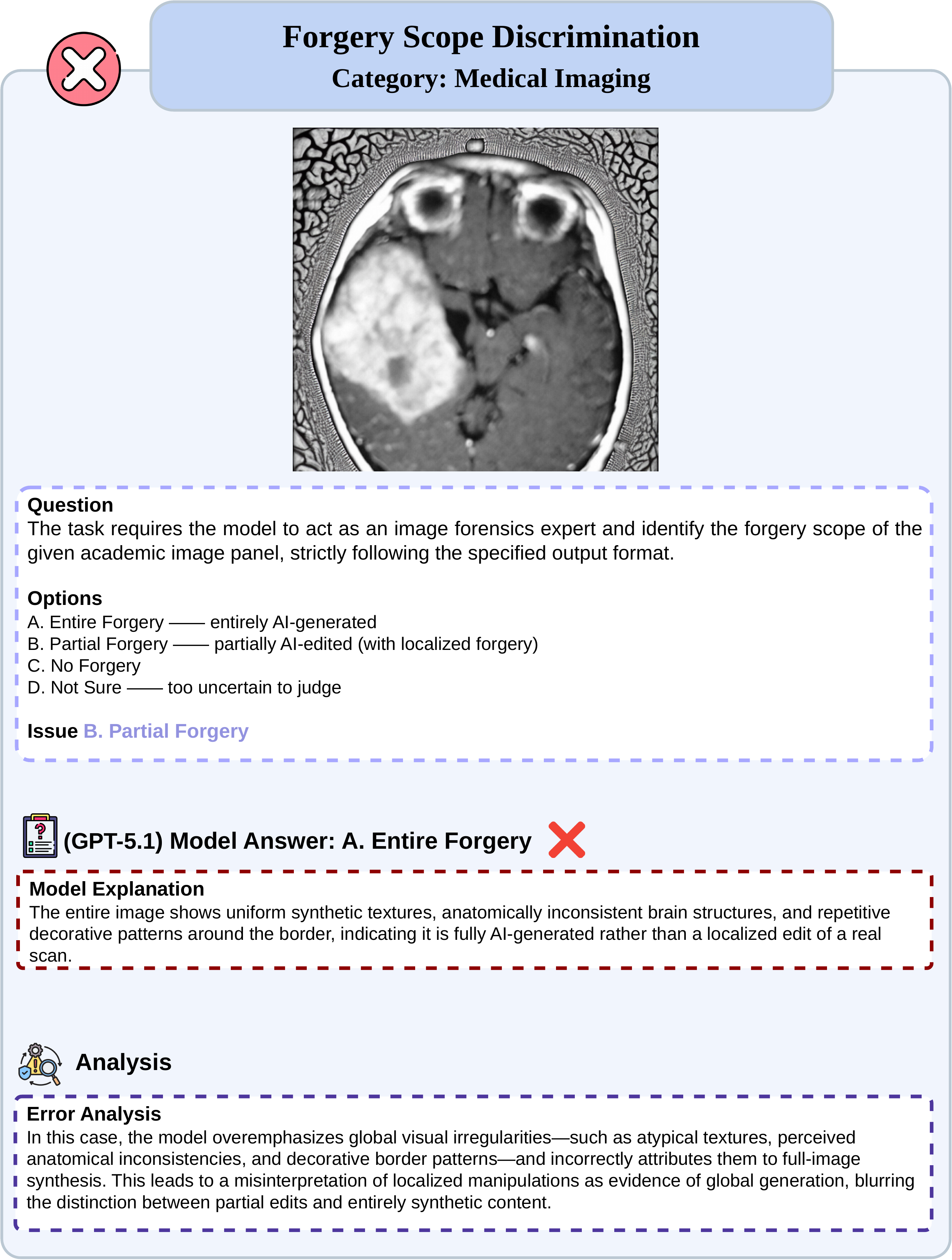}
    \caption{\textbf{Failure case by GPT-5.1 on Forgery Scope Discrimination.} The panel belongs to the Medical Imaging category.}
    \label{fig:appendix_case_FSD_error4}
\end{figure*}

%% file: appendix/figures/case/TAR_correct.tex
\begin{figure*}[t]
    \centering
    \includegraphics[width=0.90\linewidth]{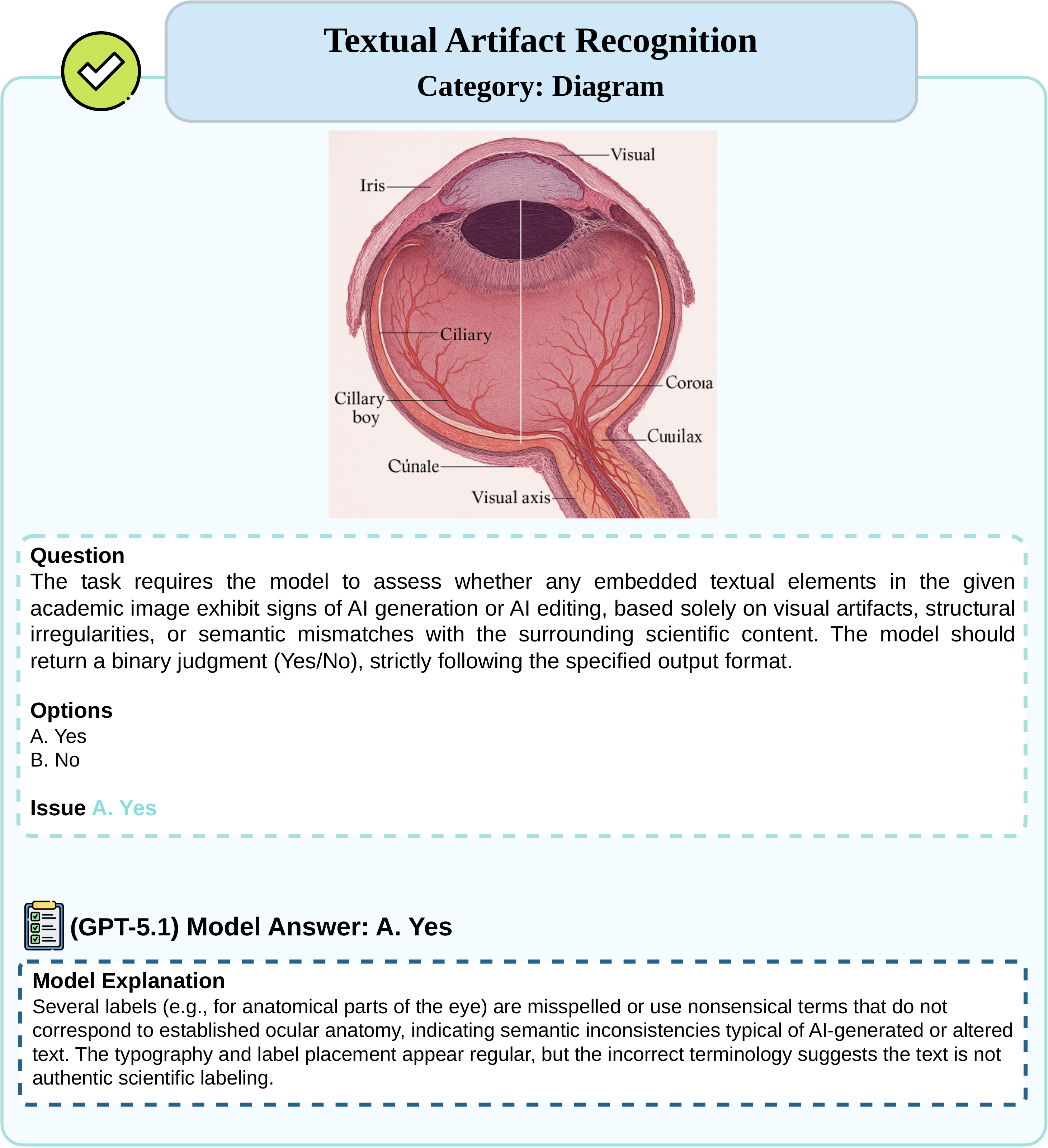}
    \caption{\textbf{Success case by GPT-5.1 on Textual Artifact Recognition.} The panel belongs to the Diagram category.}
    \label{fig:appendix_case_TAR_correct}
\end{figure*}

%% file: appendix/figures/case/TAR_error.tex
\begin{figure*}[t]
    \centering
    \includegraphics[width=0.90\linewidth]{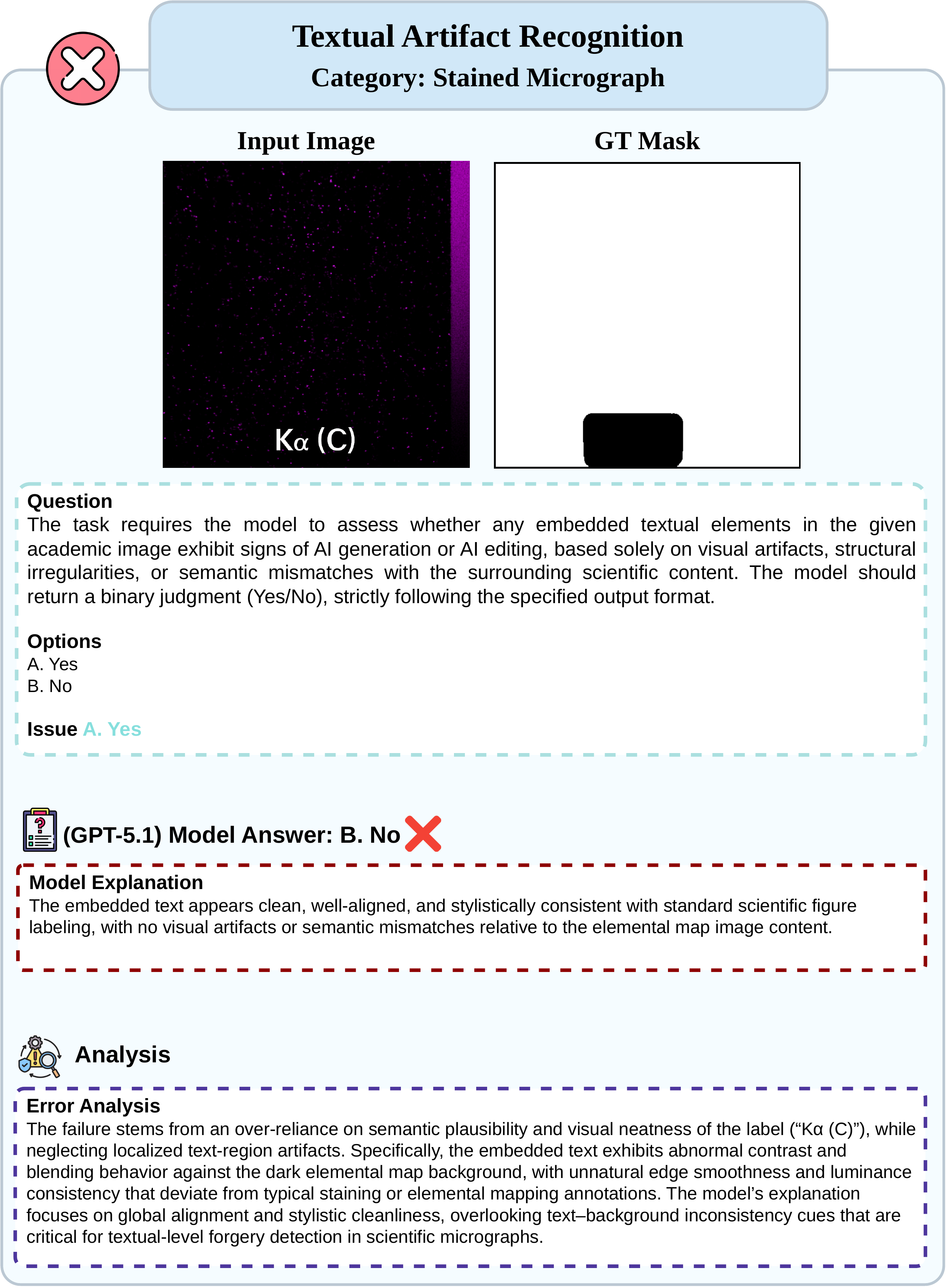}
    \caption{\textbf{Failure case by GPT-5.1 on Textual Artifact Recognition.} The panel belongs to the Stained Micrograph category and contains a localized forgery generated via \textit{Targeted Region Editing}.}
    \label{fig:appendix_case_TAR_error}
\end{figure*}

%% file: appendix/figures/case/MC_correct.tex
\begin{figure*}[t]
    \centering
    \includegraphics[width=0.90\linewidth]{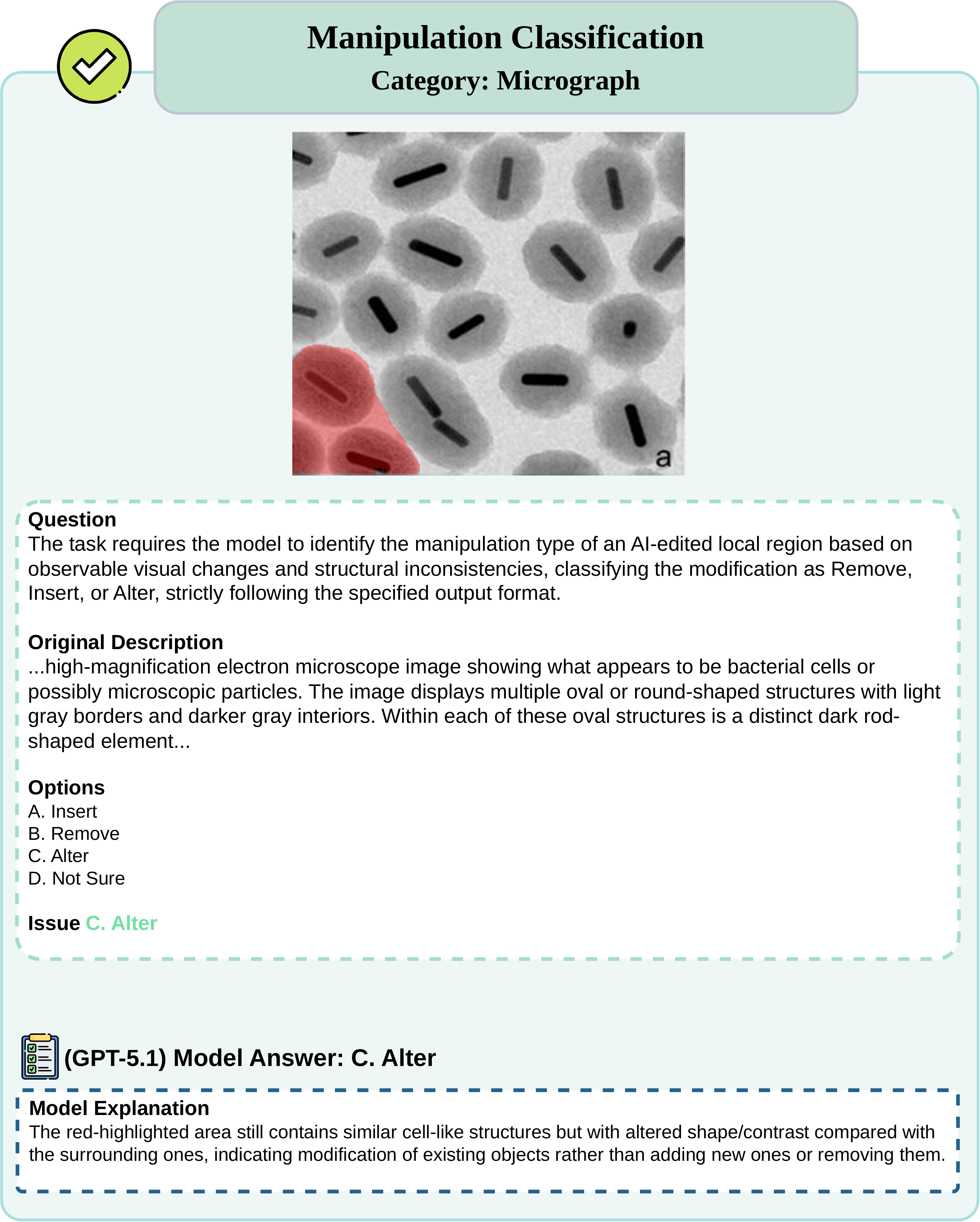}
    \caption{\textbf{Success case by GPT-5.1 on Manipulation Classification.} The panel belongs to the Micrograph category and contains a localized forgery generated via \textit{Targeted Region Editing}.}
    \label{fig:appendix_case_MC_correct}
\end{figure*}

%% file: appendix/figures/case/MC_error1.tex
\begin{figure*}[t]
    \centering
    \includegraphics[width=0.90\linewidth]{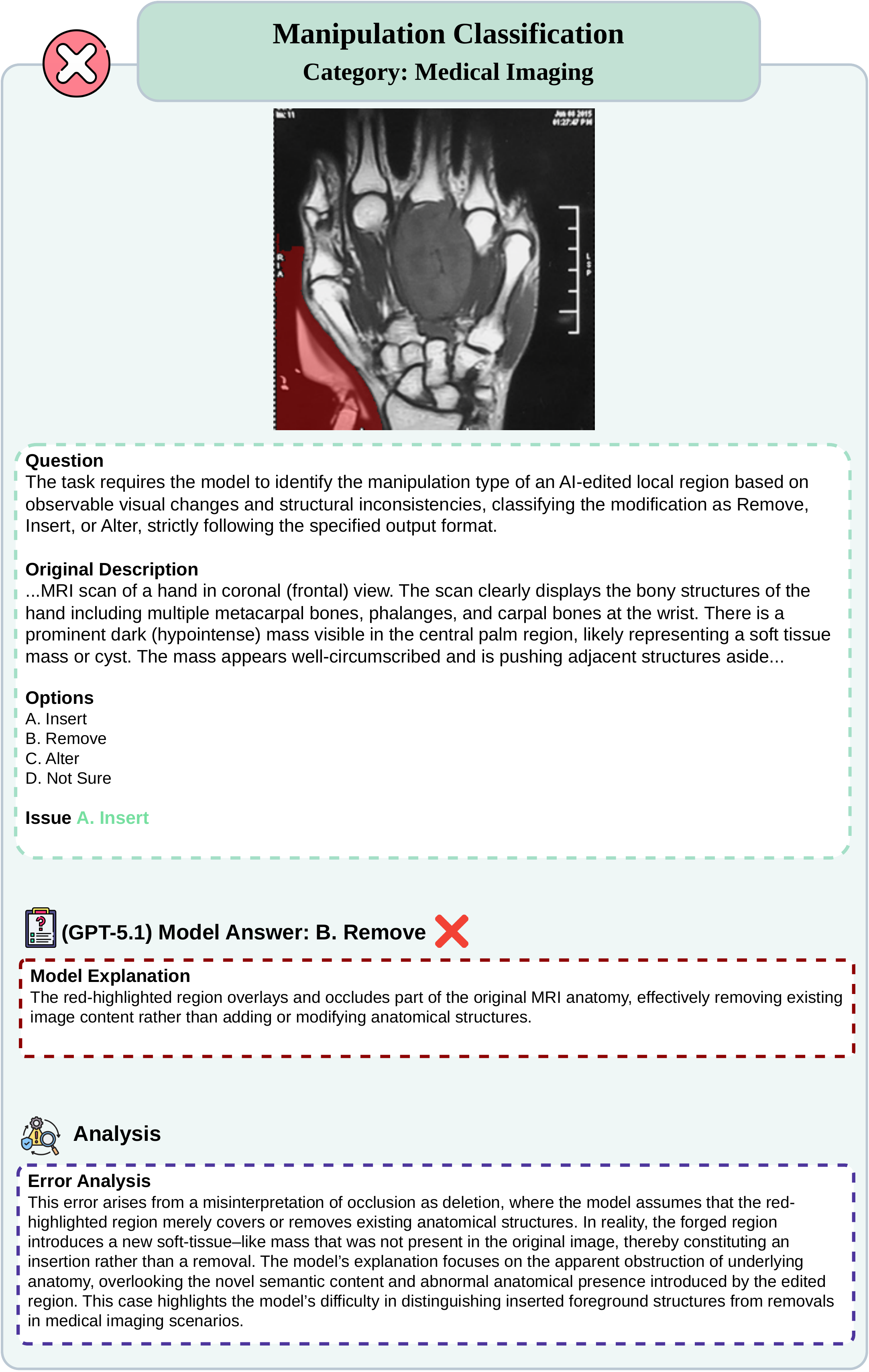}
    \caption{\textbf{Failure case by GPT-5.1 on Manipulation Classification.} The panel belongs to the Medical Imaging category and contains a localized forgery generated via \textit{Targeted Region Restoration}.}
    \label{fig:appendix_case_MC_error1}
\end{figure*}

%% file: appendix/figures/case/MC_error2.tex
\begin{figure*}[t]
    \centering
    \includegraphics[width=0.90\linewidth]{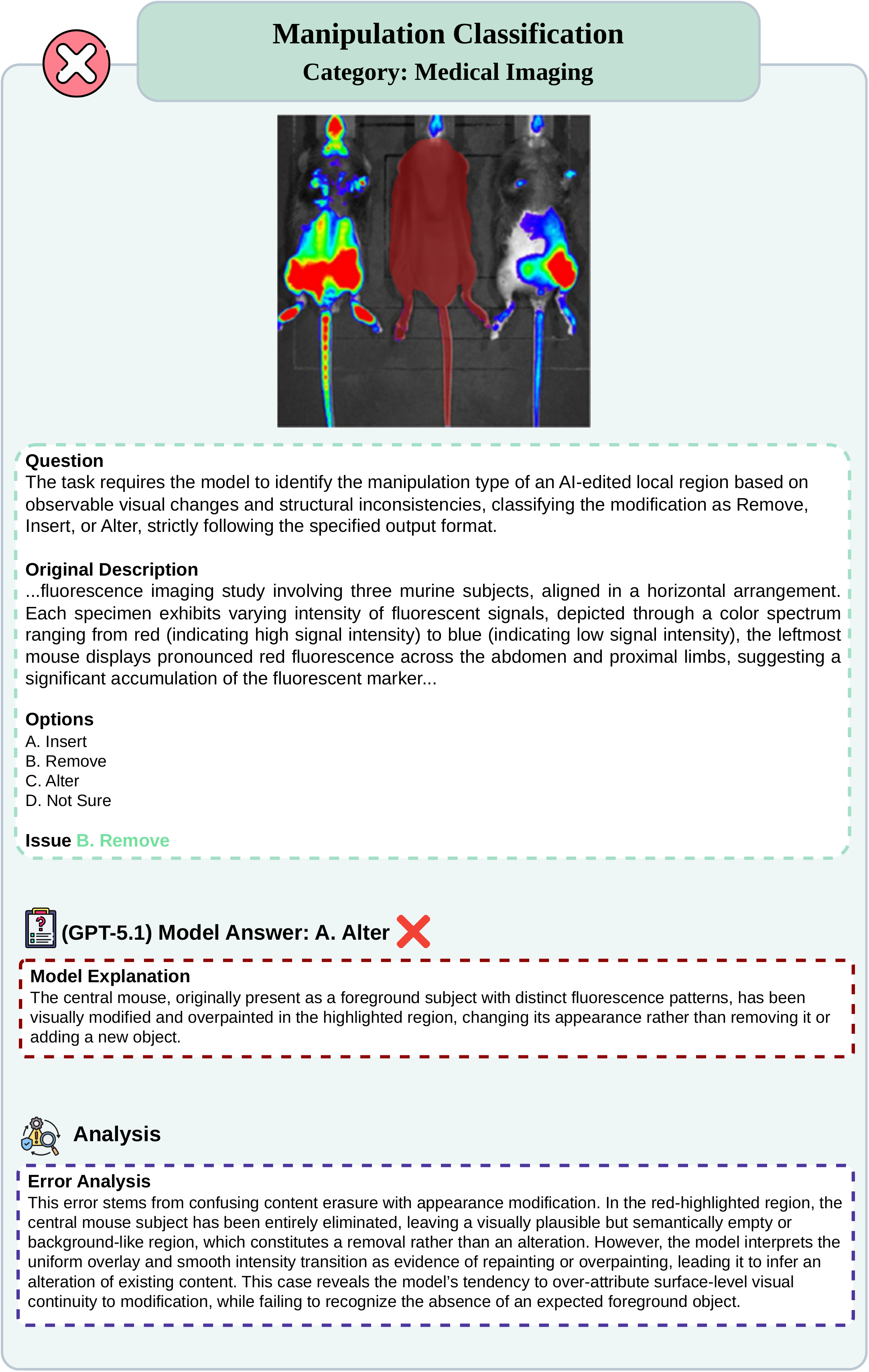}
    \caption{\textbf{Failure case by GPT-5.1 on Manipulation Classification.} The panel belongs to the Medical Imaging category and contains a localized forgery generated via \textit{Targeted Region Restoration}.}
    \label{fig:appendix_case_MC_error2}
\end{figure*}

%% file: appendix/figures/case/TP_correct.tex
\begin{figure*}[t]
    \centering
    \includegraphics[width=0.90\linewidth]{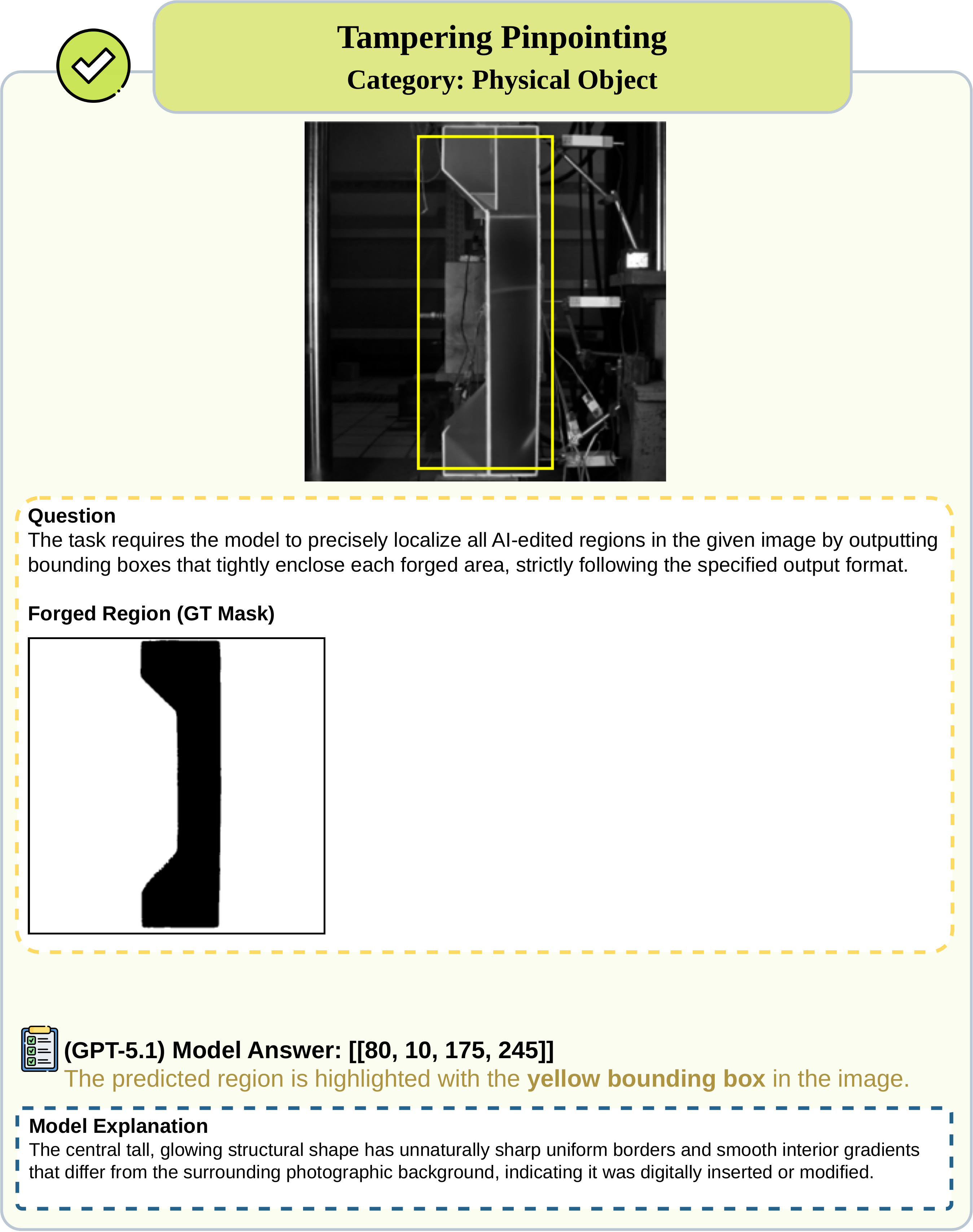}
    \caption{\textbf{Success case by GPT-5.1 on Tampering Pinpointing.} The panel belongs to the Physical Object category and contains a localized forgery generated via \textit{Targeted Region Restoration}.}
    \label{fig:appendix_case_TP_correct}
\end{figure*}

%% file: appendix/figures/case/TP_error1.tex
\begin{figure*}[t]
    \centering
    \includegraphics[width=0.90\linewidth]{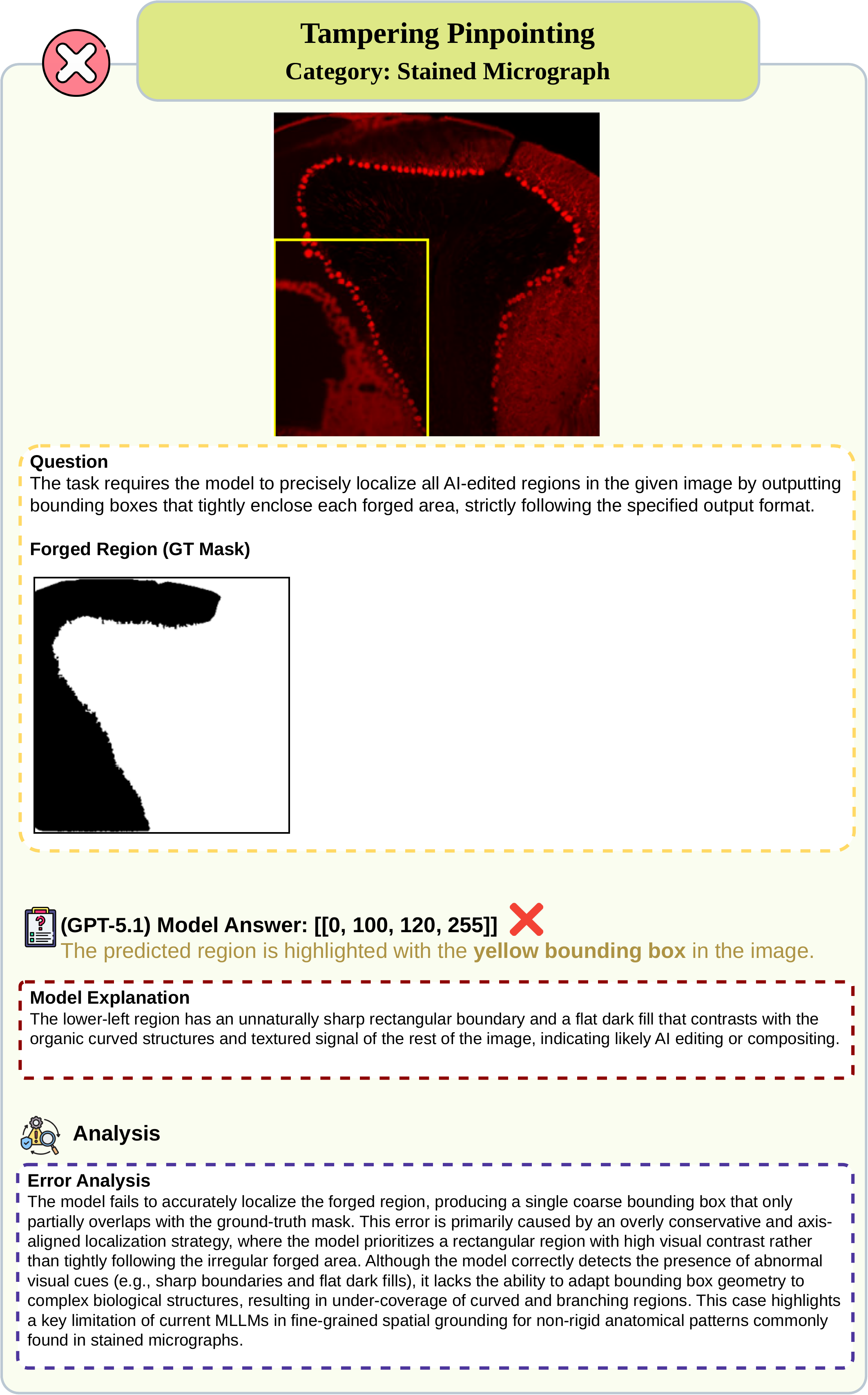}
    \caption{\textbf{Failure case by GPT-5.1 on Tampering Pinpointing.} The panel belongs to the Stained Micrograph category and contains a localized forgery generated via \textit{Targeted Region Restoration}.}
    \label{fig:appendix_case_TP_error1}
\end{figure*}